# Land use/land cover classification of fused Sentinel-1 and Sentinel-2 imageries using ensembles of Random Forests

by

Shivam Pande (Y16103079)

**Department of Civil Engineering**

Indian Institute of Technology Kanpur

June 2018

# Land use/land cover classification of fused Sentinel-1 and Sentinel-2 imageries using ensembles of Random Forests

A thesis submitted in partial fulfilment of the requirements

for the degree of

**Master of Technology**

by

Shivam Pande (Y16103079)

to the

**Department of Civil Engineering**

Indian Institute of Technology Kanpur

June 2018

# Certificate

It is certified that the work contained in this thesis entitled "*Land use/land cover classification of fused Sentinel-1 and Sentinel-2 imageries using ensembles of Random Forests*" by Shivam Pande (Y16103079) has been carried out under my supervision and has not been submitted elsewhere for a degree.

June 2018

Dr. Onkar Dikshit
Professor
Department of Civil Engineering
Indian Institute of Technology Kanpur
Kanpur-208016, India

**Statement of Thesis Preparation**

1. Thesis title: Land use/landcover classification of fused Sentinel-1 and Sentinel-2 imageries using ensembles of Random Forests
2. Degree for which the thesis is submitted: M.Tech
3. Thesis Guide was referred to for preparing the thesis.
4. Specifications regarding thesis format have been closely followed.
5. The contents of the thesis have been organized based on the guidelines.
6. The thesis has been prepared without resorting to plagiarism.
7. All sources used have been cited appropriately.
8. The thesis has not been submitted elsewhere for a degree.

Shivam
(Signature of the student)

Name: SHIVAM PANDE

Roll No.: 16103079

Department/IDP: CIVIL ENGINEERING

3

# Abstract

The information obtained from imageries generated from Synthetic Aperture Radar (SAR) and Visible-Near Infrared-Short Wave Infrared (VNIR-SWIR) can be synergistically combined through the technique of image fusion and used for land use/land cover (LULC) classification. One of the objective of this thesis is to study the effect of image fusion of SAR (in the form of texture band) and VNIR-SWIR imageries on LULC classification. Image fusion is performed using Bayesian fusion while random forests, one of the most popular supervised classification techniques has been used for classification. However, random forests have limitations such as inability to perform well with less number of features and stagnation in the accuracy after a certain number of decision trees. In addition, randomization leads to different predictions on the same test set for same classifier with same parameters. Therefore, the other objective of this thesis is to address these limitations by creating ***ensembles of random forests (RFE)*** after introducing random rotations in the training set (based on Forest-RC algorithm). Three approaches are used for rotation: ***principal component analysis (PCA)***, ***sparse random rotation (SRP)*** matrix and ***complete random rotation (CRP)*** matrix. To train and test these classifiers, SAR data from Sentinel-1 and VNIR-SWIR data from Sentinel-2 has been used for the study area of IIT-Kanpur and surrounding region. Five kinds of datasets are created for training: i) SAR, ii) SAR stacked with texture, iii) VNIR-SWIR, iv) VNIR-SWIR stacked with texture and v) VNIR-SWIR fused with texture. Using these datasets, not only the efficacy of the classifiers is studied but the effect of fusion of SAR and VNIR-SWIR data on classification has also been researched on. In addition, the execution speed of Bayesian fusion code is also increased to as high as **3000** times for 700 x 700 image. The ***SRP*** based ***RFE*** performs the best among the ensembles for the first two datasets giving average overall kappa of **61.80%** and **68.18%** respectively while ***CRP*** based ***RFE*** performs the best for the last three datasets with respective average overall kappa values of **95.99%, 96.93%** and **96.30%**. Among the datasets, highest overall kappa of **96.93%** is observed for the fourth dataset. In addition, using texture with SAR bands leads to maximum increment of **10.00%** in overall kappa while maximum increment of about **3.45%** is observed by adding texture to VNIR-SWIR bands.

# CONTENTS

# List of figures

# List of Tables

// Acknowledgement

My gratitude goes to my thesis supervisor Dr. Onkar Dikshit for introducing me to the domain of machine processing of remotely sensed data, guiding me throughout my thesis work and motivating me to read more which greatly benefitted me in my thesis.

I am thankful to Dr Bharat Lohani and Maj. Gen. (Dr) B. Nagarajan who taught me the coursework related to Geoinformatics that led me to develop insights and perspectives that further helped me in the completion of my thesis.

I acknowledge the assistance provided by the Geoinformatics Laboratory staff, without them the completion of thesis would not have been possible.

Finally, I would thank my parents, friends, seniors and juniors who provided me moral support and motivation throughout my thesis work. And of course, the Almighty to bless me with the skills and all the people mentioned above.

# 1 Introduction

In the present age, the use of satellite imagery has been continuously increasing in almost all the fields whether it be agriculture, urban planning, mining, conservation of natural resources, cartography and what not. This is because the satellite images are able to provide a synoptic view of a large area and hence more amount of information can be extracted from them in a very short time. One of the most effective techniques to obtain the information from satellite imageries is through the generation of classified maps. These maps are obtained by assigning several regions of the land/terrain a certain class (such as water, vegetation, urban etc.) based on its properties. However, the accuracy of the classified map would depend on the information contained in the satellite imagery. Therefore, to obtain maximum information from satellite imageries, one must make use of the full potential of the data obtained by satellites i.e. by making use of all the available spectrums in which the remote sensing satellites collect the data. This can be worked out in two steps:

i. Combining the information obtained from those spectrums that complement each other so that the gain of information can be maximized.
ii. Using efficient classification algorithms on the remote sensing data/satellite imagery to get a meaningful interpretation.

The motivation of this thesis finds its roots in the two points mentioned above. Firstly, the advantage of complementing datasets has been harnessed in this study by using Synthetic Aperture Radar (SAR) and Visible, Near Infrared and Short Wave Infrared (VNIR-SWIR) data. By combining both these imageries and utilising their positive characteristics, better information about the land use/land cover (LULC) can be obtained (Gungor and Shan, 2006; Kanakan, 2009).

Secondly, ***random forest***, an excellent classifier (that finds its roots in machine learning), (Hastie et al. 2017) and ***ensemble classification*** techniques have been explored for the classification of combined SAR and VNIR-SWIR datasets.

The combination of the datasets can be carried out either by stacking the bands from different sensors or by performing image fusion. Several existing image fusion algorithms can be applied to fuse these two imageries and get the desired information.

The ideas of image fusion and image classification have been introduced in sections 1.1 and 1.2 of the thesis respectively.

## 1.1 Image fusion

Image fusion as the name suggests is the process of combination of two or more sets of images synergistically (Alparone et al., 2015). Since, a single satellite sensor cannot capture all the images in all the bands of electromagnetic spectrum, it becomes a necessity to combine the beneficial information from several images and use that information in classification. Generally, those image sensors are selected that complement each other so that the missing information in one image can be compensated by another set of images. These images can either be from different sensors (such as SAR and optical images) or they can be from different times of acquisition (Pohl et al., 1998).

## 1.2 Image classification

Image classification refers to the extraction of information from an image/raster by assigning the classes to the image pixels. The image thus generated with all the pixels assigned a particular class is called a classified map. Image classification is divided into two categories:

i. **Unsupervised classification:** In this technique, the pixels in the raster are grouped into clusters based on the numerical and statistical information such as proximity of certain pixels with each other or the pixel values. Though, this kind of classification is fast and simple to perform, the classes created do not always represent the actual data. And since, there can be changes in the spectral properties of the classes, same clusters may not be used again (Hastie et al., 2017).

ii. **Supervised classification:** In this technique, certain pixels in the imagery are associated to a specific class. A number of such classes are assigned to the several pixels called "training set". A classification algorithm is then run which makes use of these training pixels and based on them assigns classes to the unclassified pixels. This classification technique is more accurate than the former one though it requires more computational time and human supervision and expertise (Hastie et al., 2017). This is further divided into two parts:

    a. **Parametric classifiers:** These classifiers assume that the data follows a certain trend or distribution. First, a desired function is selected and then the unknown parameters of the function are learnt with the help of the training data. The examples of such classifiers include maximum likelihood classifier, logistic regression, linear discriminant analysis etc. (Russell and Norvig, 2010).

    b. **Non parametric classifiers:** These classifiers do not assume the distribution of data and hence are more flexible in the classification approach. These methods search for the best fit of the training data in a mapping function and also generalize the unseen or unaccounted data. The examples of these classifiers include k-nearest neighbour, support vector machines, decision trees, random forests etc. (Russell and Norvig, 2010).

Since a single classifier is not able to provide a 100% accuracy, there have been several approaches that try to increase the classification accuracy. One of such approach is ensemble learning in which the predictions from several classifiers are combined to give a single prediction (Zaman and Hirose, 2011). Ensemble learning has been discussed in detail in section 2.1 of the thesis.

This thesis focuses on the use of ***random forest*** classifier and ***classification ensembles*** in classifying the satellite imagery.

## 1.3 Objective

The objective of this study is to evaluate the efficacy of image classification of fused SAR and VNIR-SWIR imagery using ensemble classifiers with ***random forest*** classifier as the base classifier.

### 1.3.1 Scope

In order to achieve the aforementioned objective, following points are included in the scope of work:

i. The VNIR-SWIR imagery is fused with texture band obtained from VH band of SAR imagery using Bayesian fusion (Kanakan, 2009).
ii. The existing code for fusion technique is improved in order to decrease its execution time.
iii. The impact of textural information (obtained from SAR imagery) on classification is studied.
iv. The effect of image fusion on image classification is compared to stacking of texture band to VNIR-SWIR bands.
v. Comparative analysis of classification ensembles with a single RF classifier is performed.
vi. The impact of tuning the parameters on ensembles of RF classifiers is studied.

## 1.4 Data products used

Two kinds of remote sensing data have been used in this work for classification. They are SAR data and VNIR-SWIR data.

The VNIR-SWIR data are obtained by the reflection of the electromagnetic waves (in the range 400 nm-1500 nm) from the target. The reflectivity depends on the composition of the target such as its pigmentation, moisture content of rocks and soils, texture and cellular structure of biomass. However, the VNIR images are subjected to the condition that they can only be acquired in daylight and cloudless sky otherwise the images obtained are not informative (Fairbarn, 2013). VNIR-SWIR imageries have been obtained from Sentinel-2 satellite.

SAR, another kind of valuable remote sensing data, are created when the consecutive pulses of radio waves from the satellite illuminate the target and its echo is recorded on the receiver. Since the echoes are received at different time than the time of transmission of wave, different positions are mapped. The SAR is fixed on a platform, which is in continuous motion. The motion of the platform leads to the creation of a synthetic aperture which is much longer than the actual aperture of the radar. The need of creating such an aperture arises from the fact that the angular resolution of the imagery is directly proportional to the aperture length of the radar (Doerry, 2004). The SAR images are formed by the constructive and destructive interference of the radio waves. The random interference of the echoes leads to the formation of speckles (Gangnon, 1997). The intensity of SAR imagery depends on the polarization of the waves, its frequency, incidence angle as well as the physical characteristics of the surface illuminated. The advantages with SAR images is that they can be captured at night and are not affected by the cloud cover. Besides these, radio waves can also penetrate the surfaces such as vegetation, soil and water which makes them good for measuring thickness. However, in SAR images, the adverse effects of geometric distortions like layover and fore-shortening effects creep in that lead to inaccurate predictions. Also, excessive presence of speckles in the data also hampers its utility (Doring et al, 2013). SAR data have been obtained from Sentinel-1 satellite.

### 1.4.1 Sentinel-1

Sentinel-1 is a satellite mission based on SAR imaging that provides imagery in the C-band. The mission is formed of two categories of polar satellites (Sentinel-1A and Sentinel-1B) orbiting at altitude of 700 km with the period of 6 days. The mission is a part of Copernicus programme under European Space Agency (ESA) (Sentinel-1 User Handbook, 2013). SAR data are obtained in different polarizations:

i. HH: Both transmission and reception are horizontal
ii. VV: Both transmission and reception are horizontal
iii. HV: Transmission is horizontal and reception is vertical
iv. VH: Transmission is vertical and reception is horizontal

Based on them, there are two polarization complexities:

i. **Single polarized:** Anyone of HH, VV HV, VH polarization
ii. **Dual polarized:** Combinations of VV and VH, HH and HV, HH and VV (Polarization in radar systems, 2018)

The SAR data are acquired in four modes:

i. **Stripmap:** Data are acquired in 80 km swath and spatial resolution of 5 m x 5 m.
ii. **Interferometric Wide Swath**: Data are acquired in 250 km swath and spatial resolution of 20 m x 5 m.
iii. **Extra Wide Swath:** Data are acquired 400 km wide swath with 20 m x 40 m resolution.
iv. **Wave**: Data are acquired in vignettes of 20 km x 20 km and spatial resolution of 5 m x 5 m.

Sentinel-1 offers 3 kinds of algorithms and products:

i. **Level 0:** This level consists of unprocessed data with information to support its processing. This is used for scientific purposes.

ii. **Level 1:** This product is most widely used for SAR image processing by the remote sensing community. It is obtained by calibration and processing of level 0 product. This is further divided in two products:

   a. **Single Look Complex (SLC**): This data consists of complex SAR imagery with both intensity and phase information.

   b. **Ground Range Detected (GRD):** This data are multilooked and comprises of intensity information only. These products are available in three resolutions namely, full resolution, high resolution and medium resolution. Table 1 represents specification for 3 kinds of resolutions (Level-1 Ground Range Detected, 2018).

*Table 1: Resolution of SAR products (Level-1 Ground Range Detected, 2018)*

| **Full Resolution** | | | |
|---|---|---|---|
| **Mode** | **Resolution** | **Pixel spacing** | **Number of looks ($N_H$ x $N_V$)** |
| SM | 9 m x 9 m | 4 m x 4 m | 2 x 2 |
| **High Resolution** | | | |
| SM | 23 m x 23 m | 10 m x 10 m | 6 x 6 |
| IW | 20 m x 22 m | 10 m x 10 m | 5 x 1 |
| EW | 50 m x 50 m | 25 m x 25 m | 3 x 1 |
| **Medium Resolution** | | | |
| SM | 84 m x 84 m | 40 m x 40 m | 22 x 22 |
| IW | 88 m x 87 m | 40 m x 40 m | 22 x 5 |
| EW | 93 m x 87 m | 40 m x 40 m | 6 x 2 |
| WV | 52 m x 51 m | 25 m x 25 m | 25 x 25 |

In table 1, $N_H$ and $N_V$ represent the number looks in SAR dataset in horizontal and vertical planes respectively (Level-1 Ground Range Detected, 2018).

iii. **Level 2:** This product is derived from Level 1 product. It consists of geophysical components that is used in wind and wave related applications.

### 1.4.2 Sentinel-2

Just like Sentinel-1, Sentinel-2 is also a satellite mission empowered by ESA under Copernicus programme. However, unlike the former one, it works with the Visible-Near Infrared/Short Wave Infrared (VNIR-SWIR) bands. The mission consists of two groups of satellites: Sentinel-2A and Sentinel-2B. The satellites are high resolution with wide swath and a repeat period of 5 days. The information is obtained in 13 bands out of which 5 are with 10 m resolution, 6 are with 20 m resolution and 3 are with 60 m resolution (Sentinel-2 User Handbook, 2015). There are 5 levels of products in which Sentinel-2 data are worked upon (User Guides - Sentinel-2 MSI - Level-0 Product - Sentinel Online, 2018):

**i.** **Level 0:** This level consists of compressed raw image data and is not available for users. It is used to generate the Level 1 products.

**ii.** **Level 1A:** This product is created by decompressing the Level 0 product. It is also not provided to the users for research.

**iii.** **Level 1B:** This product is derived from Level 1A product. The imagery provided is radiometrically corrected in Top of Atmosphere (TOA). The pixel coordinates in the imagery correspond to the centre of the pixel.

**iv.** **Level 1C:** It consists of 100 x 100 $km^2$ tiles in UTM/WGS84 projection. In this case, pixel coordinates correspond to top left corner of pixel. They are created when the Level 1B images are projected with the help of Digital Elevation Model (DEM).

**v.** **Level 2A:** This product is created when the Level 1C images are corrected for the Bottom of Atmosphere (BOA). It consists of 100 x 100 $km^2$ tiles in UTM/WGS84 projection. Here, the pixel coordinates correspond to top left corner of pixel.

The data are released to the users as Level 1C product by ESA. This can be further processed and converted to Level 2A product as per the need of the user. Salient features of Sentinel-2 bands are presented in Table 2 (MSI Instrument – Sentinel-2 MSI Technical Guide – Sentinel Online, 2018).

*Table 2: Wavelength and resolution for Sentinel-2 bands (User Guides - Sentinel-2 MSI - Level-0 Product - Sentinel Online, 2018)*

| | **S2A** | | **S2B** | | |
|---|---|---|---|---|---|
| **Band Number** | **Central wavelength, $\lambda_c$ (nm)** | **Bandwidth, $\Delta\lambda$ (nm)** | **Central wavelength, $\lambda_c$ (nm)** | **Bandwidth, $\Delta\lambda$ (nm)** | **Spatial resolution (m)** |
| **1** | 443.9 | 27 | 442.3 | 45 | 60 |
| **2** | 496.6 | 98 | 492.1 | 98 | 10 |
| **3** | 560.0 | 45 | 559.0 | 46 | 10 |
| **4** | 664.5 | 38 | 665.0 | 39 | 10 |
| **5** | 703.9 | 19 | 703.8 | 20 | 20 |
| **6** | 740.2 | 18 | 739.1 | 18 | 20 |
| **7** | 782.5 | 28 | 779.7 | 28 | 20 |
| **8** | 835.1 | 145 | 833.0 | 133 | 10 |
| **8a** | 864.8 | 33 | 864.0 | 32 | 20 |
| **9** | 945.0 | 26 | 943.2 | 27 | 60 |
| **10** | 1373.5 | 75 | 1376.9 | 76 | 60 |
| **11** | 1613.7 | 143 | 1610.4 | 141 | 20 |
| **12** | 2202.4 | 242 | 2185.7 | 238 | 20 |

## 1.5 Study area and data used

IIT Kanpur and surrounding area is chosen as the area of study. The map of the area is presented in Figure 1 (Google earth V 7.1.8.3036, 2018).

- Latitude: $26^{O}29'10''$ N to $26^{O}33'00''$ N
- Longitude: $80^{O}13'05''$ E to $80^{O}21'40''$ E

The satellite imagery from Sentinel-1A and Sentinel-2B have been used as the target for classification. The specifications of the datasets have been tabulated in Table 3 (ESA, 2018). Figures 2 and 3 (Copernicus open access hub, 2018) respectively show the VH band of Sentinel-1 imagery and the colour composite of RGB bands of Sentinel-2 imagery for the study area.

*Table 3: Specifications of datasets used*

| S. No. | Sensor | Date of Acquisition | Data Characteristics |
|---|---|---|---|
| 1 | Sentinel-1A | 26-January-2018 | Polarisation: DV (VH and VV)<br>• C-Band<br>• Resolution: 20 m x 22 m<br>• Pixel spacing: 10 m x 10 m |
| 2 | Sentinel-2B | 24-January-2018 | Number of bands: 13<br>• 10 m resolution: 4 (2, 3, 4, 8)<br>• 20 m resolution: 6 (5, 6, 7, 8a, 11, 12)<br>• 60 m resolution: 3 (1, 9, 10)<br><br>Product level: MSIL1C |

## 1.6 Equipment and software used

All the operations were carried out on 64-bit windows 10 based machine equipped with processor Intel(R) Core(TM) i7-4770 CPU @ 3.40 GHz, 8GB RAM.

The image preprocessing, image registration and generation of texture bands have been done in SNAP software. ENVI is used for training data collection while eCognition is used for creating the ground truth image. Matlab R2017a has been used to perform fusion of Sentinel-1A and Sentinel-2B data. The implementation of ensembles of random forest classifiers have been performed in MATLAB as well using Machine Learning Toolbox. The parallel processing utility of MATLAB has been used to increase the speed of classification ensembles.

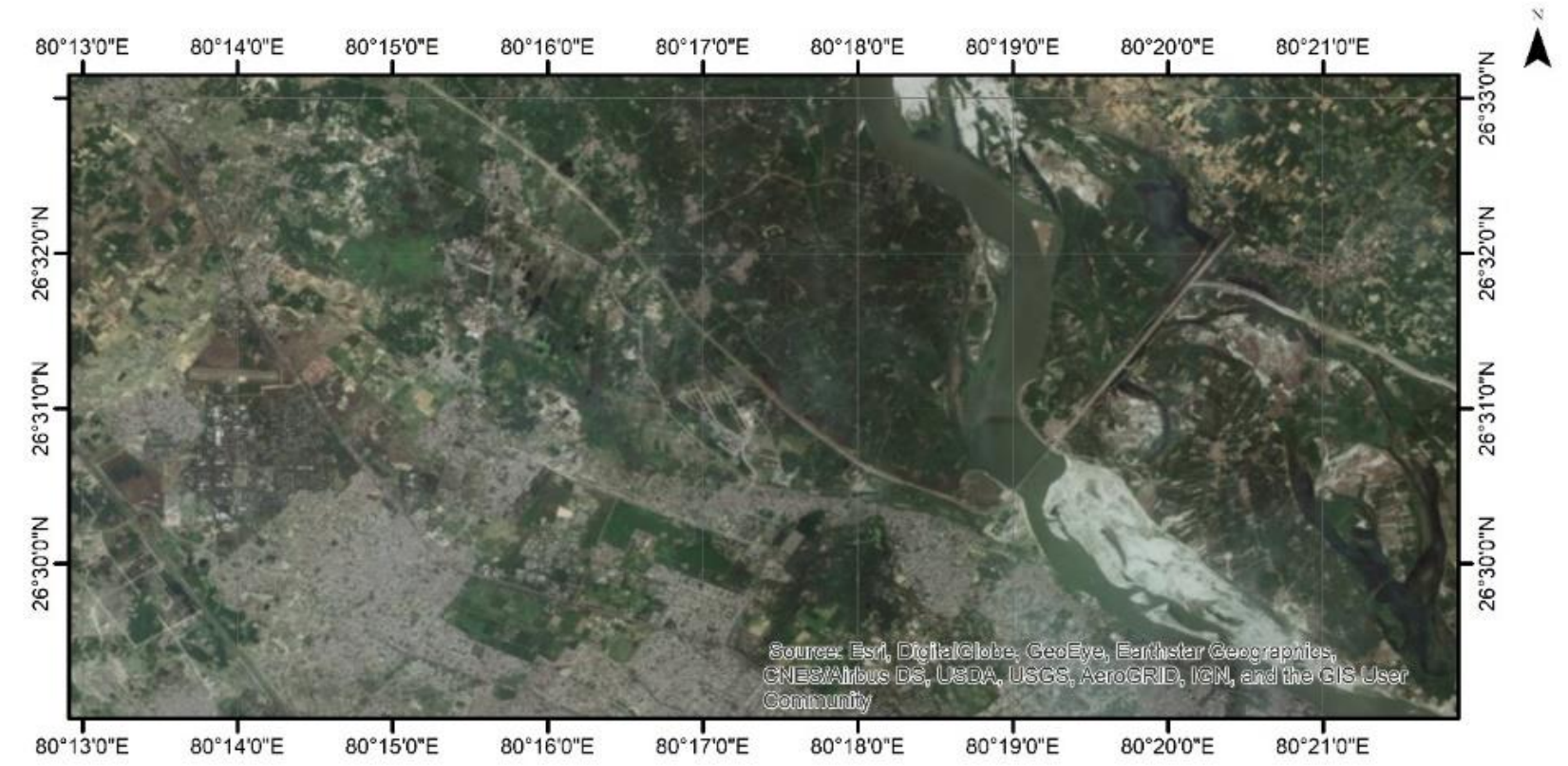

*Figure 1: Study area comprising IIT Kanpur and surroundings. Google earth V 7.1.8.3036. (September 2, 2018). Kalyanpur, Kanpur, India. 26° 29' 10" N- 26° 33' 00" N, 80° 13' 05" E- 80° 21' 40" E, Eye alt 14588 m. DigitalGlobe 2018. http://www.earth.google.com [June 2, 2018].*

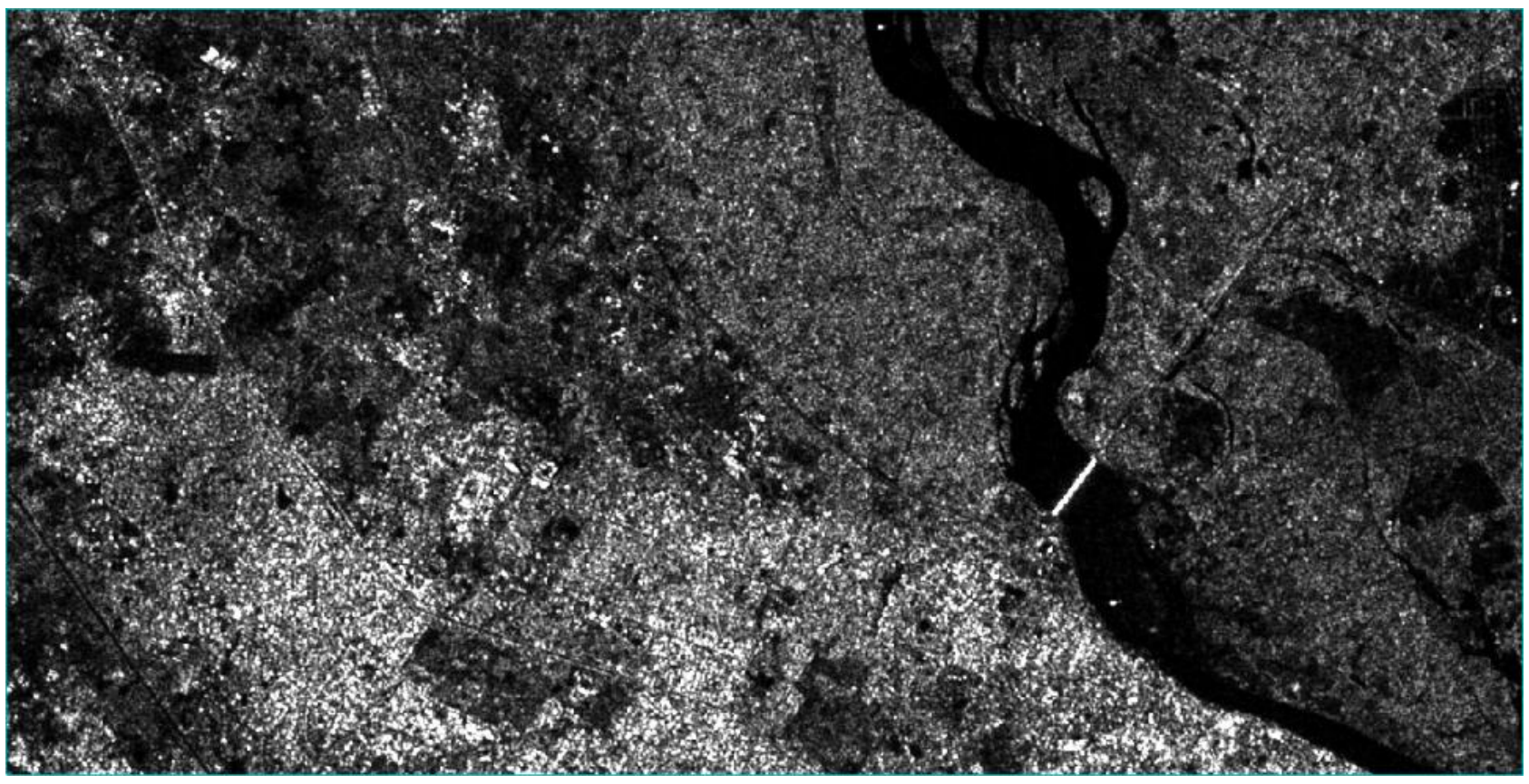

*Figure 2:VH band of S1A_IW_GRDH_1SDV_20180126T124610_20180126T124639_020326_022B7B_3CF2 (Copernicus open access hub, 2018)*

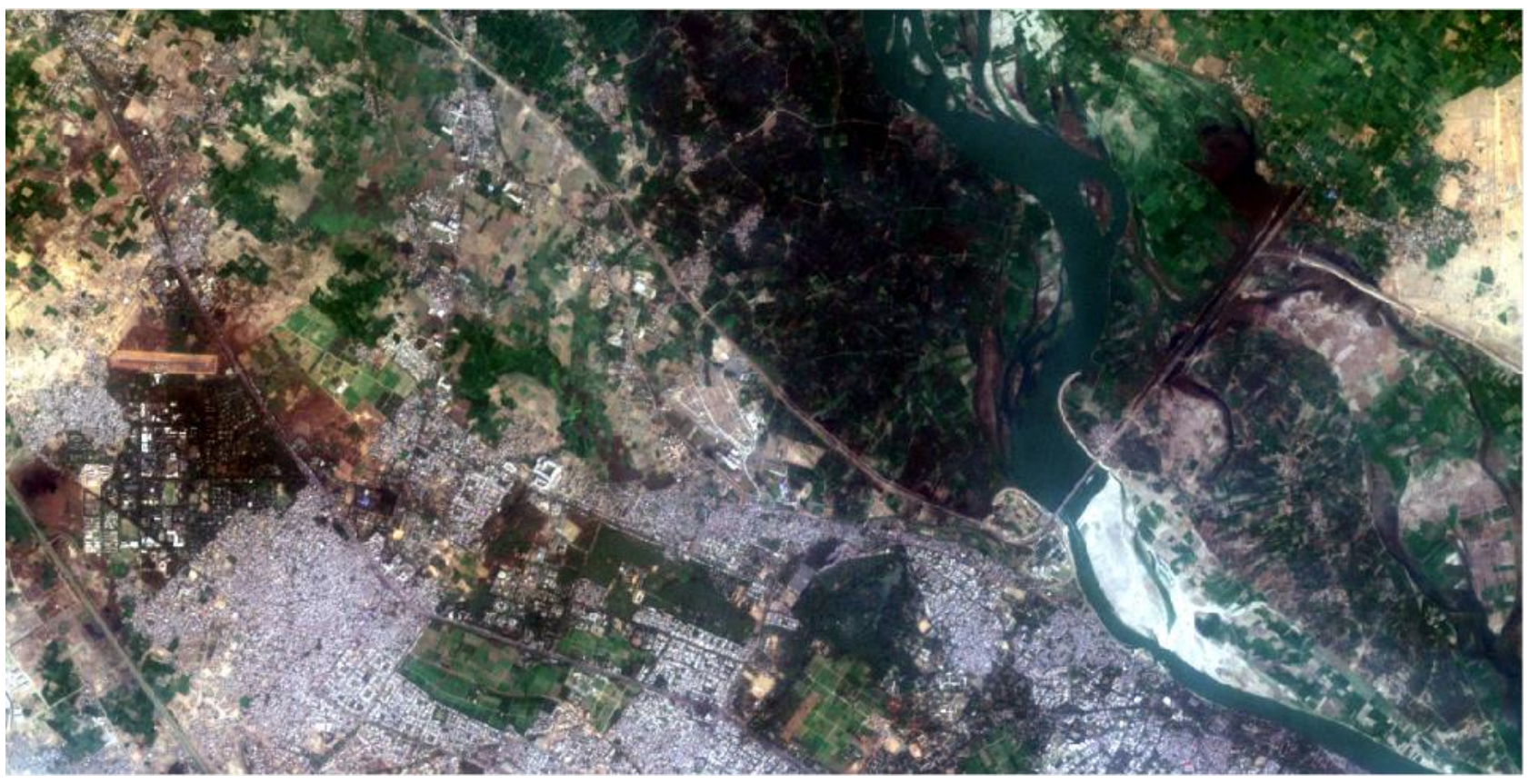

*Figure 3: False colour composite of band 4, band 3 and band 2 of S2B_MSIL2A_ 20180124T051109_N0206_R019_T44RMQ_20180124T175137 (Copernicus open access hub, 2018)*

## 1.7 Structure of thesis

The thesis is divided into five chapters. **Chapter 1** introduces us to the idea of the thesis, familiarizes us to the data sources (Sentinel-1 and 2) and presents the objective and scope of the work. **Chapter 2** is the literature review of the relevant work that has been done in satellite image classification in the recent time and innovative ideas that have been developed since then. **Chapter 3** presents the theoretical and mathematical background of the techniques used in the thesis. **Chapter 4** includes the methodology followed in achieving the research objectives. **Chapter 5** discusses the results of the experiments and studies performed. **Chapter 6** concludes the thesis, provides the inferences based on the results and discusses the future scope of the work done.

# 2 Literature review

Over the course of years, there have been significant work done in the field of image fusion and image classification because of the increasing need to extract more information from remote sensing data and achieve higher classification accuracies. In addition, auxiliary information such as image texture is also being used to improve the classification accuracy. In this chapter, light is shed on recent developments related to image fusion, image classification and texture analysis.

## 2.1 Image fusion

Image fusion, as already mentioned in section 1.1, is the synergistic combination of several images i.e. combining the images in such a way such that the resulting product is more informative than the original products (Alparone et al., 2015). In case, the images are obtained from remote sensing satellites, the technique is termed as satellite image fusion. The satellite images required to be fused could be multi-temporal, multi-spectral or multi-spatial (Chang and Bai, 2015). The fusion methods can be classified into three levels: Pixel level fusion, feature level fusion and decision level fusion (Pohl et al., 1998).

### 2.1.1 Pixel level fusion

It is considered as the lowest level of fusion. After carrying out the image registration, the images are fused at the pixel-by-pixel level. This means that the digital number in a specific pixel of the image is not affected by the digital number in any other pixel of the image. The example of such kind of fusion could be averaging of two images or generation of Normalised Burn Ratio (NBR) band. NBR is given as:

$$NBR = \frac{NIR - SWIR}{NIR + SWIR} \tag{2.1}$$

Here, *NIR* is the reflectance in Near Infrared band and *SWIR* is reflectance in Short Wave Infrared band (Normalized Burn Ratio, 2015). Here, it is observed that the fusion is simply an arithmetic operation between two kind of bands (Chang and Bai, 2017). The examples of this kind of fusion include fusion using ***principal component analysis, intensity hue saturation, Brovey transform, multiscale transform*** etc. (Jagalingam and Hegde, 2014).

### 2.1.2 Feature level fusion

At this level of fusion, the distinctive features from the imageries are identified, segmented and extracted before the fusion is carried out. These features are then blended/fused together instead of the pixels. The features can either be radiometric (for e.g. intensities) or geometric (height, size etc.). The new feature space obtained after fusing the original features tends to be more informative than the original feature space. It must be kept in mind that features must be extracted carefully to get desirable fused products (Chang et al., 2016).

### 2.1.3 Decision level fusion

It is the kind of image fusion level where the features from several imageries are combined using external decision rules to get a meaningful common interpretation of the features in the fused image (Pohl et al., 1998). This quality of the features greatly influences the fusion results because the decisions based on it (Chang et al, 2014). The decision rules can either be hard, soft or both. Hard decision methods include weighted sum scores, Boolean methods and M-of-N method. Soft decision methods include Bayesian, Dempster-Shafer and fuzzy approach (Chang and Bai, 2017).

## 2.2 Image classification

This study focusses on the use of ensemble learning techniques such as random forests for image classification. The subsequent sections throw light on the several kinds of ensemble learning techniques and random forest classifier and explain its limiting factors and research gaps.

### 2.2.1 Ensemble learning techniques

Ensemble learning is an approach in machine learning in which several learners are trained on the same dataset. These ensemble learners develop an individual hypothesis for each learner, which are then combined to generate a single result (Zhou, n.d.). The individual learners are referred as base/weak learners. Generally, a single type of base learner is used in the ensemble to maintain its homogeneity. The idea of using a set of learners to improve the prediction accuracy has been in practice for a long time. However, the formal work on ensemble learning came into picture from the works of Hansen and Salamon (1990), where they created an ensemble using neural networks as base classifiers and Schapire (1990), where he boosted those weak learners which could perform just better than guessing.

Ensemble learners are generally discussed at four levels (Hawadiya, 2015). These four levels are discussed in subsequent sections.

#### *2.2.1.1 Feature level*

This level involves training of base classifiers using various subsets of features from the same training set. This kind of ensemble technique helps to tackle the problem of high dimensionality in the dataset. In case of high dimensional data, if the number of training samples are not sufficient, then training the model becomes an impossible task because of the formation of low rank matrices. So, to deal with this problem, the feature set is broken down into smaller number of subsets and each base classifier is then trained on this subset of features. (Benediktsson, 2008; Su, 2014). The results of all such classifiers are then combined to get a s single prediction.

#### *2.2.1.2 Data level*

At this level, the data for the same area from several sources is used to train each base learner. The data could be multi temporal data or from different sensors. This type of learning is generally performed in case of climate studies or change detection in the terrain over a period of time. Waske et al. (2009) used Random Forest ensemble learning on multi-temporal SAR imagery and obtained sufficient good results.

#### *2.2.1.3 Classifier level*

At this level, several kinds of base leaners are used to train the datasets so that the strong aspect of each classifier can be incorporated in the final prediction to increase the accuracy of the resulting classifier.

Hongfen et al. (2008) used a classification ensemble consisting of minimum distance classifier, Mahalanobis classifier, maximum likelihood classifier and Support Vector Machine (SVM) and used it to classify the Quick Bird imagery. The results showed an increment of 3.95% in accuracy in comparison to single SVM classifier.

#### *2.2.1.4 Combination level*

At this level, the predictions from several base learners are combined using various voting mechanisms such as majority voting or weighted majority voting algorithm. The weights in the latter case are applied by taking into account the overall accuracy of the classifier. The boosting method of ensemble learning that is discussed section 2.2.2 is an example of such level.

### 2.2.2 Types of ensemble learning techniques

#### *2.2.2.1 Bagging*

Bagging stands for Bootstrap Aggregation. It is a prediction model that combines several predictors and takes their aggregated result as the prediction. In case of classification, plurality/majority voting technique is used to get the most probable prediction for the class.

Bootstrapping refers to a technique of randomly selecting samples (with replacement) from the learning sets. Each of the training sample is then is used to train the individual learner. The training process occurs parallel for all the trees. This technique is observed to have increased the prediction accuracy by introducing significant disturbance in the dataset. The specific trait of bagging is that in it, every element is equally probable to appear in the new training set (Breiman, 1984).

#### *2.2.2.2 Boosting*

Boosting is another ensemble mechanism. It is different from bagging in several aspects. First of all, the classifiers are trained sequentially. This is because boosting tries to provide weights to the observations on the basis of success of the previous classifier and hence some of them occur more often in the training samples. After each training step, the weights are redistributed. The weights for the misclassified data are increased to take into account the difficult cases so that the subsequent learners could focus on them while training. Adaboost is one of the most famous algorithms of boosting and has shown promising results (Freund and Schapire, 1999).

#### *2.2.2.3 Stacking*

Also called 'stacked generalization', it is a technique that is used to combine different kinds of classifiers i.e. it works at the classifier level of ensemble learning. Stacking generally works at two levels. Level-0 (called base learner) consists of a set of weak learners while level-1 (called stacking model learner) consists of a set of stronger learners. The dataset is first fed to the learners at level-0. The prediction results collected from those learners are then used to collect a new dataset in which each of the prediction is related to the real value it was expected to predict. Then, this dataset is provided the level-1 model and the results are called as final results (Syarif et al., 2012).

The random forest classifier used in this work is a feature level classifier that follows the bagging technique of creating ensemble (Breiman, 2001).

### 2.2.3 Voting techniques in ensemble learning

Voting techniques work on the labels of the predicted classes. For the number of classifiers *T*, the prediction of class by each classifier for each pixel is assigned a certain probability value. These probability values depend on the weights of the classifiers. Finally, all the probabilities are combined and the class label with maximum probability is assigned to the pixel under consideration (Poliker, 2011). These techniques are divided as:

#### *2.2.3.1 Majority voting*

In this case, each classifier is given equal weightage. Under the assumption that the classifiers are independent, majority voting always improves the performance of the individual classifiers for a large number of classifiers. Let there be a two class problem for an ensemble with *T* classifiers then the ensemble decision will be accurate if $\frac{T}{2}+1$ classifiers choose the right class. Now, if each classifier has probability $p$ for choosing the correct class, then the probability of ensemble of choosing the $k$ correct classifiers ( $P_{ens}$ ) will be given as:

$$P_{ens} = \sum_{k=\frac{T}{2}+1}^{T} {}^{T}C_k \, p^k (1-p)^{T-k} \tag{2.2}$$

For $T \to \infty$, if $p > 0.5$

Then,

$$P_{ens} \to 1 \tag{2.3}$$

And if $p < 0.5$

$$P_{ens} \to 0 \tag{2.4}$$

It is to be noted that $p > 0.5$ is necessary as well as sufficient for a two class problem. However, it is not necessary for a multiclass problem though it would still be sufficient (Poliker, 2011).

#### *2.2.3.2 Weighted majority voting*

In this method, each classifier is given a vote. The most common method of voting is based on overall accuracy of the individual classifier. The classifier with maximum overall kappa is given the highest weight and so on. Based on these weights the probabilities are decided and then voting is performed (Poliker, 2011).

### 2.2.4 Random forests

***Random forest (RF)*** classifier is an ensemble classification approach developed by Leo Breiman (2001). This classifier combines several decision trees using ***bagging*** approach of ensemble learning. Each decision is however trained only on a subset of features instead of the full feature set. The results obtained from each decision tree are finally combined using majority voting to get the final prediction. The explanation of random forest requires the understanding of decision trees, which is used as a base classifier in it. Decision trees are discussed in detail in section 3.4.1.1 of this thesis.

The decision trees used in bagging are independent and identically distributed (i.i.d.). This means that the bias for a single tree is same as that of the bagged trees and hence variance reduction is the only way for improvement. The reduction in the variance is achieved by selecting the subset of variables/features randomly while growing trees. So, if the total number of features are $p$ then the subset $m$ would be such that $m \le p$. The default value of $m$ is kept to be $\sqrt{p}$ .

#### 2.2.4.1 Stagnation and overfitting in random forests

When there are too many variables but only limited number of relevant variables, then random forests do not perform well because the probability of selecting relevant variable at each split of decision tree decreases. In such a case, the decision trees are trained on irrelevant/noisy data and hence leads to underperforming of random forests (Hastie et al., 2017).

From the perspective of increment in the number of trees, there has been no record of overfitting. However, after reaching a certain value of trees, the further increase in the accuracy stagnates to a constant value. The number of such trees have been limited to 64-128 (Oshiro et al., 2012).

### 2.2.5 Forest-RC and oblique random forests

In the original paper on Random Forests (Breiman, 2001), the method called Forest-RC was discussed. According to it, in case of less number of features, taking a higher portion from them for splitting would lead to high correlation. To avoid this, more features can be generated by randomly combining the features in linear fashion. After this combination, the new feature set can be used to find the most optimum splits. Since, feature space is being rotated in this method, the forests created are also called *oblique random forests* (in contrast, the random forest mentioned in section 2.2.4 are sometimes called orthogonal random forests) (Menze et al., 2011). This method is further discussed in Chapter 3.

## 2.3 Texture analysis

Though there is no formal definition of texture, it can be crudely defined as the consistency of a surface or the variation on the surface with respect to colour, shapes, density and sizes (Gimel'farb, 2006). With the point of view of an imagery, texture can be defined as the variation in the digital numbers of the image (Haralick, 1973). Haralick defined 14 parameters that could be considered as the measure of defining texture namely, angular second moment, contrast, correlation, sum of squares, inverse difference moment, sum average, sum variance, sum entropy, entropy, difference variance, difference entropy, two information measures of correlation and maximal correlation coefficient. These texture parameters can be computed using Grey Level Co-occurrence Matrix (GLCM). It is one of the earliest techniques developed and being used to calculate the texture parameters for an image (Sebastian et al., 2012). Analogous to *"pixel"* in an image, the smallest unit of a texture band is called *"texel"*, which is the shortened version for texture element (Tamakuwala, 2013). Kumar and Dikshit (2014) showed that incorporating textural information along with the spectral information in hyperspectral imagery provides statistically better classification accuracies than using only spectral features. The procedure of creating GLCM and generating texture bands has been discussed in section 3.4.

## 2.4 Summary and research gaps

After carrying out the literature review, following points are considered to work upon in this study:

a. The random forests have shown to exhibit stagnation in accuracy after a number of decision trees, so techniques are required to be employed to increase their accuracy.
b. It is also observed that random forests are giving different results on the test set even after being trained on the same training set. This happens due to randomization involved in training and therefore it is decided to create the ensembles of random forests to counteract this issue.
c. The Forest-RC algorithm mentioned in section 2.2.5 has not been explored as frequently as orthogonal random forests and therefore this algorithm is being incorporated in the generation of ensembles.
d. Two kinds of data products (SAR and VNIR-SWIR) have been used to study the effect of ensembles on the accuracies obtained for both of them.

e. Since, texture features have been shown to improve classification accuracy when added to spectral features, therefore they are incorporated as well to study how the classifier behaves with the involvement of texture in the imagery.

Based on these research gaps, the objectives are defined in section 1.3.

# 3 Theoretical and mathematical background

This chapter explains in detail the theory and mathematics of the various procedures involved in completion of this research work. Firstly, the image preprocessing of Sentinel-1 (SAR) and Sentinel-2 (VNIR-SWIR) imagery is discussed. After this, the image co registration of Sentinel-1 and Sentinel-2 imagery is discussed. This is followed by the Bayesian fusion technique of image fusion. Subsequently, the image classification techniques are discussed. Finally, the accuracy analysis is discussed using kappa coefficients and error matrices.

## 3.1 Image preprocessing

Satellite images directly obtained from sensors are subjected to several distortions. These distortions may occur in the image due to several factors such as the refraction of the electromagnetic rays in layers of atmosphere, presence of clouds, constructive and destructive interference of EM waves in case of radio waves and several other factors. Hence, before carrying out any image related operation, it becomes important to correct the images for such distortions. Sometimes it is also argued that image preprocessing is undesirable since it leads to change in the image characteristics. However, if it is carried out intelligently, image preprocessing can greatly benefit the further operations on images (Krig, 2014).

### 3.1.1 Preprocessing of Sentinel-1 data

The preprocessing of Sentinel-1 data involves the following steps:

#### 3.1.1.1 *Calibration*

SAR calibration provides an imagery in which a relationship can be established between the radar backscatter and the digital numbers of the image. Firstly, the digital numbers are calculated for the backscattered complex signal as:

$$DN = \sqrt{I^2 + Q^2} \tag{3.1}$$

Here, I is the real part while Q is the imaginary part for the backscattered complex signal (Radiometric calibration of Terrasar-X data beta naught and sigma naught coefficient calculation, 2014).

Secondly, a new scaling must be applied after reverting the application scaling of the processor. To accomplish this task, four Look-Up Tables (LUTs) are provided in Level-1 products either to produce or to return to the Digital Number (DN). A range-dependent gain that includes absolute calibration constant is applied by the LUTs. A constant offset is also applied for the GRD products.

All the information needed to convert digital numbers to radiometrically calibrated backscatter, is found in the product. The $\sigma_o$ (backscatter coefficient) and $\beta_o$ (radar brightness) are derived from image intensity values using a calibration vector included in the product.

The equation mentioned below applies the radiometric calibration:

$$value(i) = \frac{|DN_i|^2}{A_i^2} \tag{3.2}$$

$value(i)$ : the calibrated value and is either $\sigma_o$ or $\beta_o$ (depending on the calibration LUT)

$A_i$ : gain value/ calibration constant

$DN_i$ : digital number

In case any pixel falls between the LUTs, bi-linear interpolation is used (El-Darymli et al., 2014).

#### *3.1.1.2 Terrain correction*

The methodology of Range Doppler Terrain Correction (RDTC) is used to correct the terrain for distortions due to tilt of the satellite sensor. These corrections minimises these distortions so that the image is close to the actual scene. Range Doppler orthorectification algorithm (Small and Schubert, 2008) is followed to carry out this correction. This algorithm makes use of available state vectors, parameters to convert slant range product to ground range product, radar timing annotations and DEM data to geocode SAR imageries from a single 2D raster. The schematic diagram of terrain correction is shown in figure 4 (User guide - SNAP, 2018).

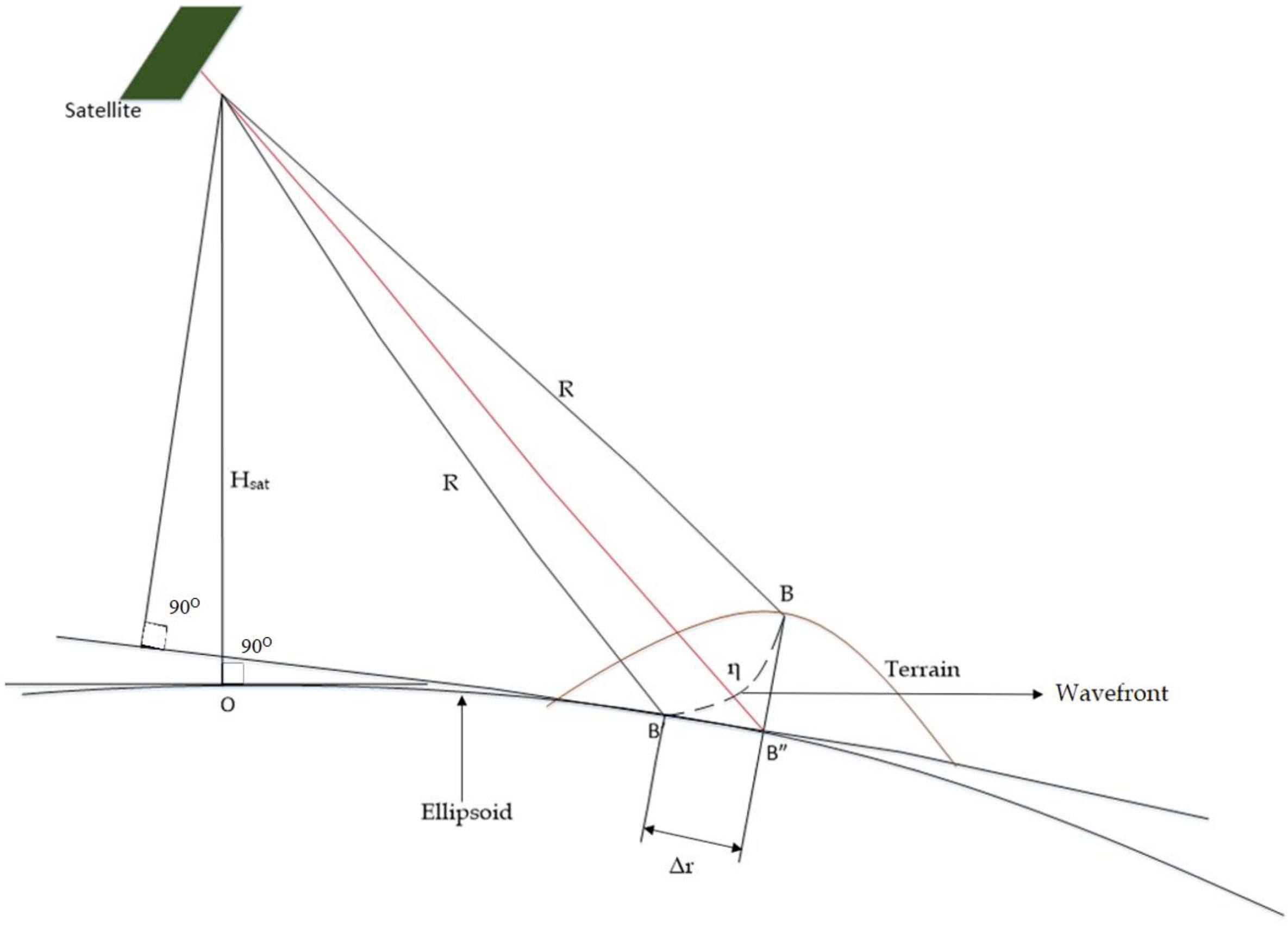

*Figure 4: Terrain Correction (User guide - SNAP software, 2018)*

Here,

$H_{sat}$ : height of satellite from the ellipsoid

$B$ : observed point on the surface

$B'$ : point where $B$ is imaged

$B''$ : point where $B$ should have been imaged

$\Delta r$ : distance between $B'$ and $B''$

### 3.1.2 Preprocessing of Sentinel-2 data

#### *3.1.2.1 Atmospheric correction*

When the solar radiation passes through the atmosphere, it scatters because of interaction with the particles present in the several layers of the atmosphere such as dust, water vapour etc. In addition, when the radiation interacts with the earth's surface, it is diffused in several directions. The satellite sensors not only receive the direct radiation but also the indirect radiation due to scattering. To take into account these effects, we need to apply atmospheric correction to satellite imagery (Hadjimitsis et al., 2010).

## 3.2 Creation of new bands

Since in Sentinel-1 imagery, the number of bands are limited to 2 (sigmaVH and sigmaVV bands), four new bands are created by combining the original bands to increase the number of features in classification (Abdikan et al., 2016):

- Average : $\frac{VV+VH}{2}$
- Difference : $VV-VH$
- Ratio : $\frac{VV}{VH}$
- DEM band (generated after terrain correction)

Similarly, for Sentinel-2 imagery, a new RDVI (Renormalized Difference Vegetation Index) band is created using Red and NIR bands to account for vegetation present in the imagery (Roujean and Breon, 1995).

$$RDVI = \frac{NIR - Red}{\sqrt{NIR + Red}} \tag{3.3}$$

## 3.3 Image registration

Image registration is a technique by which two (or more) images are superimposed over each other. The need for registration arises when the images corresponding to the same scene are captured from different kind of sensors, from different angles, different times etc. but their information is required to be used simultaneously such as in case of image fusion. In this process, one image can be considered as master image and the other image(s) (also called slave images) are then registered on it (Zitová et al., 2003).

## 3.4 Generating texture features

GLCM (discussed in section 2.3) has been used to generate texture band. GLCM matrix is a square matrix with size $N_g$ x $N_g$ where, $N_g$ is the number of grey levels in the satellite imagery. To generate element $[i, j]$ of the GLCM matrix, first define a direction along which the adjacency of the neighbouring digital numbers is required to be observed. The directions can be $0^o$, $90^o$, $45^o$ or $135^o$. Now, count the number of times the pixel that has a value $N_i$ and is next to a pixel whose value is $N_j$ and store these value in a matrix of size $N_g$ x $N_g$.

$$Frequency\ Matrix = \begin{bmatrix} P(1,1) & P(1,2) & \cdots & P(1,N_g) \\ P(2,1) & P(2,2) & \cdots & P(2,N_g) \\ \vdots & \vdots & \ddots & \vdots \\ P(N_g,1) & P(N_g,2) & \cdots & P(N_g,N_g) \end{bmatrix} \tag{3.4}$$

$P(i, j)$ : Relative frequency with which two neighbouring grey levels $N_i$ and $N_j$ are separated in the image.

To normalize the frequency matrix, divide it by the summation of all the elements present in it.

$$GLCM = \frac{1}{R}\begin{bmatrix} P(1,1) & P(1,2) & \cdots & P(1,N_g) \\ P(2,1) & P(2,2) & \cdots & P(2,N_g) \\ \vdots & \vdots & \ddots & \vdots \\ P(N_g,1) & P(N_g,2) & \cdots & P(N_g,N_g) \end{bmatrix}$$

$$= \begin{bmatrix} p(1,1) & p(1,2) & \cdots & p(1,N_g) \\ p(2,1) & p(2,2) & \cdots & p(2,N_g) \\ \vdots & \vdots & \ddots & \vdots \\ p(N_g,1) & p(N_g,2) & \cdots & p(N_g,N_g) \end{bmatrix} \quad (3.5)$$

where,

R : summation of all the elements in the frequency matrix

$p(i, j)$ : normalised joint probability that how often pixel value $N_i$ is adjacent to pixel value $N_j$ along the chosen direction (Haralick, 1979)

The Haralick texture parameters are derived from mathematical manipulations among the elements of GLCM matrix. For e.g. the expression for homogeneity band (also called inverse difference moment) (Linda and George, 2000) that is used in this study is given as:

$$Homogeneity = \frac{\sum_{i=1}^{N_g}\sum_{j=1}^{N_g} p(i, j)}{1+(i-j)^2} \quad (3.6)$$

*Homogeneity* has been used in this study because it is used with classifiers that are based on statistics (Gotileb and Kreyszi, 1990).

To generate a texture band for a certain Haralick parameter, say *homogeneity*, move a window of size $q$ x $q$ (where $q$ is an odd number less than the smallest dimension of the image) over the image and for each instance, generate a GLCM, calculate the required parameter and assign it to the pixel over which the window is centred (Geoinformatics Tutorial: Creating a Texture Image with a GLCM Co-Occurence Matrix using Scikit-Image and Python, 2016).

## 3.5 Image fusion

As already mentioned in Chapter 1, it is a process that synergistically combines the images from two or more sources to cumulate their information. The image fusion technique that is used in this study is Bayesian fusion (Kanakan, 2009). It is discussed in detail below:

### 3.5.1 Bayesian fusion

The idea behind this technique is that when the variables of interest, say $Z$, cannot be directly observed then they can be studied through observable variables, say $Y$, by means of an error model given as:

$$Y = g(Z) + E \tag{3.7}$$

Here,

$g(Z)$ is a functional vector

$E$ is a vector of errors stochastically independent of $Z$

For set of observations $Y$, i.e. $y = (y_0, ..., y_n)^T$ are available , a conditional probability density function is formulated:

$$f(z \mid y) = f(z_0, ..., z_n \mid z_0 + e_0, ..., z_n + e_n) \tag{3.8}$$

and $e = (e_0, ..., e_n)$ is associated error

Using Bayes' theorem (Hastie et al., 2017):

$$f(z_i \mid y_i) = \frac{f(y_i \mid z_i) f(z_i)}{\sum_{i=1}^{N} f(y_i \mid z_i) f(z_i)} = \frac{1}{A} f(y_i \mid z_i) f(z_i) \tag{3.9}$$

Here, $A$ is normalization constant.

Conditional cumulative distribution for $F(y|z)$ is given as:

$$F(y|z) = P(Y < y|z) = P(g(Z) + E < y|z)$$

$$= P(E < y - g(Z)) = F_E(y - g(z)) \quad (3.10)$$

This gives,

$$f(y|z) = \frac{\partial}{\partial y} F(y|z) = f_E(y - g(z)) \quad (3.11)$$

Using 3.10 in equation 3.8 and removing $A$

$$f(z|y) \propto f_Z(z) f_E(y - g(z)) \quad (3.12)$$

Here, $f(.)$ is the *a priori* pdf of $Z$ and $f_E(.)$ is *a priori* pdf of errors $E$.

The formulation for Bayesian Fusion is given as:

Let

$Z = (Z_1, ..., Z_n)^T$ be fused SAR and VNIR-SWIR product

$Y = ((Y_M)^T, y_S)^T$ be the set of observed variables

$Y_M = (Y_{M,1}, ..., Y_{M,n})^T$ be digital numbers in VNIR-SWIR bands

$y_S$ the corresponding digital number in SAR band

$E = ((E_M)^T, E_S)^T$ be the vector of errors assumed to be zero-mean so that

$Y_M = g_M(Z) + E_M$ and $Y_S = g_S(Z) + E_S$

Since, $E_M$ and $E_S$ are independent

$$f(z \mid y_M, y_S) \propto f_Z(z).f_{E_M}(y_M - g_M(z)).f_{E_S}(y_S - g_S(z)) \quad (3.13)$$

To account for the weights, the equation can be modified as:

$$f(z \mid y_M, y_S) \propto f_Z(z).f_{E_M}(y_M - g_M(z))^{2(1-w)}.f_{E_S}(y_S - g_S(z))^{2w} \quad (3.14)$$

Here, $w$ is the weight parameter to increase the SAR content in the fused product at the cost of VNIR-SWIR content. Its value lies between $0$ and $1$ where $0$ discards SAR and 1 discards VNIR-SWIR. The value of $w = 0.5$ leads to back to equation 3.7.

Following assumptions are made in implementation of Bayesian fusion:

i. $Y_M$ is directly related to $Z$ since they are directly observed bands.

ii. $E_M$ is multivariate Gaussian, i.e. $E_M \sim N(0, \sum_M)$ where $\sum_M$ is estimated covariance matrix calculated from VNIR-SWIR image.

iii. $E_S$ is Gaussian with $E_S \sim N(0, \sum_S)$.

iv. Functional $g_S(z)$ is expressed as a linear regression model. It links VNIR-SWIR pixels to SAR pixels. So, $g_S(z) = \alpha + z^T \beta$. Here, $\alpha$ and $\beta = (\beta_1, ..., \beta_n)^T$ are regression parameters. RMS error is used to calculate variance of $E_S$, $\sigma_S^2$.

v. For *a priori* pdf of $z$, there is a non-informative *prior* pdf with $f_Z(z) \propto 1$.

Using these assumptions, the values of $\mu_w$ and $\sum_w$ are obtained as:

$$\mu_w = \sum\nolimits_w \left( 2(1-w) \sum\nolimits_M^{-1} y_M + 2w \frac{1}{\sigma_s^2}(y_s - \alpha)\beta \right) \quad (3.15)$$

$$\sum\nolimits_w^{-1} = 2(1-w) \sum\nolimits_M^{-1} + 2w \frac{1}{\sigma_s^2} \beta\beta^T \quad (3.16)$$

The value of $\mu_w$ gives the pixel value for the fused image. It is clearly visible that it is a function of $w$ and control the proportions of SAR and VNIR-SWIR images respectively.

## 3.6 Image classification

The random forest classifier and classification ensembles that use random forest as base classifier have been used in this study to perform satellite image classification. They are discussed as follows:

### 3.6.1 Random forest classifier

Random forests have already been introduced in section 2.3 of the thesis. As already mentioned, these forests use decision trees as their base classifier for ensemble formation and then use majority voting scheme to determine the single output. The decision trees have been discussed below:

#### *3.6.1.1 Decision trees*

Decision trees are the models that are used to predict the value of a discrete valued function. The use of decision trees can be dated back to 1966 (Hunt et al. 1966). They are graphically shown to be hierarchical in structure. The first node (at the top) is called *root node*. It does not have any incoming edge. Then come the *internal nodes*. These nodes have the outgoing and incoming edges. The internal represent a condition on the feature and edges/arrows show the result of the consequence. The nodes that form the end of the decision tree are called *leaf nodes*. The leaf nodes contain the classification result (class label) for the required variable.

Classification and regression trees (CART) is a type of methodology that can be used to fit the decision trees. (Breiman et al., 1984). The thesis makes use of the classification trees helps in labelling the image data. The trees are created using the following steps:

i. Let there be *n* number of observations and *m* number of samples. Select (with replacement), a sample from training data.
ii. From *m* features, select a subset of features that give the best fit.
iii. Create the largest possible trees.

Figures 5 (a) and 5 (b) are used to explain the working of a decision tree.

The best fitting features can be selected by either using the Gini index or by calculating the *Information Gain* for the features.

Gini Index is a measure of homogeneity or heterogeneity of the features. It is directly proportional to the class heterogeneity. For a successful split, the Gini index for child node must be less than that of the parent node. When Gini index is calculated as zero, it means the splitting is finished and we are at the end of the tree (Raileanu, 2004).

$$Gini\ Index = 1 - \sum_{i=1}^{K} p_k^2 \qquad (3.17)$$

In equation 3.11, $p_k$ represents the proportion of instance that belong to a class.

Information Gain is the measure of amount of information a feature gives about a class. In other words, it measures the reduction in the entropy of the system. The information gain for the system is given as:

$$I(A) = H\left(\frac{p}{p+n}, \frac{n}{p+n}\right) - EH(A) \qquad (3.18)$$

where, $EH(A)$ is Expected Entropy and is given as:

$$EH(A) = \sum_{i=1}^{K} \frac{p_i + n_i}{p+n} H\left(\frac{p_i}{p_i+n_i}, \frac{n_i}{p_i+n_i}\right) \qquad (3.19)$$

Here,

$p$ and $n$ are number of positive and negative samples respectively.

$p_i$ and $n_i$ are $i^{th}$ samples from positive and negative samples (Russell and Norvig, 2010).

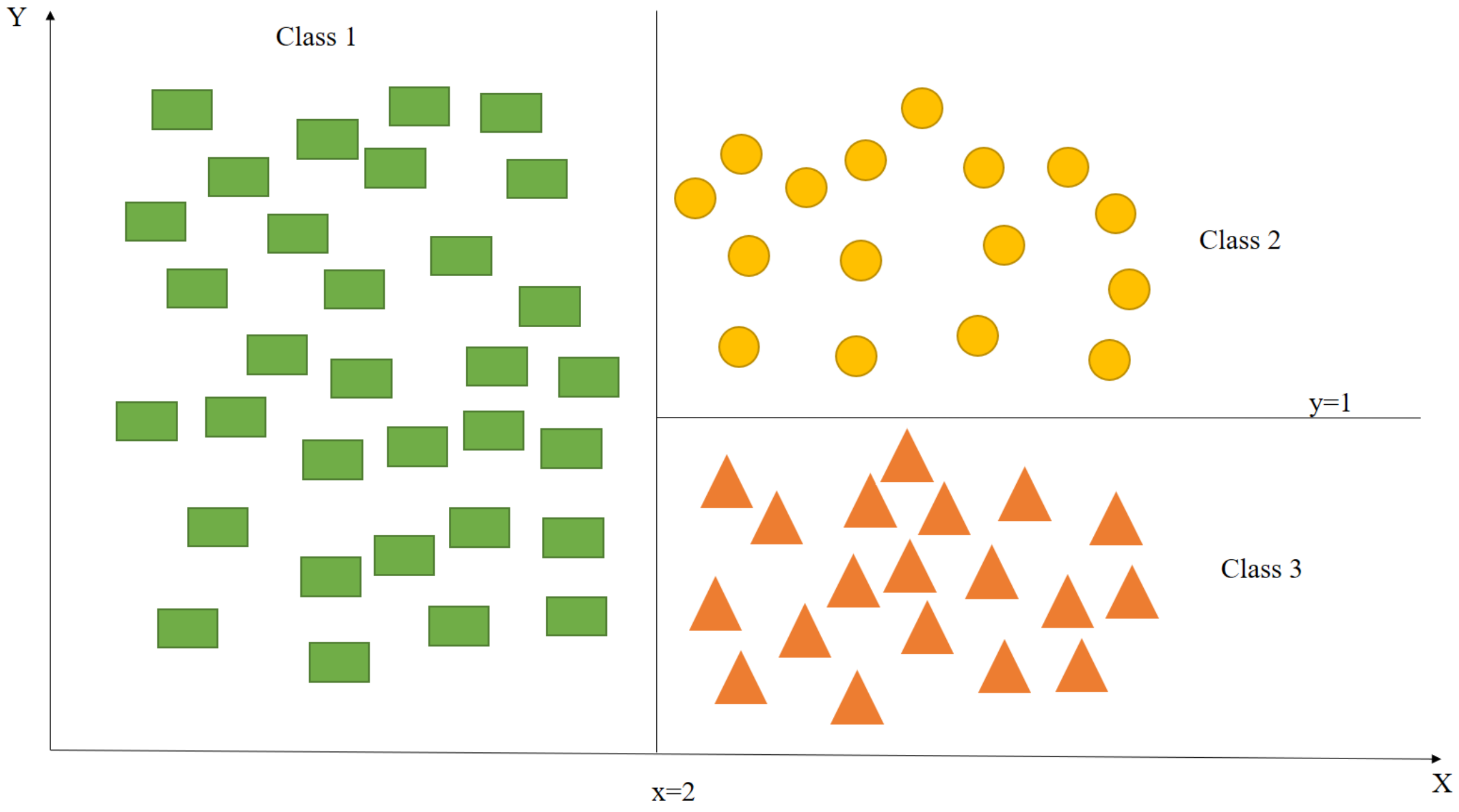

*Figure 5(a): Explanatory dataset for decision tree*

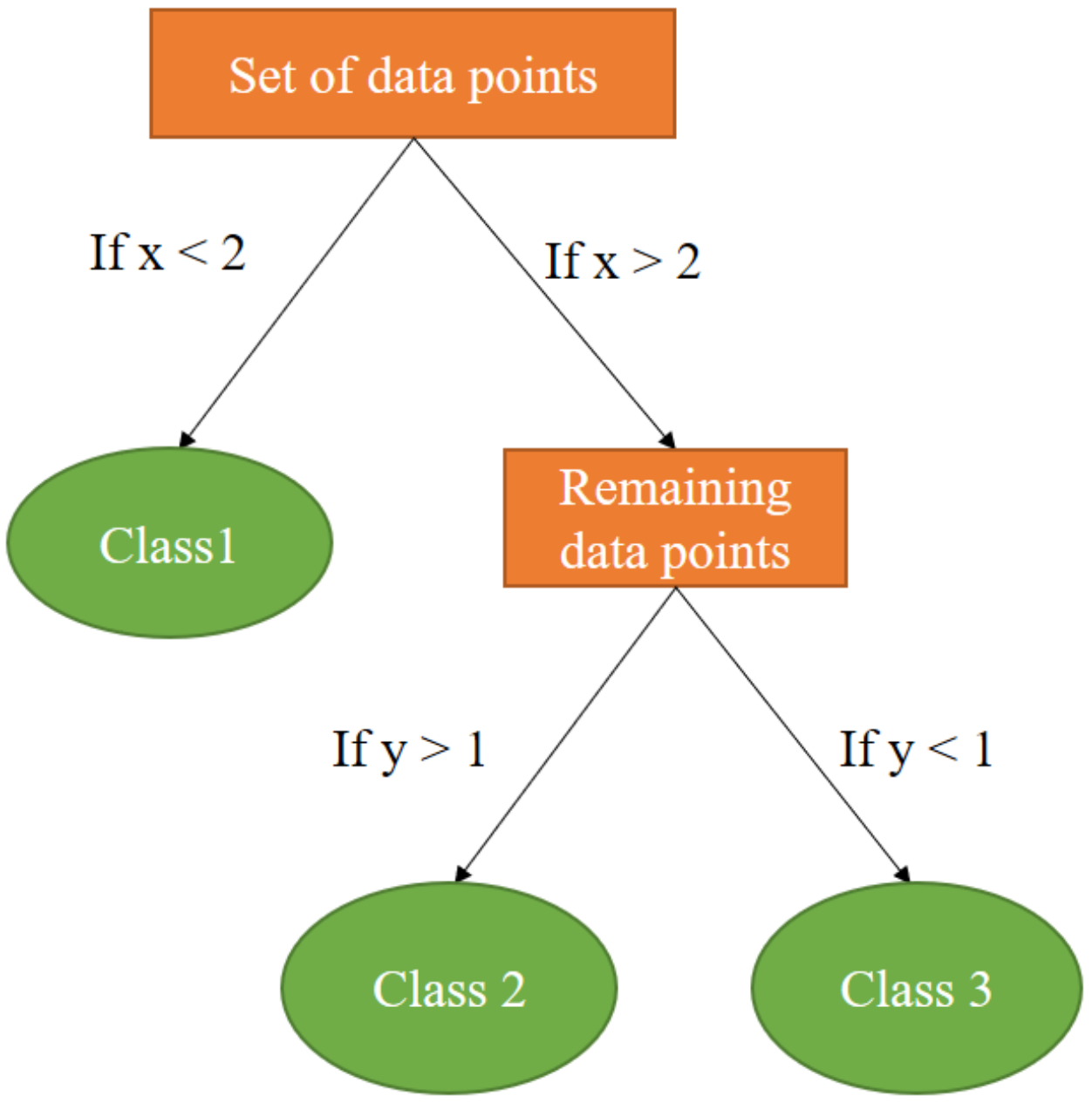

*Figure 5(b): Block diagram for decision tree*

An ensemble of such trees is then created to generate a random forest model. The user has the liberty to tune the number of trees and the number of randomly selected variables used to create the forests. The number of trees are by default kept to square root of number of variables, however, the user can change these values as well.

After the creation of forests, a portion of the data are used in training while the remaining portion is used in testing of the model. The training data are called *in bag* samples while the testing data are called *out of bag* samples.

The Gini Index is also used to calculate the Variable Importance (VIMP) (Ishwaran, 2007) i.e. how significant a variable is in classification. The more a variable's exclusion affects the classification accuracy, more is the importance of that variable for classification. This is called Mean Decrease in Accuracy or Mean Increase in Error and is calculated over the out of bag samples. Variable importance may also be called as *Permutation Importance.* The importance of VIMP measures are presented in Breiman, (2002) and Liaw and Wiener, (2002).

### 3.6.2 Forest-RC and oblique random forests

This method has already been introduced in section 2.3.2. In this method, the features that are used in classification are linearly combined to obtain more useful information from the dataset. So, if there is a feature set defined as $F$, the new feature set can be generated by multiplying the subset of original features by a rotation matrix, i.e.

$$F_{new} = R^T F \tag{3.20}$$

Because of this rotation, these kind of random forests are also known as ***oblique random forests.***

Such kind of feature rotation techniques have also been applied in other studies such as Xia et al., (2016) where SVM was used as a base classifier while PCA rotation and random matrices were used to generate a linear combination of features.

In this study, a different variant of Forest-RC algorithm is implemented in which instead of using the decision tree, a random forest is used as a base classifier. A similar approach has also been followed in Xia et al. (2018). The ensemble of random forests is then trained on the oblique feature set. Three types of techniques have been used to generate the linear combinations of the features. They are as follows:

i. Principal component analysis
ii. Sparse random projection matrix
iii. Complete random projection matrix

#### *3.6.2.1 Principal component analysis (PCA)*

It is a technique of multivariate analysis that is used to maximize the variance of linear combination of features. This is done by projecting the feature space along such axes that maximum information is extracted out of each feature and is contained in the new features. The procedure to evaluate the principal components is presented henceforth (Principal Component Analysis (PCA) Procedure | STAT 505, 2018).

Let $A$ be a random vector given as:

$$A = \begin{pmatrix} A_1 \\ A_2 \\ \vdots \\ A_n \end{pmatrix} \tag{3.21}$$

The variance-covariance matrix for $A$ can be given as:

$$\Sigma_{AA} = \begin{pmatrix} \sigma_1^2 & \sigma_{12} & \cdots & \sigma_{1n} \\ \sigma_{21} & \sigma_2^2 & \cdots & \sigma_{2n} \\ \vdots & \vdots & \ddots & \vdots \\ \sigma_{n1} & \sigma_{n2} & \cdots & \sigma_n^2 \end{pmatrix} \tag{3.22}$$

Now, these features are rotated using a linear combination:

$$\begin{aligned} B_1 &= c_{11}A_1 + c_{12}A_2 + \cdots + c_{1n}A_n \\ B_2 &= c_{21}A_1 + c_{22}A_2 + \cdots + c_{2n}A_n \\ \vdots \quad & \quad \vdots \qquad \vdots \qquad\qquad \vdots \\ B_n &= c_{n1}A_1 + c_{n2}A_2 + \cdots + c_{nn}A_n \end{aligned} \tag{3.23}$$

Here, $c_{i,j}$ is the rotation coefficient for rotation in the new space.

Since $B$ is a vector dependent on a random variable, it also has its variance that is given as:

$$\Sigma_{B_{ii}} = \sum_{p=1}^{n}\sum_{q=1}^{n} c_{ip}c_{iq}\sigma_{pq} \tag{3.24}$$

$$\Sigma_{B_{ii}} = c_i^{'}\Sigma_{AA}c_i \tag{3.25}$$

Also, the covariance between $B_i$ and $B_j$ can be given as:

$$\Sigma_{B_{ij}} = \sum_{p=1}^{n}\sum_{q=1}^{n} c_{ip}c_{jq}\sigma_{pq} \tag{3.26}$$

The principal components are found by subjecting the rotational coefficients to the certain constraints. The components with their corresponding constraints are described below:

**a. First principal component:**

It is the linear combination of features that corresponds to maximum variance and hence maximum possible information. So, we obtain such value of coefficients $c_{11}$, $c_{12}, \ldots, c_{1n}$ maximize the variance but subject it to the constraint that the sum of the square of coefficients is 1. That is,

Maximize:

$$\Sigma_{B_{11}} = \sum_{p=1}^{n} \sum_{q=1}^{n} c_{1p} c_{1q} \sigma_{pq} \tag{3.27}$$

$$\Sigma_{B_{11}} = c_1^{'} \Sigma_{AA} c_1 \tag{3.28}$$

Subject to:

$$c_1^T c_1 = \sum_{j=1}^{n} c_{1j}^2 = 1 \quad \text{(condition of orthogonality)}$$

(3.29)

**b. Second principal component:**

This is the linear combination of features that account for the remaining variance such that its correlation with the previous component is 0.

Maximize:

$$\Sigma_{B_{22}} = \sum_{p=1}^{n} \sum_{q=1}^{n} c_{2p} c_{2q} \sigma_{pq} \tag{3.30}$$

$$\Sigma_{B_{22}} = c_2^T \Sigma_{AA} c_2 \tag{3.31}$$

Subject to:

$$c_2^T c_2 = \sum_{j=1}^{n} c_{2j}^2 = 1 \quad \text{(condition of orthogonality)} \qquad (3.32)$$

and

$$\Sigma_{B_{12}} = \sum_{p=1}^{n}\sum_{q=1}^{n} c_{1p} c_{2q} \sigma_{pq} = 0 \qquad (3.33)$$

**c. $i^{th}$ principal component:**

Similarly, the $i^{th}$ principal component can be found by maximizing:

$$\Sigma_{B_{ii}} = \sum_{p=1}^{n}\sum_{q=1}^{n} c_{ip} c_{iq} \sigma_{pq} \qquad (3.34)$$

Subjected to:

$$c_i^T c_i = \sum_{j=1}^{n} c_{ij}^2 = 1 \quad \text{(condition of orthogonality)} \qquad (3.35)$$

and

$$\Sigma_{B_{1j}} = \sum_{p=1}^{n}\sum_{q=1}^{n} c_{1p} c_{jq} \sigma_{pq} = 0 \qquad (3.36)$$

$$\Sigma_{B_{2j}} = \sum_{p=1}^{n}\sum_{q=1}^{n} c_{2p} c_{jq} \sigma_{pq} = 0$$

$$\vdots \qquad \vdots \qquad \vdots$$

$$\Sigma_{B_{(j-1)j}} = \sum_{p=1}^{n}\sum_{q=1}^{n} c_{(j-1)p} c_{jq} \sigma_{pq} = 0$$

This makes sure that all the principal components are uncorrelated.

Following procedure is carried out to find the principal components:

a. Normalize the dataset using the Z-normalization, i.e.

$$Z = \frac{x - \mu}{\sigma} \tag{3.37}$$

b. Find the variance-covariance matrix of the normalized dataset.

$$\Sigma_{AA} = \begin{pmatrix} \sigma_1^2 & \sigma_{12} & \cdots & \sigma_{1n} \\ \sigma_{21} & \sigma_2^2 & \cdots & \sigma_{2n} \\ \vdots & \vdots & \ddots & \vdots \\ \sigma_{n1} & \sigma_{n2} & \cdots & \sigma_n^2 \end{pmatrix} \tag{3.38}$$

Here, $A$ represents the normalised feature vector.

c. Find the eigenvalues for the variance covariance matrix by:

$$\left| \begin{pmatrix} \sigma_1^2 - \lambda & \sigma_{12} & \cdots & \sigma_{1n} \\ \sigma_{21} & \sigma_2^2 - \lambda & \cdots & \sigma_{2n} \\ \vdots & \vdots & \ddots & \vdots \\ \sigma_{n1} & \sigma_{n2} & \cdots & \sigma_n^2 - \lambda \end{pmatrix} \right| = 0 \tag{3.39}$$

The roots of the equation will give the eigenvalues.

d. Arrange the eigenvalues in the descending order. Here, the highest eigenvalue will correspond to the first principal component and so on.

$$\lambda_1 \geq \lambda_2 \geq \ldots \geq \lambda_n$$

The sum of all these eigenvalues will come out to be 1.

e. Find the eigenvectors for all the corresponding eigenvalues.

$$[c_1, c_2, \ldots, c_n]$$

Here, $c_1$ corresponds to $\lambda_1$ and so on.

f. Multiply the eigenvectors with the original feature matrix to get the new feature matrix.

#### *3.6.2.2 Sparse random projection (SRP) matrix*

The sparse matrix is generated by drawing integers randomly from a Gaussian distribution with 0 mean and standard deviation 1. The matrix so formed only picks a subset of features at a time (since most of the elements are zero, their multiplication with the digital numbers of certain bands gives 0 and hence their effect is not considered) and then sends it to the base classifier (Xia et al., 2016).

$$\begin{bmatrix} [R_{11},\ldots,R_{1p}] & 0 & \cdots & 0 \\ 0 & [R_{21},\ldots,R_{2p}] & 0 & 0 \\ \vdots & \vdots & \ddots & \vdots \\ 0 & 0 & \cdots & [R_{q1},\ldots,R_{qp}] \end{bmatrix} \sim N(0,1)$$

Here,

$q$ : number of rows in the super matrix

$p$ : number of vectors in each diagonal cell of the super matrix

$R_{ij}$ : vector of random rotation coefficients

The matrix formation has been explained in section 4.6.2.2.2 of the thesis.

#### *3.6.2.3 Complete random projection (CRP) matrix*

Here also, the matrix is generated by drawing integers randomly from a Gaussian distribution with $mean = 0$ and $standard\ deviation = 1$ . However, unlike the previous matrix, here all the cells are occupied with a random integer, so the rotation is performed on the complete training set instead of a subset of features. It could be considered as a special case for the previous method (Xia et al., 2016).

$$\begin{bmatrix} r_{11} & r_{12} & \cdots & r_{1p} \\ r_{21} & r_{22} & \cdots & r_{2p} \\ \vdots & \vdots & \ddots & \vdots \\ r_{p1} & r_{p2} & \cdots & r_{pp} \end{bmatrix} \sim N(0,1)$$

## 3.7 Accuracy analysis

Accuracy analysis is what one would call "last but not the least" part of classification process. From classification point of view, it gives us information about how many instances in the dataset were misclassified. Since there are multiple classes involved in, the multinomial tests for accuracy is used. The most popular approach of calculating accuracy analysis parameters in multinomial tests is through the "confusion matrix". A confusion matrix is used to compare the predicted values obtained from the classifier to the reference values originally provided with the test set. It is a $c \times c$ matrix where $c$ is the number of classes in the dataset. The diagonal elements $(i,i)$ in the matrix represent the classes that are correctly classified. The off diagonal terms represent the instances that are misclassified. Table 4 (Rossiter, 2014) represents the confusion matrix. The misclassification is divided into two categories:

a. **Commission error:** This is the error for the case when the certain instance is reported in the class to which it does not belonged. They are reported on the $(i, j)$ cell of confusion matrix where $i$ is kept fixed and $j$ is varied.

*Table 4: Representation of confusion matrix*

| **Classes** | | $C_1$ | $C_2$ | $\cdots$ | $C_k$ | | **Row Total** |
|---|---|---|---|---|---|---|---|
| | | | | | | | |
| $C_1$ | | $a_{11}$ | $a_{12}$ | $\cdots$ | $a_{1,k}$ | | $a_{1+}$ |
| $C_2$ | | $a_{12}$ | $a_{22}$ | $\cdots$ | $a_{2,k}$ | | $a_{2+}$ |
| $\vdots$ | | $\vdots$ | $\vdots$ | $\ddots$ | $\vdots$ | | $\vdots$ |
| $C_k$ | | $a_{k,1}$ | $a_{k,2}$ | $\cdots$ | $a_{k,k}$ | | $a_{k+}$ |
| | | | | | | | |
| **Column Total** | | $a_{+1}$ | $a_{+2}$ | $\cdots$ | $a_{+k}$ | | $a$ |

b. **Omission error:** This is the error for the case when the certain instance is not reported in the class to which it actually belonged. They are reported on the $(i, j)$ cell of confusion matrix where $j$ is kept fixed and $i$ is varied.

Here,

$i$ : class predicted by the classifier

$j$ : class as given in the reference dataset

The notations used in the confusion matrix along with their explanations are tabulated in table 5 (Rossiter, 2014) below:

*Table 5: Notations in confusion matrix (Rossiter, 2014)*

| Notation | Explanation | Expression |
|---|---|---|
| | | |
| $a_{i,j}$ | The cell element in confusion matrix representing the element for class $j$ mapped as class $i$ | $a_{i,j}$ |
| $a_{i+}$<br><br>$a_{+j}$ | Marginal sum along the rows (predicted class)<br><br>Marginal sum along the columns (original class) | $\sum_{j=1}^{k} a_{i,j}$<br><br>$\sum_{i=1}^{k} a_{i,j}$ |
| $a_{+i}$<br><br>$a_{j+}$ | Marginal sum along the columns (predicted class)<br><br>Marginal sum along rows (original class) | $\sum_{j=1}^{k} a_{j,i}$<br><br>$\sum_{i=1}^{k} a_{j,i}$ |
| $a$ | Total number of instances in the test set | $\sum_{i=1}^{k}\sum_{j=1}^{k} a_{i,j}$<br>Or<br>$\sum_{j=1}^{k} a_{i+}$<br>Or<br>$\sum_{i=1}^{k} a_{+j}$ |

Sometimes, it is preferred to represent the confusion matrix as a proportion of the total number of instances where each element of the matrix is divided by the total number of instances in the dataset. It makes the calculation of accuracy statistics easier. The notations for proportionated confusion matrix are shown in table 6 (Rossiter, 2014).

*Table 6: Notations for normalised confusion matrix (Rossiter, 2014)*

| **Notation** | **Explanation** | **Expression** |
|---|---|---|
| | | |
| $p_{i,j}$ | Proportion of the cell element in confusion matrix representing the element for class $j$ mapped as class $i$ | $\frac{a_{i,j}}{a}$ |
| $p_{i+}$<br><br>$p_{+j}$ | Proportional marginal sum along the rows (predicted class)<br><br>Proportional marginal sum along the columns (original class) | $\sum_{j=1}^{k} p_{i,j}$<br><br>$\sum_{i=1}^{k} p_{i,j}$ |
| $p_{+i}$<br><br>$p_{j+}$ | Proportional marginal sum along the columns (predicted class)<br><br>Proportional marginal sum along rows (original class) | $\sum_{j=1}^{k} p_{j,i}$<br><br>$\sum_{i=1}^{k} p_{j,i}$ |

Generally, two statistics are used to evaluate the accuracy of the classification. They are:

a. Naïve statistics

b. Kappa statistics

### 3.7.1 Naïve statistics

The statistics tabulated above can be directly used to assess the overall accuracy or accuracy for class wise accuracies. These statistics are simple to compute. They are tabulated in table 7 (Hawadiya, 2015).

*Table 7: Naïve statistics (Hawadiya, 2015)*

| **Notation** | **Name** | **Expression** |
|---|---|---|
| | | |
| $A_O$ | Overall Accuracy | $\frac{\sum_{i=1}^{k} a_{i,i}}{a}$ OR $\sum_{i=1}^{k} p_{i,i}$ |
| $C_i$ | User's Accuracy | $\frac{a_{i,i}}{a_{i+}}$ OR $\frac{p_{i,i}}{p_{i+}}$ |
| $O_j$ | Producer's Accuracy | $\frac{a_{j,j}}{a_{+j}}$ OR $\frac{p_{j,j}}{p_{+j}}$ |
| $s_{i+}$ | Standard error for User's Accuracy | $\sqrt{\frac{C_i\left(1-C_i\right)}{a_{i+}}}$ |
| $s_{+j}$ | Standard error for Producer's Accuracy | $\sqrt{\frac{O_j\left(1-O_j\right)}{a_{+j}}}$ |

### 3.7.2 Kappa statistics

Developed by Cohen in 1960, this statistic aims to compare two classifiers i.e. if one classifier can be distinguished from another and which one among them performs better. It was introduced in the field of remote sensing by Congalton et al. (1983) and correct formulae for its variance are mentioned in Hudson and Ramm (1987). The kappa statistics are used along with the binomial tests (one-tailed and two-tailed tests) (Rossiter, 2014) to compare the classifiers. The kappa statistics are tabulated in table 8 (Hawadiya, 2015).

*Table 8: Kappa statistics (Hawadiya, 2015)*

| Notation | Name | Expression |
|---|---|---|
| | | |
| $\theta_1$ | Intermediate parameters | $\frac{1}{a}\sum_{i=1}^{k} a_{i,i}$ or $\sum_{i=1}^{k} p_{i,i}$ |
| $\theta_2$ | | $\frac{1}{a^2}\sum_{i=1}^{k} a_{i+}a_{+i}$ or $\sum_{i=1}^{k} p_{i+}p_{+i}$ |
| $\theta_3$ | | $\frac{1}{a^2}\sum_{i=1}^{k} a_{i,i}\left(a_{i+}+a_{+i}\right)$ or $\sum_{i=1}^{k} p_{i,i}\left(p_{i+}+p_{+i}\right)$ |
| $\theta_4$ | | $\frac{1}{a^3}\sum_{i=1}^{k}\sum_{j=1}^{k} a_{i,j}\left(a_{j+}+a_{+i}\right)^2$ or $\sum_{i=1}^{k}\sum_{j=1}^{k} p_{i,j}\left(p_{j+}+p_{+i}\right)^2$ |
| $\hat{\kappa}$ | Overall Kappa | $\frac{\theta_1-\theta_2}{1-\theta_2}$ |
| $\sigma_{\hat{\kappa}}$ | Standard error of overall kappa | $\sqrt{\frac{1}{a}\left(\frac{\theta_1(1-\theta_1)}{(1-\theta_2)^2}+\frac{2(1-\theta_1)(2\theta_1\theta_2-\theta_3)}{(1-\theta_2)^3}+\frac{(1-\theta_1)^2(\theta_4-4\theta_2^2)}{(1-\theta_2)^4}\right)}$ |
| $\hat{\kappa}_{i+}$ | Kappa user's accuracy | $\frac{p_{i,i}-p_{i+}.p_{+i}}{p_{i+}-p_{i+}.p_{+i}}$ |
| $\sigma_{\hat{\kappa}_{i+}}$ | Standard error of user's accuracy | $\sqrt{\frac{1}{2}\left[\frac{p_{i+}-p_{i,i}}{p_{i+}^3.(1-p_{+i})^3}.\left\{(p_{i+}-p_{i,i})(p_{i+}.p_{+i}-p_{i,i})+p_{i,i}(1-p_{i+}-p_{+i}+p_{i,i})\right\}\right]}$ |
| $\hat{\kappa}_{+j}$ | Kappa producer's accuracy | $\frac{p_{j,j}-p_{j+}.p_{+j}}{p_{j+}-p_{j+}.p_{+j}}$ |
| $\sigma_{\hat{\kappa}_{+j}}$ | Standard error of producer's accuracy | $\sqrt{\frac{1}{2}\left[\frac{p_{+j}-p_{j,j}}{p_{+j}^3.(1-p_{j+})^3}.\left\{(p_{+j}-p_{j,j})(p_{+j}.p_{j+}-p_{j,j})+p_{j,j}(1-p_{+j}-p_{j+}+p_{j,j})\right\}\right]}$ |

### 3.7.3 Hypothesis testing

These are simple tests, carried out to compare the two classifiers and check if one performs distinguishably better than the other. This thesis used one-tailed test to compare the classifiers. Following hypothesis is formed for comparison:

$$H_0 : \kappa_1 > \kappa_2$$

$$H_a : \kappa_1 \leq \kappa_2$$

This test requires the value of $Z$ statistic, which is computed as:

$$Z = \frac{\kappa_1 - \kappa_2}{\sqrt{\sigma_{\kappa_1}^2 + \sigma_{\kappa_2}^2}} \tag{3.40}$$

Here, $\kappa_1$ and $\kappa_2$ are kappa coefficients for the two classifiers. The $Z$ value is compared to the tabulated value of $Z$ at 95% level of confidence, which is given as $Z_{table} = 1.645$. If $Z > Z_{table}$ holds true, then the classifier 1 performs better than classifier 2 (Rossiter, 2014).

# 4 Methodology

This chapter discusses in detail all the methods followed that lead to the completion of this study. The methodology has been demonstrated in figure 6.

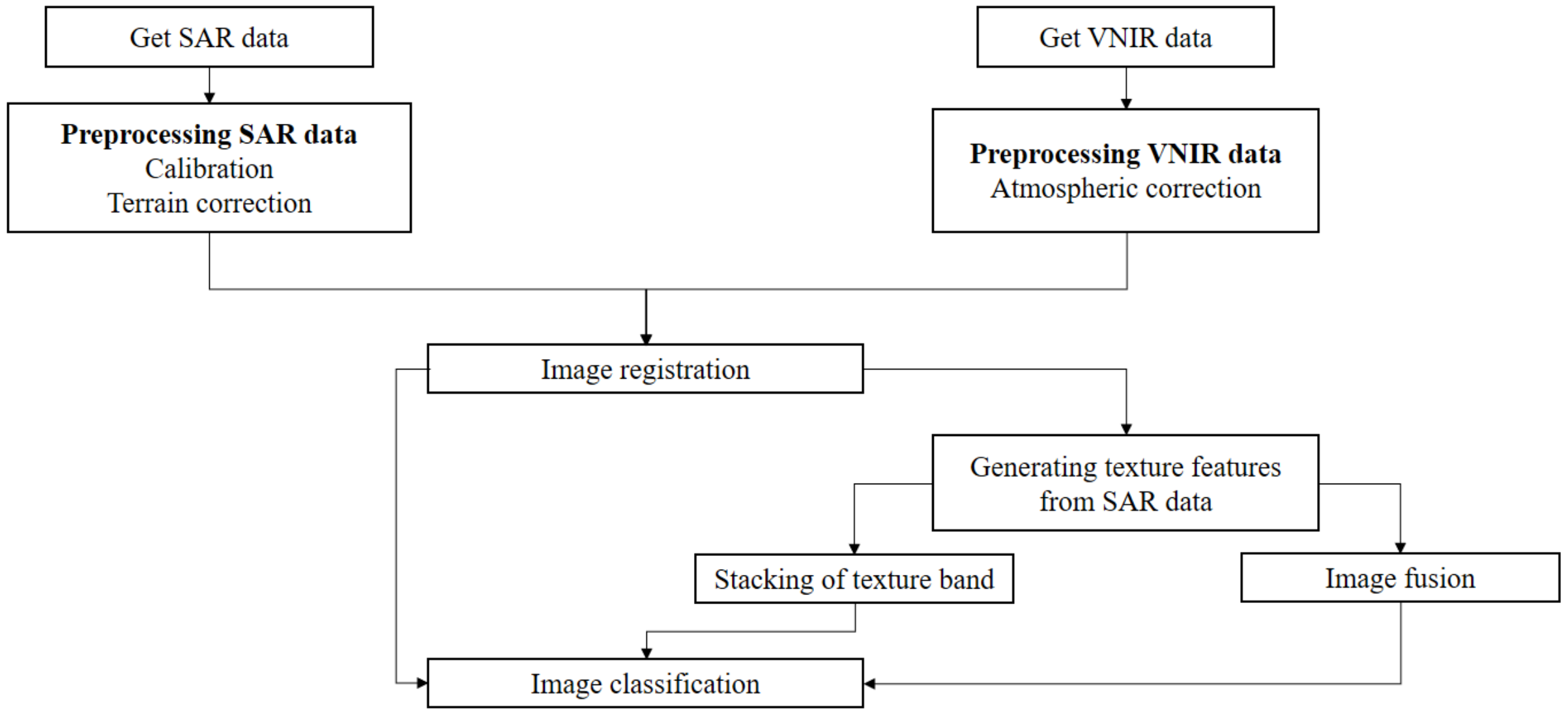

*Figure 6: Methodology block diagram*

## 4.1 Obtaining image data

The Sentinel-1A and Sentinel-2B satellite imageries are obtained from the official website of ESA, (Copernicus Open Access Hub, 2018) since the data are freely available. The Sentinel-1A data was collected for the date 26-Jan-2018 while for Sentinel-2B, it was collected for the date 24-Jan-2018.

## 4.2 Image preprocessing

The image preprocessing steps discussed in chapter 3 have been applied over the satellite imageries.

### 4.2.1 Preprocessing of Sentinel-1A imagery

Calibration and terrain correction have been used on the satellite imagery for its preprocessing as described in sections 3.1.1.1 and 3.1.1.2 respectively.

#### 4.2.1.1 Calibration

SNAP software is used to carry this process out. The result of this step is the sigma band for each of the polarisation that is generated by making use of amplitude and intensity bands. The result of calibration is sigma0 bands for VH and VV polarizations.

#### 4.2.1.2 Terrain correction

Range Doppler Terrain Correction from SNAP software is used to carry out this procedure. Bilinear interpolation method is used for image resampling and digital elevation model (DEM) resampling. WGS84 datum is used as map projection for generating DEM. This method applies the correction on the sigma bands obtained after calibration and also generates a DEM band.

### 4.2.2 Preprocessing of Sentinel-2B imagery

Atmospheric correction (section 3.1.2.1) has been used to preprocess Sentinel-2B imagery.

#### 4.2.2.1 Atmospheric correction

Sen2cor plugin in SNAP software is used to apply atmospheric correction on Sentinel-2 imagery. The input is the L1C product which is converted to L2A product. The general methodology involves correction for effects of cirrus clouds followed by scene classification, leading to Aerosol Optical Thickness (AOT) and water vapour retrieval and finally Top of Atmosphere (TOA) to Bottom of Atmosphere (BOA) conversion (Louis, 2016). Radiative Transfer Model and Digital Elevation Model (DEM) are provided as auxiliary data since they are required to carry out the atmospheric correction.

In this study, the Sentinel-2 bands are first resampled at 10 m resolution and then processed with sen2cor. After processing, out of the 13 bands, 3 bands of 60 m resolution (band 1, band 9 and band 10) have been discarded since, they are cirrus bands and do not provide land information. The working of sen2cor is shown in the flowchart in figure 7 (Jerome et al., 2016).

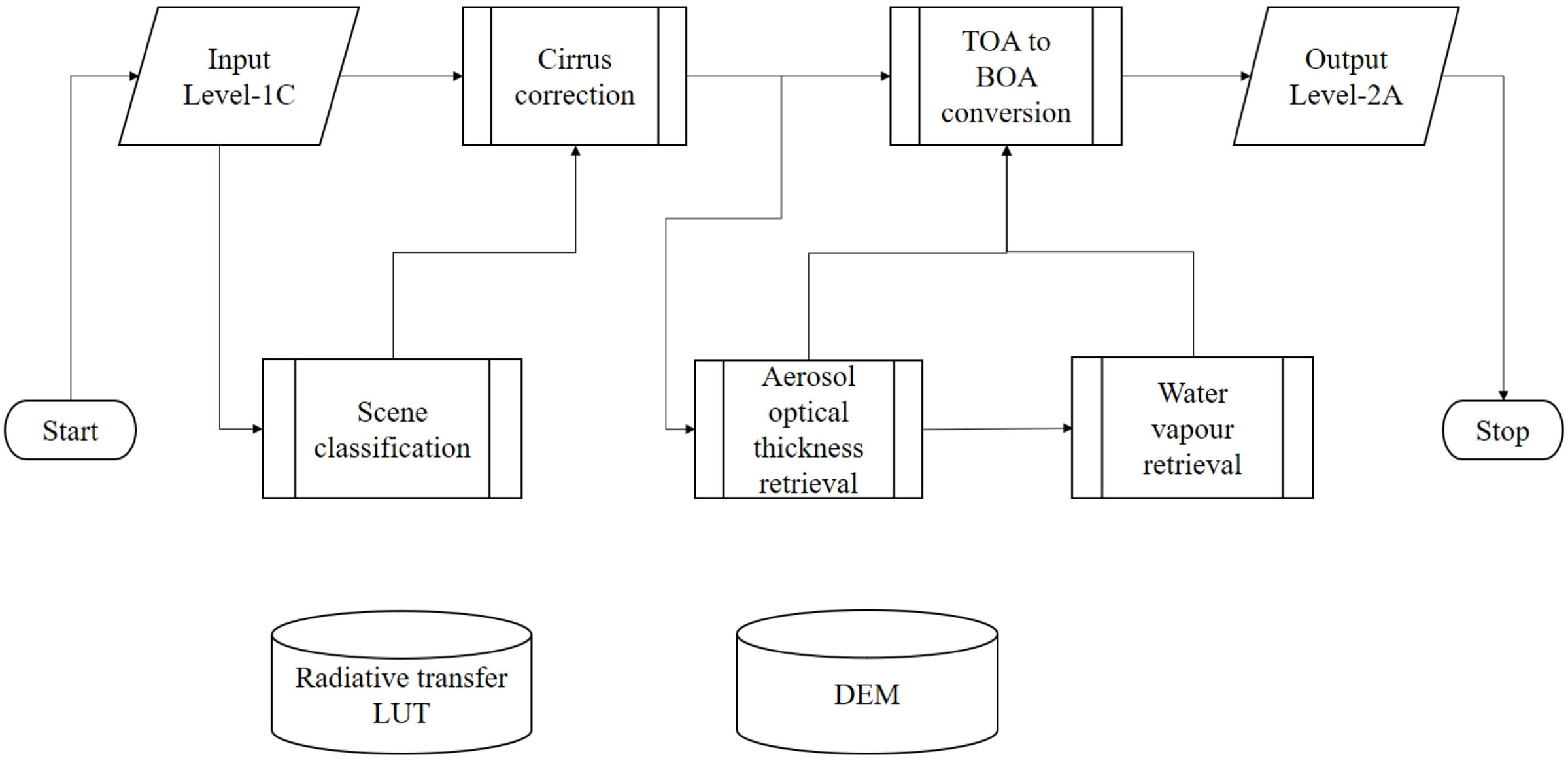

*Figure 7: Flowchart to demonstrate atmospheric correction using sen2cor processor (Jerome et al., 2016)*

## 4.3 Creation of new bands

The new bands for Sentinel-1A and Sentinel-2B imagery as discussed in section 3.2 are created using the Band Maths utility of SNAP software except the DEM band for Sentinel-1 imagery. It is generated after carrying out the terrain correction.

## 4.4 Image registration

The image registration is performed using the collocation tool in SNAP. Collocating two products implies that the pixel values of one product (the slave) are resampled into the geographical raster of the other (the master). The resampling has been done using cubic convolution technique. Here, the VNIR-SWIR bands are used as the master product whereas SAR bands are used as slave product.

## 4.5 Computing texture features

The Grey Level Co-occurrence Matrix tool from SNAP software has been used to calculate the texture parameters from the SAR image. The window size was kept to be $9 \times 9$ (as this was the smallest window size available in SNAP and so trimming of the edges was minimum) and the angle is set to ALL (to account for texture feature along all 4 directions) while computing the texture parameters.

## 4.6 Incorporating texture features in classification

Texture features are incorporated with VNIR-SWIR bands using two methods. These methods are described in subsequent sections.

### 4.6.1 Stacking of texture band with Sentinel-1 and Sentinel-2 bands

The homogeneity band generated from VH polarisation is simply stacked with the bundle of SAR from Sentinel-1 bands as well as VNIR-SWIR bands from Sentinel-2 imagery. These datasets are then sent for the classification process.

### 4.6.2 Image fusion

The Bayesian fusion technique of image fusion is adopted to fuse the homogeneity band for VH polarisation. The weighting parameter of 0.6 (60% SAR) is used because the work by Kanakan (2009) showed that the best classification accuracies were obtained at 60% SAR content.

#### *4.6.2.1 Speed enhancement of Bayesian fusion*

The Bayesian fusion method is implemented on MATLAB software. Though the algorithm was coded correctly, the code was inefficient and hence the implementation of fusion was taking too long to complete. Therefore, certain modifications are made in the code, which have increased its implementation speed greatly and without any information loss. The increment factor in the fusion speed has been presented in Chapter 5.

## 4.7 Image classification

This process involves three steps:

i. Training data selection

ii. Classification

iii. Accuracy analysis

### 4.7.1 Training data selection

The training data is obtained using the ENVI software and per pixel methodology of pixel selection is followed. The colour composite of red, blue and green bands of the preprocessed Sentinel-2 imagery was uploaded on the ENVI platform. Then, seven classes were visibly identified on the image namely, water, sand, built-up area, trees, grasses, algae and unused land. Then, for each class, the pixels were selected such that they were uniformly distributed throughout the image. The training data was saved as *.roi* files, which contained the image coordinates and geodetic coordinates in WGS-84 datum. This region of interest (ROI) file can be then superimposed on any of the Sentinel-2 band to obtain the pixel values from the corresponding cells. Table 9 shows class names with corresponding number of training pixels.

*Table 9: Classes with number of training pixels*

| Class name | Number of training pixels |
|---|---:|
| Water | 174 |
| Sand | 141 |
| Built-up area | 704 |
| Trees | 368 |
| Grass | 597 |
| Algae | 26 |
| Unused land | 511 |

In all the cases except *algae* class, it is made sure that the number of pixels lie in the range of 10n-100n (where n is the number of features, Lilleisand and Keifer, 1999). The *algae* class was present in a limited region and hence only a few samples are collected for it.

To obtain the data from Sentinel-1 bands, the collocated Sentinel-1 bands were opened in the ENVI software. Then, the same *.roi* was uploaded on the Sentinel-1 bands. This helped in extracting the corresponding pixel values from Sentinel-1 dataset as well.

The ground truth image for the area is created in eCognition software. The colour composite of band 2, band 3 and band 4 of the preprocessed Sentinel-2 imagery was opened in eCognition software and then using the spectral difference segmentation approach, the image was divided into segments and the segments were allocated a class using visual examination. Figures 8 (a) and 8 (b) respectively show the ground truth image with classes and spatial distribution of training pixels for each class.

### 4.7.2 Classification

Four classification methodologies have been used in this study:

i. Random forests
ii. Ensemble of random forests using PCA
iii. Ensemble of random forests using Sparse Random Projection
iv. Ensemble of random forests using Complete Random Projection

#### *4.7.2.1 Random forests*

The random forests are created with number of decision trees fixed to 30 (obtained after tuning). The number of features to be split at each node (called *mtry*) is fixed at square root of total number of features (recommended by Breiman, 2001) (Hastie et. al, 2017). Algorithm 1 explains the creation of random forests.

#### *4.7.2.2 Oblique ensembles of random forests*

The three methods discussed in section 3.6.2 to create oblique ensembles of random forests have been implemented on the dataset. Random forest has been used as base classifier. The

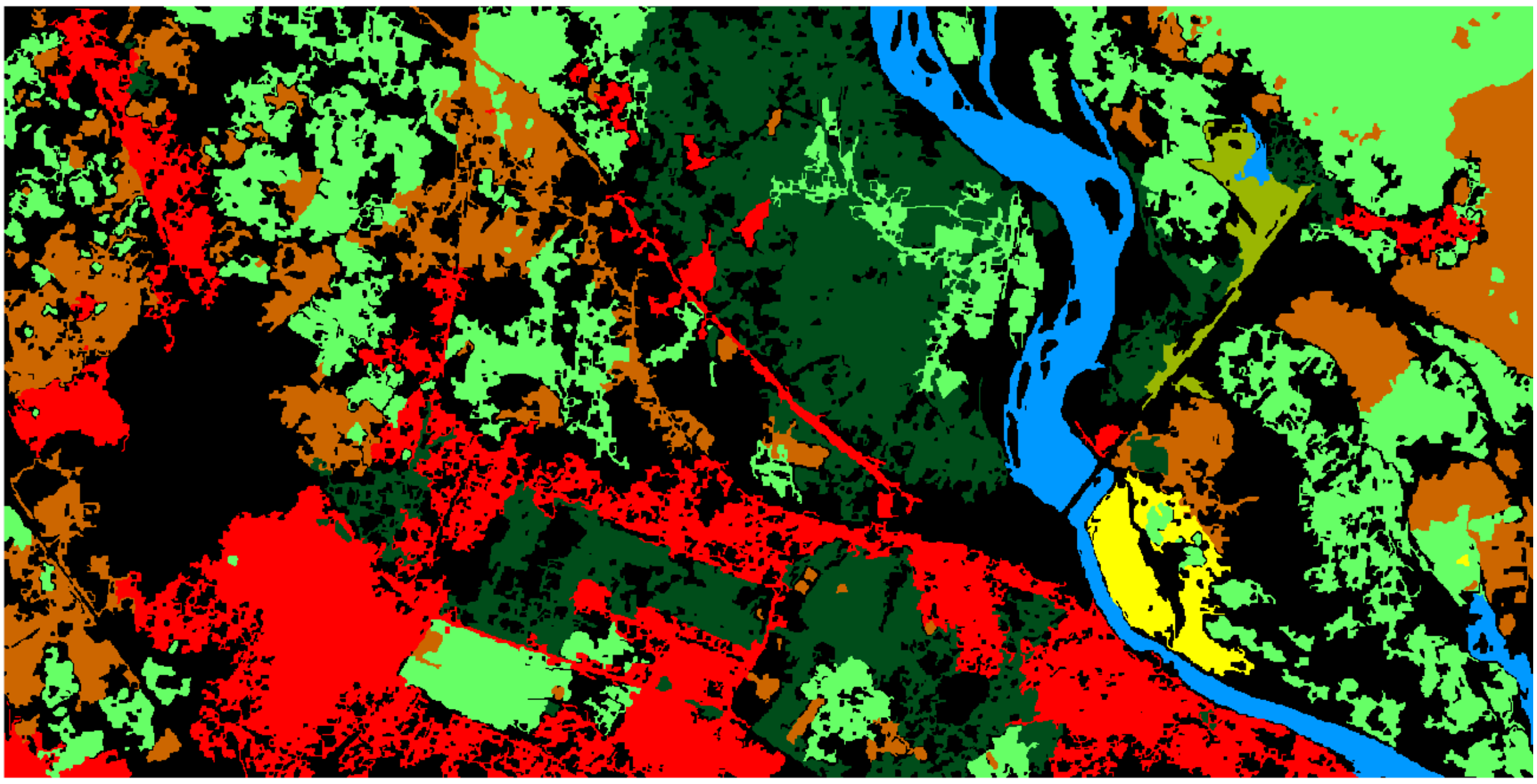

*(a)*

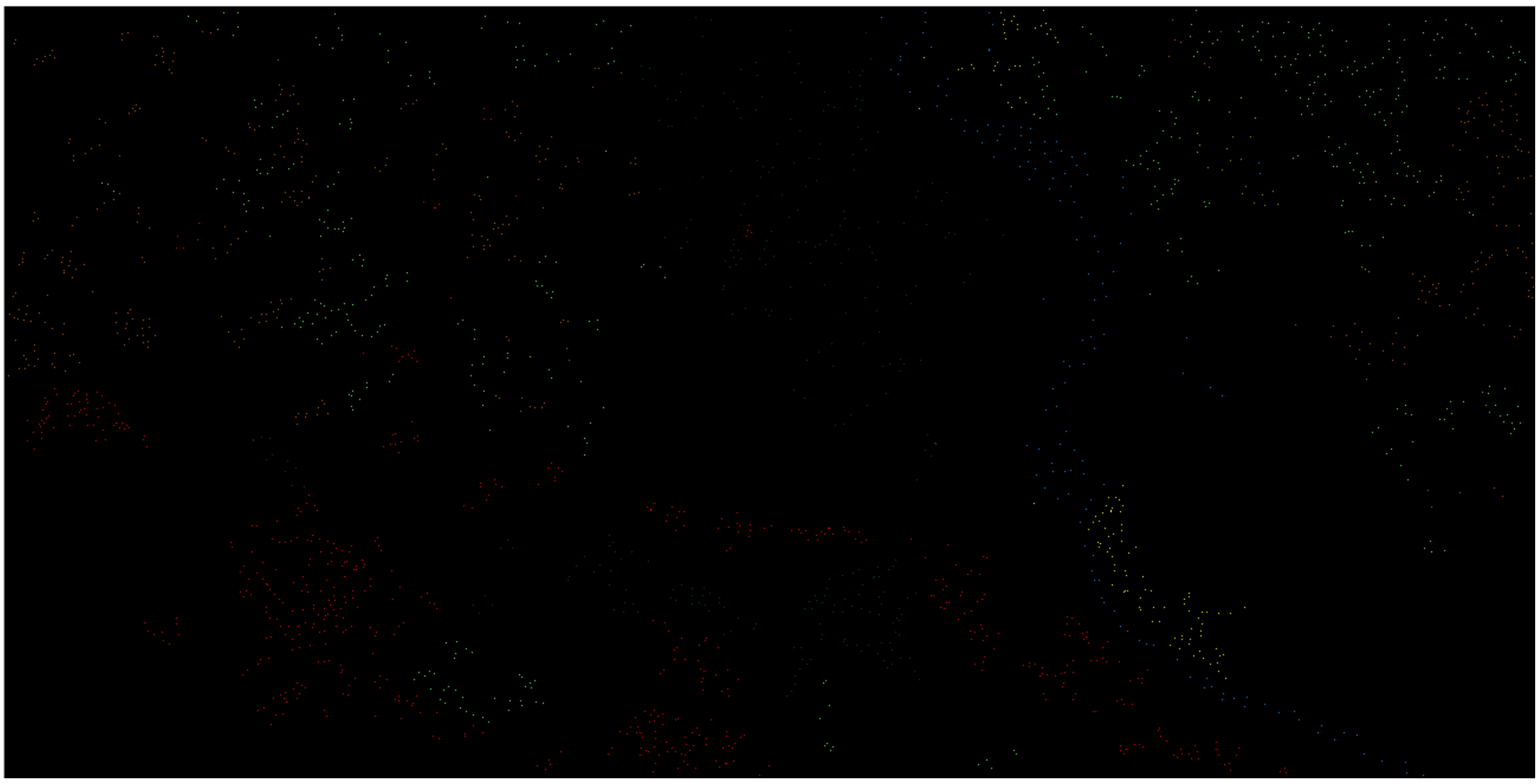

*(b)*

*Figure 8(a): Ground truth image with classes*
*Figure 8(b): Spatial distribution of training pixels*

number of random forests in ensembles are varied progressively to obtain a tuned value with number of decision tress in each random forest was fixed to 30 (chosen after tuning). The value of *mtry* for the random forest was kept fixed at default value of square root of number of features. The methodology to create the ensembles has been discussed in sections 4.7.2.2.1, 4.7.2.2.2 and 4.7.2.2.3.

### 4.7.3 Accuracy analysis

The *naïve* and *kappa* measures of accuracy analysis as discussed above have been used to carry out accuracy analysis of the classifiers used. For each of the classification carried out, the confusion matrix is generated using the reference test data and predictions obtained from the classifier. This confusion matrix is then used to calculate the *naïve* and *kappa* statistics. To get the value of $Z$ statistic for hypothesis testing kappa, random forest classifier has been used as the reference classifier and has been compared with the ensembles of random forests.

---

**Algorithm 1:** Procedure to create a random forest classifier

---

**Input:** Training dataset with *n* number of data points and *m* number of features with each data point having an associated class label. $C$ number of base classifiers (decision trees).

**Output:** Predicted labels for the testing dataset along with confusion matrix and naïve and kappa statistics.

1. Divide the dataset into training set and testing set.
2. For each of the decision tree
   a. Randomly select data points from the training set with replacement and square root of number of total features for splitting.
   b. Train the decision tree classifier with the training set and predict the classes for the data in the testing set.
3. Apply majority voting technique on *C* classifiers to get a single prediction.
4. Generate the confusion matrix from predicted classes and reference classes and calculate naïve and kappa statistics.

---

#### *4.7.3.1.1 Ensemble formation using PCA*

Algorithm 2 demonstrates the procedure to create the ensembles of random forests using PCA rotation.

---

**Algorithm 2:** Procedure to create random forest ensemble using PCA rotation

---

**Input:** Training dataset with *n* number of data points and *m* number of features with each data point having an associated class label. $C$ number of base classifiers (random forests).

**Output:** Predicted labels for the testing dataset along with confusion matrix and naïve and kappa statistics.

**1.** Divide the dataset into training set and testing set.

**2.** For the training set, divide the feature set into $p$ ($1 < p < m$, (Xia et al., 2016)) subsets.

**3.** Let each of the $p$ subsets contain $q$ features. The last subset may contain less number of features if the feature set is not entirely divisible by the number of subsets.

**4.** For each base learner

   a. Randomly select $50\% - 75\%$ (Xia et al., 2016) of the training data by bootstrap technique (by replacement), for each of the subsets and perform principal component analysis on this subset. A sparse matrix of the principal components will be obtained.
   b. Arrange the columns of the principal components matrix such that it could be multiplied with the training data.
   c. Multiply the matrix so formed with the original training set as well as testing set to get the new training and testing sets in the rotated plane.
   d. Train and test the classifiers on the new training and testing sets.
   e. Calculate the overall kappa value.

**5.** Give weight to each classifier equal to its overall kappa value.

**6.** Perform weighted majority voting (section 2.2.3.2) on the testing set using these weights and generate the final prediction.

**7.** Generate the confusion matrix from predicted classes and reference classes and calculate naïve and kappa statistics.

---

#### *4.7.3.1.2 Ensemble formation using SRP matrix*

Algorithm 3 demonstrates the procedure to create the ensembles of random forests using SRP matrix.

**Algorithm 3:** Procedure to create random forest ensemble using SRP matrix

**Input:** Training dataset with *n* number of data points and *m* number of features with each data point having an associated class label. $C$ number of base classifiers (random forests)

**Output:** Predicted labels for the testing dataset along with confusion matrix and naïve and kappa statistics.

1. Divide the dataset into training set and testing set.
2. For the training set, divide the feature set into $p$ ($1 < p < m$, (Xia et al., 2016)) subsets.
3. Let each of the $p$ subsets contain $q$ features. The last subset may contain less number of features if the feature set is not entirely divisible by the number of subsets.
4. For each base learner
   a. Generate a square matrix $R_{qq}$ of random integers for each subset such that
   $$R_{qq} \sim N(0,1)$$
   b. Arrange the columns of the random matrix such that it could be multiplied with the training data.
   c. Multiply the matrix so formed with the original training set as well as testing set to get the new training and testing sets in the rotated plane.
   d. Train and test the classifiers on the new training and testing sets.
   e. Calculate the overall kappa value.
5. Give weight to each classifier equal to its overall kappa value.
6. Perform weighted majority voting (section 2.2.3.2) on the testing set using these weights and generate the final prediction.
7. Generate the confusion matrix from predicted classes and reference classes and calculate naïve and kappa statistics.

#### *4.7.3.1.3 Ensemble formation using CRP matrix*

Algorithm 4 demonstrates the procedure to create the ensembles of random forests using CRP matrix.

---

**Algorithm 4:** Procedure to create random forest ensemble using CRP matrix

---

**Input:** Training dataset with *n* number of data points and *m* number of features with each data point having an associated class label. $C$ number of base classifiers (random forests).

**Output:** Predicted labels for the testing dataset along with confusion matrix and naïve and kappa statistics.

1. Divide the dataset into training set and testing set.
2. For each base learner
   a. Generate a square random matrix of size $m \text{ x } m$ where, $m$ is total number of features.
   $$R_{mm} \sim N(0,1)$$
   b. Multiply the matrix so formed with the original training set as well as testing set to get the new training and testing sets in the rotated plane.
   c. Train and test the classifiers on the new training and testing sets.
   d. Calculate the overall kappa value.
3. Give weight to each classifier equal to its overall kappa value.
4. Perform weighted majority voting (section 2.2.3.2) on the testing set using these weights and generate the final prediction.
5. Generate the confusion matrix from predicted classes and reference classes and calculate naïve and kappa statistics.

---

## 4.8 Datasets used

Five kinds of datasets have been created using the different kinds of bands from Sentinel-1 and Sentinel-2 products and the experiments are then performed on them. The datasets are listed below:

i. SAR bands
ii. SAR bands stacked with homogeneity band generated from VH band
iii. VNIR-SWIR bands
iv. VNIR-SWIR bands stacked with homogeneity band generated from VH band
v. VNIR-SWIR bands fused with homogeneity band generated from VH band

## 4.9 Experimental setup

The experimental setup consists of experiments performed for analysis of speed of image fusion and performance of classifiers in image classification.

### 4.9.1 Image fusion

The value of *w* (weighing factor) is kept to be 0.6 (section 4.6.2).

**i. Experiment 1:** In this experiment, the subsets of VNIR-SWIR bands and homogeneity band generated from VH band ranging from sizes 100 x 100 to 700 x 700 (all the bands are included in VNIR-SWIR imagery) are fused using the original and modified codes for Bayesian fusion and the execution time for both the codes is recorded. This process is repeated 10 times and the mean of the recorded execution time for each of the codes is calculated. Then, the mean execution time taken by original code is divided by the mean execution time taken by the modified code to get the increment factor in the speed of fusion.

ii. **Experiment 2:** In this experiment, the image size is fixed to 400 x 400 and the number of bands in VNIR-SWIR imagery are varied from 2-11 (for band 1, both the codes perform identically). For each of such cases, the homogeneity band is fused with VNIR-SWIR bands using both the codes of Bayesian fusion and execution time is recorded for each case. This process is repeated 10 times for each code and then the mean execution time is calculated. Then, the mean execution time taken by original code is divided by the mean execution time taken by the modified code to get the increment factor in the speed of fusion.

### 4.9.2 Image classification

The training data are selected from the datasets using stratified random sampling technique. For each of the dataset, training data has been increased progressively from 2% to 20% increasing at interval of 2% while the remaining portion of the dataset is reserved for testing. Training has been performed on smaller percentage of dataset because for larger datasets, the performance of the classifiers is statistically similar. Following experiments have been performed to achieve the thesis objective:

i. **Experiment 1:** In this experiment, the random forest classifier is run on the dataset where VNIR-SWIR bands is stacked with homogeneity band (chosen because this dataset has maximum number of features) at 10% of training data with number of decision trees increasing from 1 to the number at which the value of overall kappa tends to stagnate. For each number of decision tree, the random forest classifier is run 25 times and average overall kappa and average standard deviation for overall kappa is calculated. These average values are used to compare the performance of random forests on varying the number of decision trees. The number of decision trees is tuned to the value at which the average overall kappa starts to stagnate. This value of decision trees is used in the random forests while creating their ensembles.

ii. **Experiment 2:** In this experiment, keeping the value of decision trees fixed to the value obtained after tuning, random forest classifier is run all the datasets 25 times and the average overall kappa value along with average standard deviation for overall kappa is recorded for each of the datasets. In addition, the execution time is noted for each case as well and average execution time is calculated. Using average overall kappa, the image classification accuracies for different datasets are compared.

iii. **Experiment 3:** Using the tuned values of the number of decision trees, the ensembles are created for all the datasets. For each instance, the classifier is trained 25 times and the values of average overall kappa with corresponding average standard deviation and average producer and user kappa along with their corresponding average standard deviations are recorded. In addition, the average execution time for the ensembles is recorded as well. Using these values, following targets are achieved:

   **a.** The classification accuracies for different datasets are compared by using the average overall kappa values.

   **b.** Classification efficacy of random forests ensembles are compared to a single random forest for all the datasets using hypothesis testing (section 3.7.3) to identify if the ensembles are performing better than a single random forest or not.

   **c.** The average producer kappa values for the different classes for each of the dataset are compared to see how accurate the classes have been identified in the test set (Tso and Mather, 2009).

   **d.** The execution time for the ensembles is compared to that of a single random forest to judge the comparative execution speeds of the classifiers. For this VNIR-SWIR stacked with texture bands is used as dataset and training data are kept at 20%. This is done to ensure that maximum possible time is taken by each classifier so that the comparison becomes easier since the difference would be clearly visible. The number of base classifiers in ensembles is kept to be 30 (an arbitrary value, just to calculate the execution time).

iv. **Experiment 4:** In this experiment, the ensembles (using random forests as base classifier) are created with the algorithms discussed in section 4.7.3.1, keeping the number of decision trees in each random forest fixed to the tuned value. These ensembles are run on each of the dataset 25 times and number of forests in each ensemble is tuned to the value at which stagnation in the value of average overall kappa occurs. While tuning, the value of percentage of training data is fixed at 20% (maximum training percentage to avoid any unwanted fluctuations due to randomization) and the number of forests in the ensembles are increased progressively from 1-10 (since accuracy is increasing rapidly in this range), then in the intervals of 2 and 5 from 12 -30 (since the accuracy starts to show signs of stagnation) and finally 40 and 50 to ensure that stagnation has occurred.

v. **Experiment 5:** The classified imageries are generated using the tuned value of the parameters in the classifiers for all the datasets. The classified imageries are then visually compared with the ground truth image.

# 5 Results and Discussion

The results in accordance with the experiments mentioned in section 4.9 have been reported in this chapter.

## 5.1 Image fusion

The effect of fusion is compared visually and using classification accuracies. Figures 9 and 10 show the visual results before and after fusion using RGB bands for VNIR-SWIR imagery.

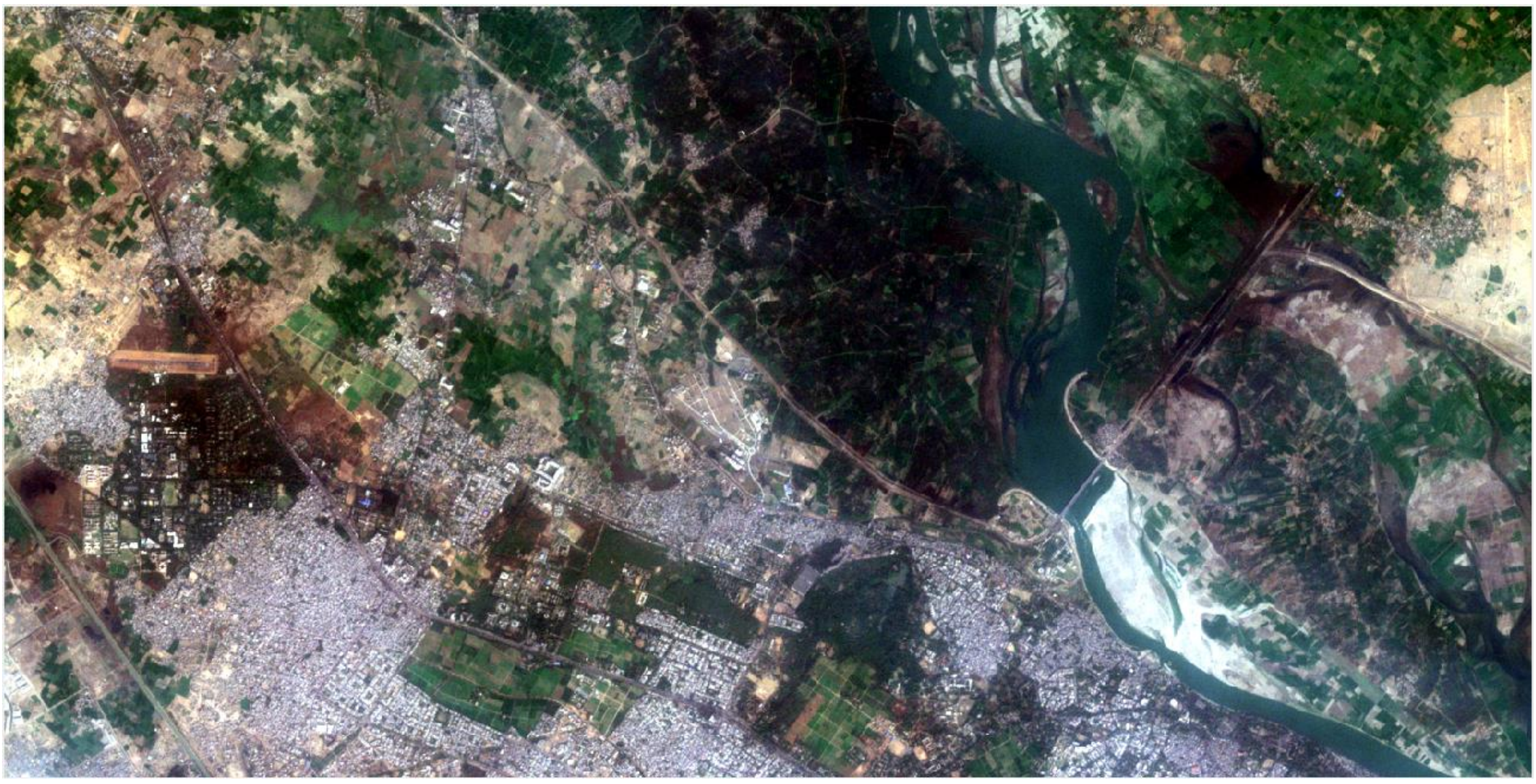

*Figure 9: False colour composite of band 4, band 3 and band 2 of bands of original VNIR-SWIR image*

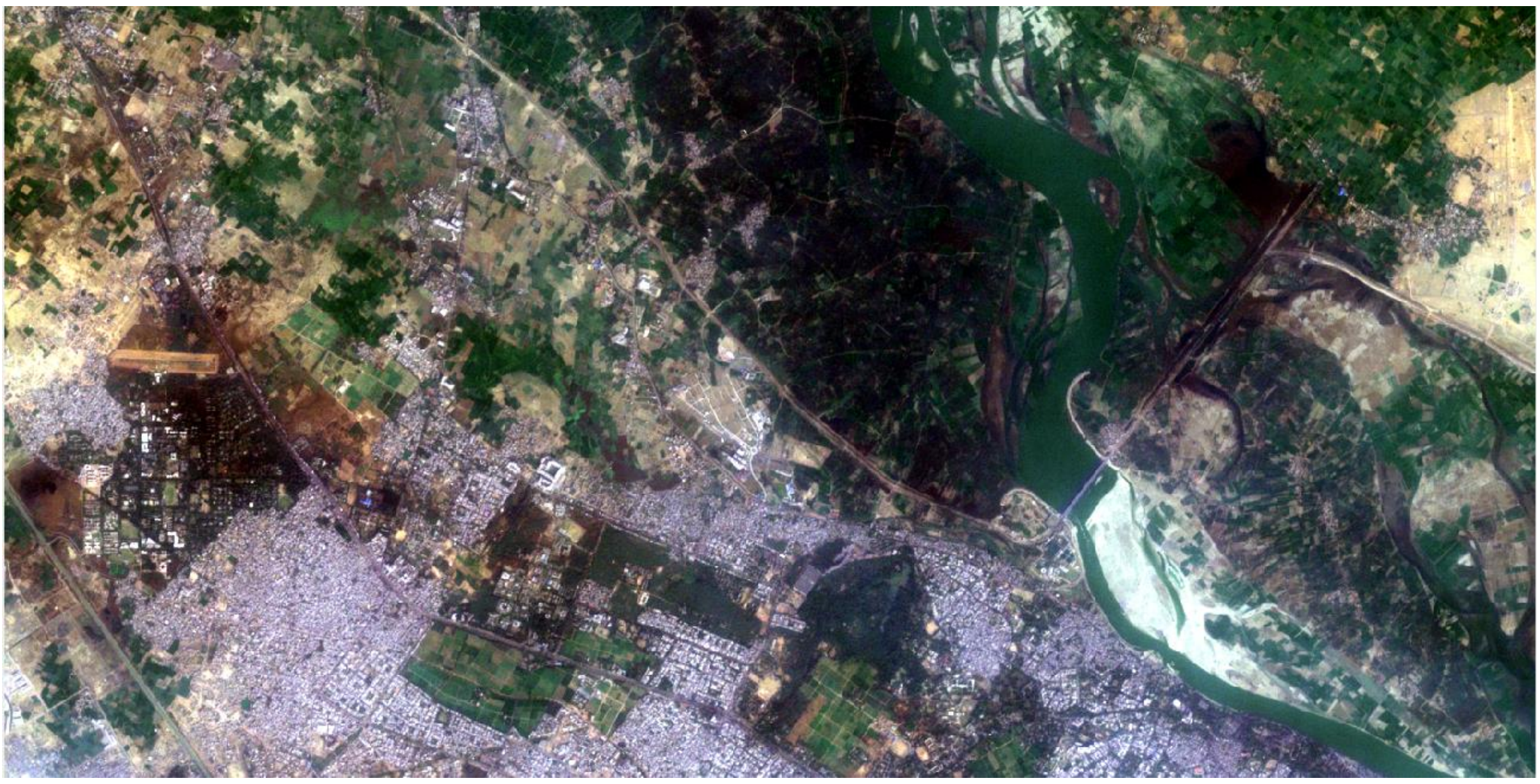

*Figure 10: False colour composite of band 4, band 3 and band 2 of VNIR-SWIR image fused with Homogeneity band generated from VH band*

### 5.1.1 Increment in the speed of Bayesian fusion

The results for the experiments conducted in section 4.9.1 have been discussed in this section.

**Experiment 1:**

The experiment shows that as the size of the image increases, the execution speed of the original code degrades. For the image of size 100 x 100, the execution time for modified code is about 1/35 times of than the original code while for a 700 x 700 image, the execution time of modified code reduces to about 1/3000 times than that of the original code. The time taken by both the codes and the increment factors are represented in table 10 and their trends can be visualised in figure 11.

**Experiment 2:**

This experiment shows that as the number of bands in the imagery increases, the execution time of the original Bayesian fusion code increases. For an image with 2 bands, the original code works about 75 times slower than the modified code while for 11 bands, the original code works about 552 times slower than the modified code. The time taken by both the codes and the increment factors are represented in table 11 and their trends can be visualised in figure 12.

*Table 10: Execution time and increment factor for modified code for implementing Bayesian fusion for different sizes of imagery*

| Image Size | Time taken by original code ($t_1$ sec) | Time taken by modified code ($t_2$ sec) | Increment factor ($t_1/t_2$) |
|---|---|---|---|
| | | | |
| 100 x 100 | 1.0356 | 0.0289 | 35.8045 |
| 200 x 200 | 3.8207 | 0.0720 | 53.0327 |
| 300 x 300 | 18.4066 | 0.1295 | 142.0854 |
| 400 x 400 | 90.5465 | 0.2104 | 430.2925 |
| 500 x 500 | 382.5063 | 0.3163 | 1209.2170 |
| 600 x 600 | 931.0949 | 0.4507 | 2066.0530 |
| 700 x 700 | 1841.2399 | 0.6057 | 3039.7360 |

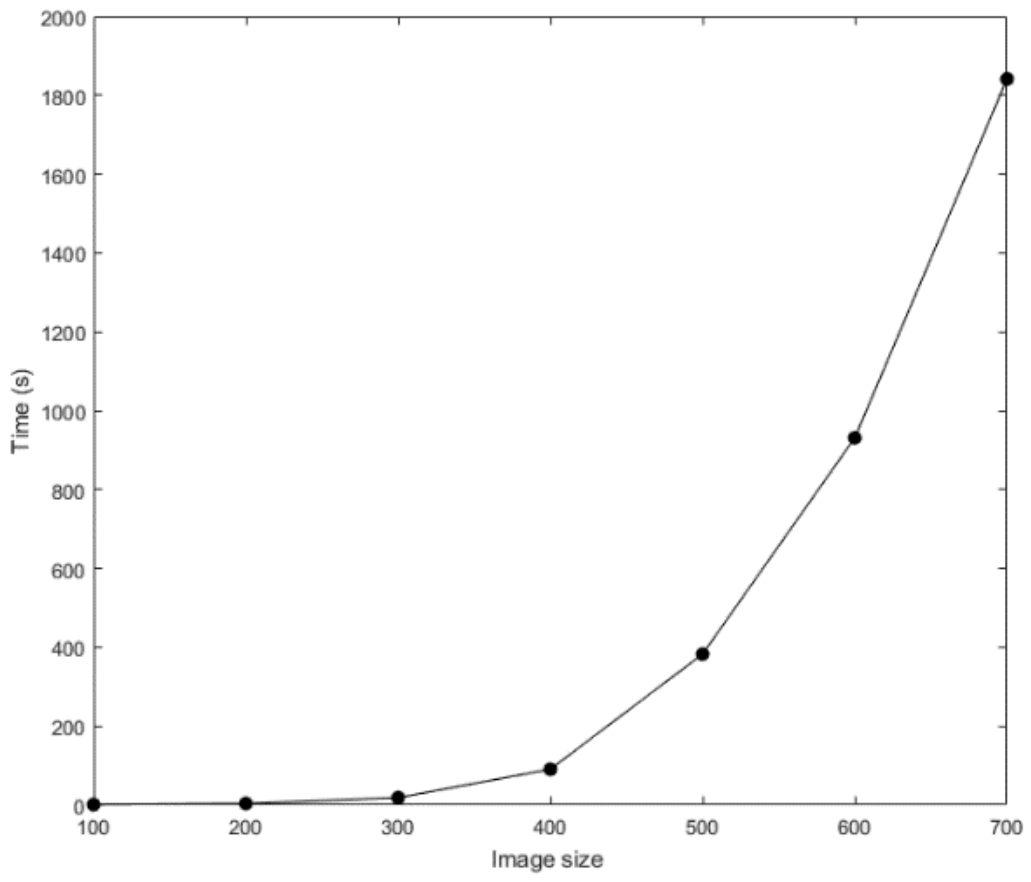

*(a)*

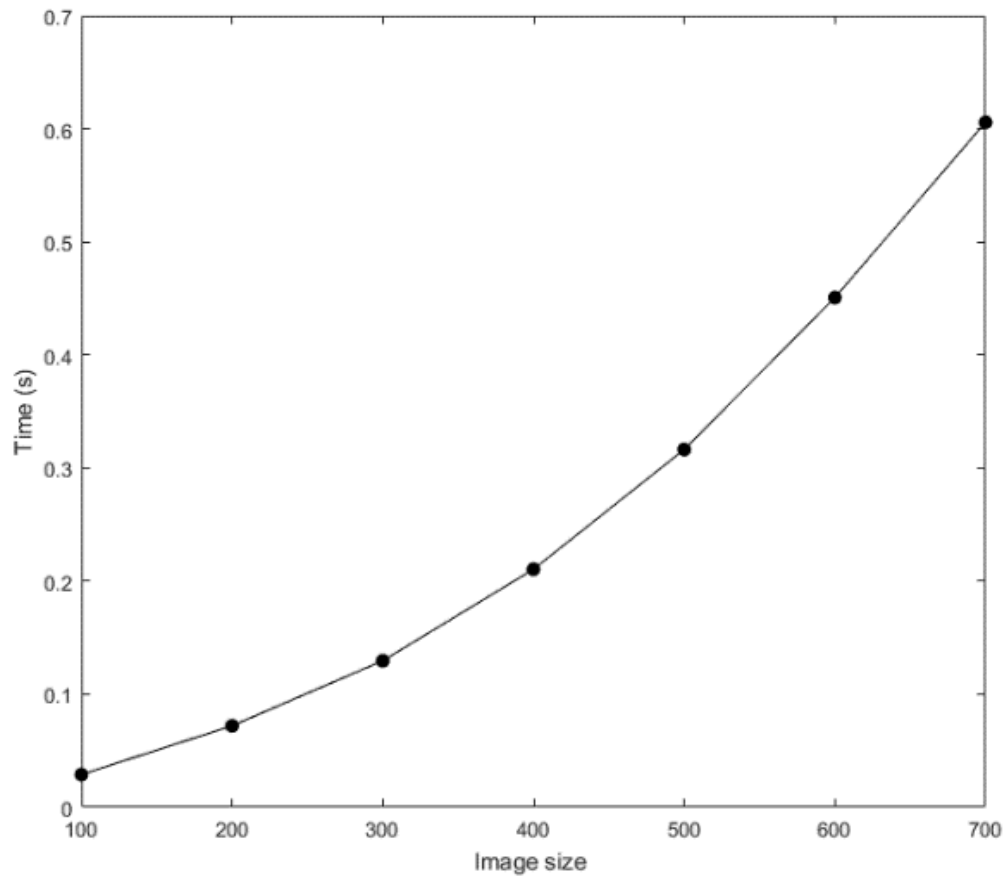

*(b)*

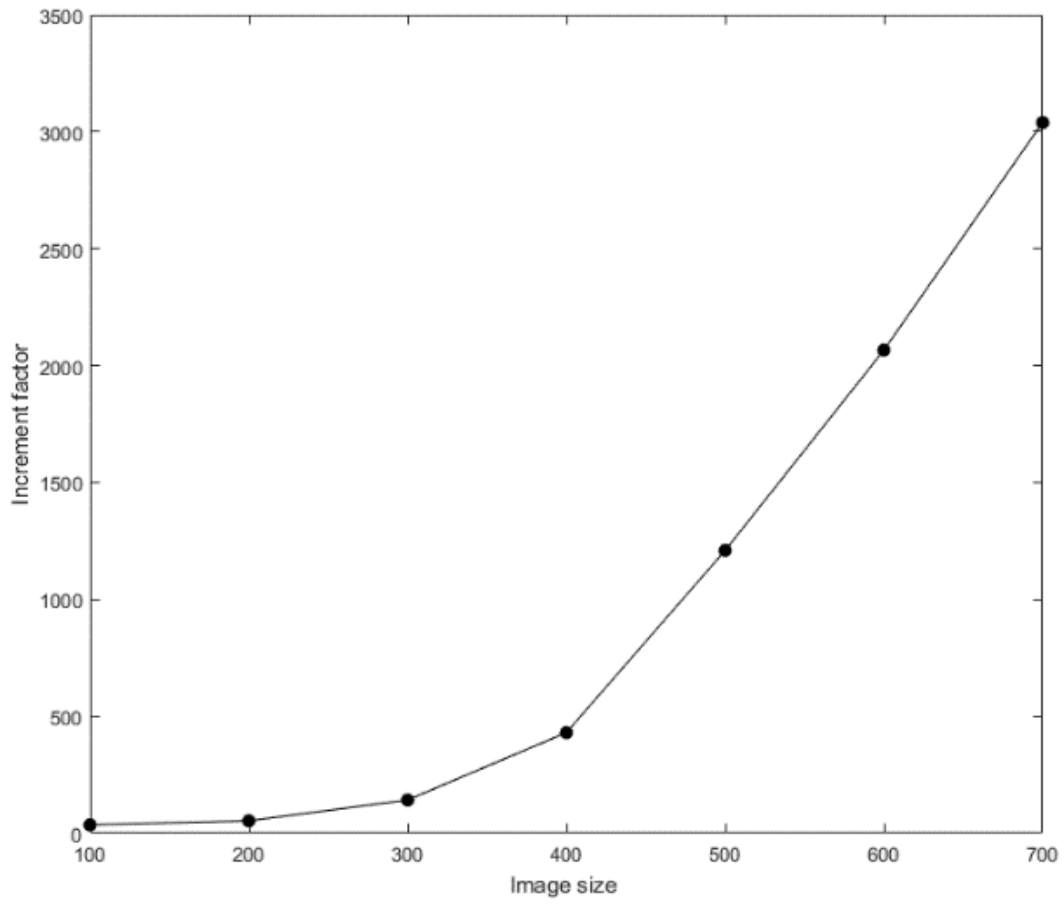

*(c)*

*Figure 11(a): Execution time for original code for Bayesian fusion*
*Figure 11(b): Execution time for modified code for Bayesian fusion*
*Figure 11(c): Increment factor for modified code for Bayesian fusion*

*Table 11: Execution time and increment factor for modified code for implementing Bayesian fusion for different number of bands*

| Number of bands | Time taken by original code ($t_1$ sec) | Time taken by modified code ($t_2$ sec) | Increment factor ($t_1/t_2$) |
|---|---|---|---|
| 2 | 8.8915 | 0.1205 | 75.0971 |
| 3 | 13.6501 | 0.1271 | 107.8207 |
| 4 | 18.4920 | 0.1379 | 136.3717 |
| 5 | 24.1002 | 0.1464 | 165.8651 |
| 6 | 32.0408 | 0.1564 | 205.5215 |
| 7 | 43.5568 | 0.1668 | 264.7830 |
| 8 | 58.5215 | 0.1755 | 334.4086 |
| 9 | 67.8913 | 0.1877 | 369.7783 |
| 10 | 83.4844 | 0.1973 | 430.5539 |
| 11 | 112.5061 | 0.2053 | 552.0417 |

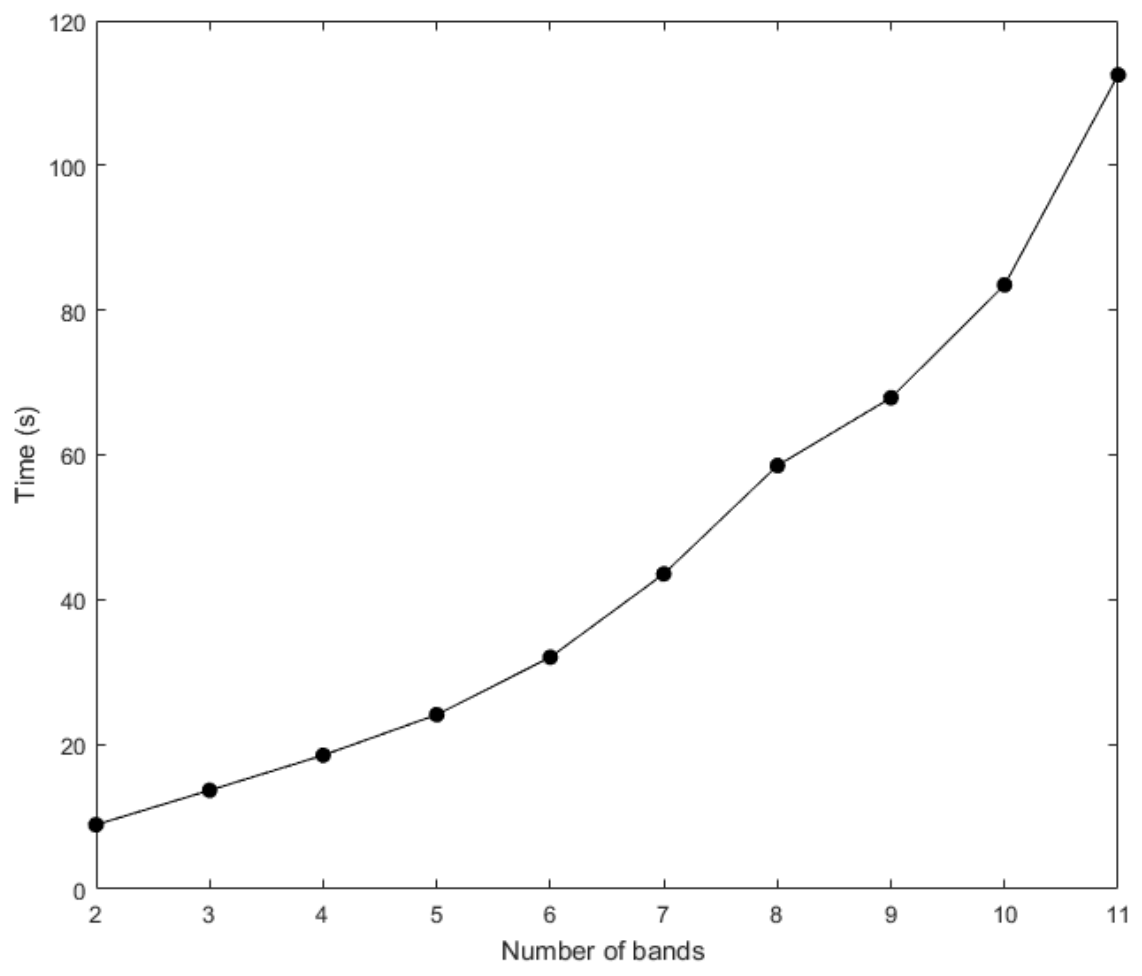

*(a)*

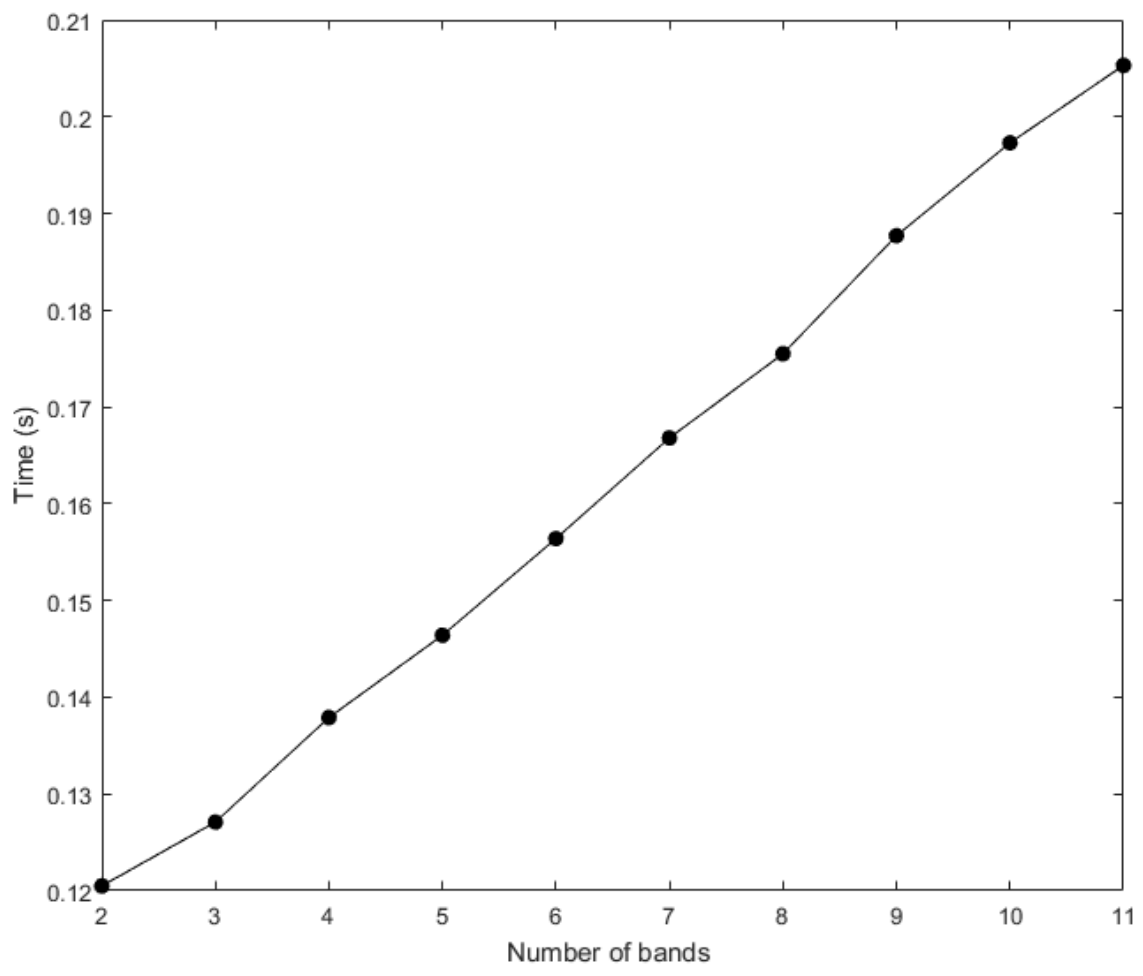

*(b)*

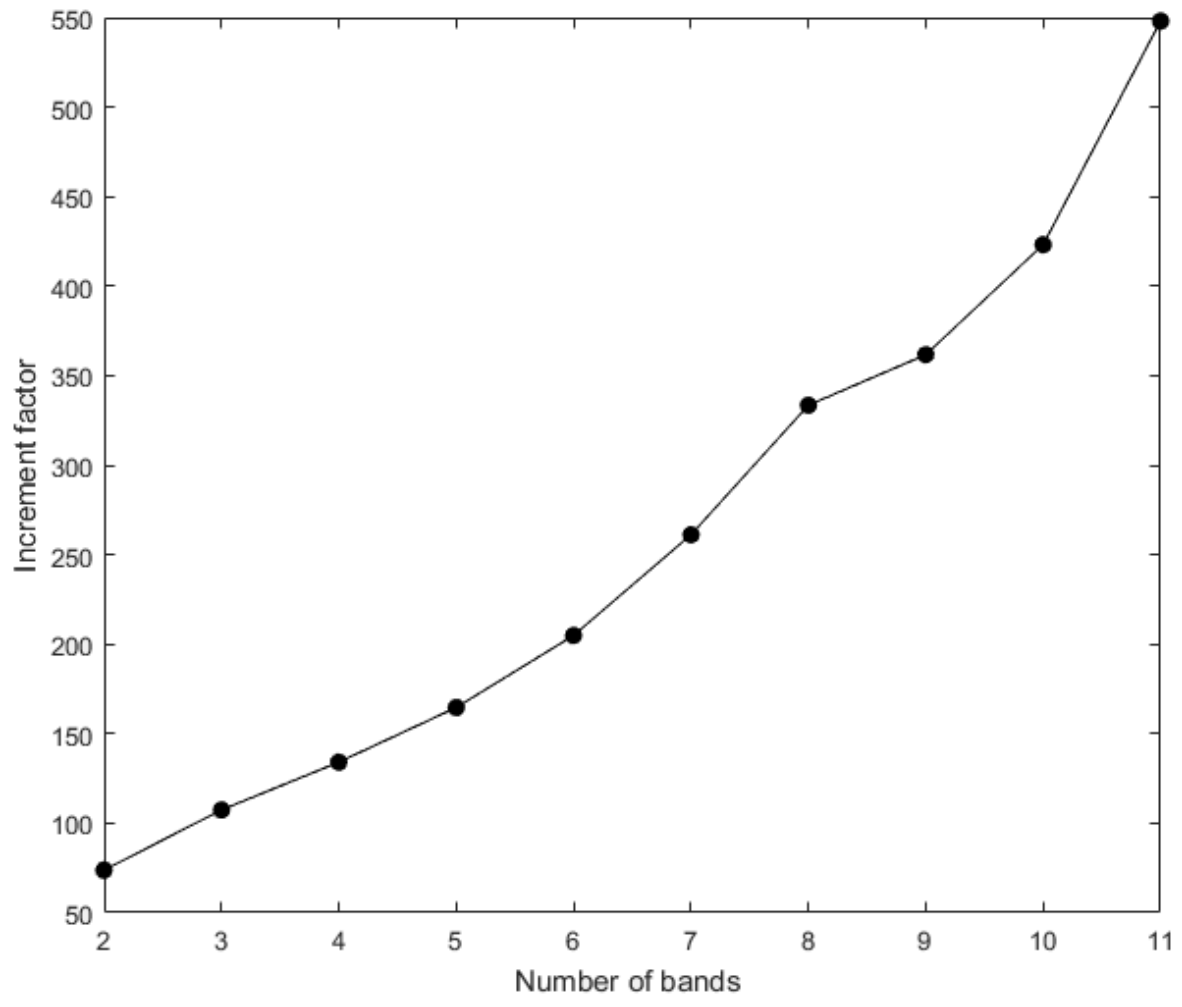

*(c)*

*Figure 12(a): Execution time for original code for Bayesian fusion*
*Figure 12(b): Execution time for modified code for Bayesian fusion*
*Figure 12(c): Increment factor for modified code for Bayesian fusion*

## 5.2 Image classification

For image classification, the accuracy analysis as mentioned in section 4.6.2 is carried out and the overall kappa obtained have been plotted against percentage of training data used for each of the classifier used. Using these plots, one classifier can be compared to another. Similarly, such plots are developed to compare the overall kappa for each kind of datasets.

### 5.2.1 Tuning of number of decision trees for random forests

The overall kappa values corresponding to varying the number of decision trees in the random forest classifier have been tabulated in table 11.

*Table 12: Overall kappa for corresponding number of trees in Random Forest*

| Number of trees | Average overall kappa |
|---|---|
| 1 | 0.8853 |
| 2 | 0.8782 |
| 3 | 0.9164 |
| 4 | 0.9178 |
| 5 | 0.9292 |
| 6 | 0.9283 |
| 7 | 0.9338 |
| 8 | 0.9342 |
| 9 | 0.9361 |
| 10 | 0.9382 |
| 11 | 0.9393 |
| 12 | 0.9387 |
| 13 | 0.9389 |
| 14 | 0.9407 |
| 15 | 0.9421 |
| 16 | 0.9406 |
| 17 | 0.9414 |
| 18 | 0.9419 |
| 19 | 0.9429 |
| 20 | 0.9415 |
| 21 | 0.9411 |
| 22 | 0.9441 |
| 23 | 0.9437 |
| 24 | 0.9446 |
| 25 | 0.9433 |
| 26 | 0.9434 |
| 27 | 0.9449 |
| 28 | 0.9453 |
| 29 | 0.9436 |
| 30 | 0.9442 |

Figure 13 shows the trend that overall kappa follows with change in the number of decision trees.

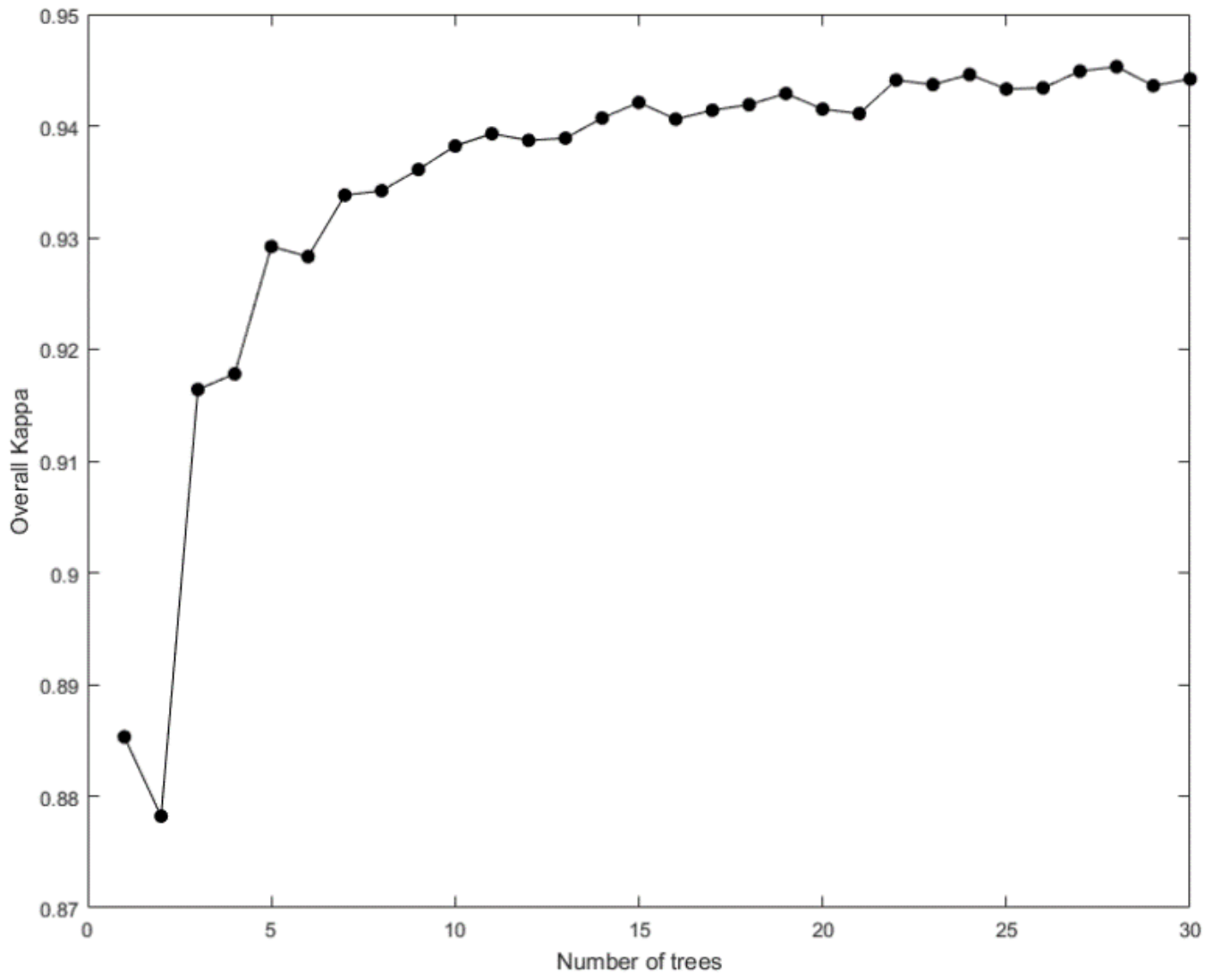

*Figure 13: Tuning curve for random forest*

The optimum values for the number of tree is calculated by performing hypothesis testing between the first classifier and the *i^th^* classifier. The Z-values obtained for all the classifiers are compared and the number of decision tress at which the classifier has highest Z-value is considered as the most optimum value that can be used for training the random forest.

From the tuning of random forest, the value of optimum number of decision trees comes out to be 28.

### 5.2.2 Comparison of accuracies on different datasets for same classifier

In this section, for the same kind of classifier, classification accuracies are compared for the five kind of datasets used. The value of trees in the decision trees for random forests and ensembles is kept to be 30 (nearest multiple of 10 to the tuned value of 28). The value of *number of forests* is kept fixed at 20 (an arbitrary value to suggest that results would follow the same trend for all such cases). Tables 13-16 display the overall kappa values for various datasets. Figures 14-17 display the trend with which average overall kappa varies for all the classifiers used on all 5 datasets.

*Table 13: Overall kappa for different datasets for Random Forest*

| | SAR | SAR stacked with texture band | VNIR-SWIR | VNIR-SWIR stacked with texture band | VNIR-SWIR fused with texture band |
|---|---|---|---|---|---|
| **Percentage training data** | *Overall kappa* | | | | |
| **2** | 0.5072 | 0.5822 | 0.8628 | 0.8973 | 0.8714 |
| **4** | 0.5504 | 0.6125 | 0.9067 | 0.9191 | 0.9038 |
| **6** | 0.5767 | 0.6229 | 0.9186 | 0.9356 | 0.9218 |
| **8** | 0.5754 | 0.6375 | 0.9221 | 0.9379 | 0.9291 |
| **10** | 0.5870 | 0.6450 | 0.9275 | 0.9457 | 0.9309 |
| **12** | 0.5918 | 0.6551 | 0.9361 | 0.9487 | 0.9422 |
| **14** | 0.5960 | 0.6610 | 0.9384 | 0.9512 | 0.9455 |
| **16** | 0.6011 | 0.6621 | 0.9385 | 0.9523 | 0.9457 |
| **18** | 0.6061 | 0.6694 | 0.9408 | 0.9554 | 0.9484 |
| **20** | 0.6048 | 0.6735 | 0.9460 | 0.9577 | 0.9481 |

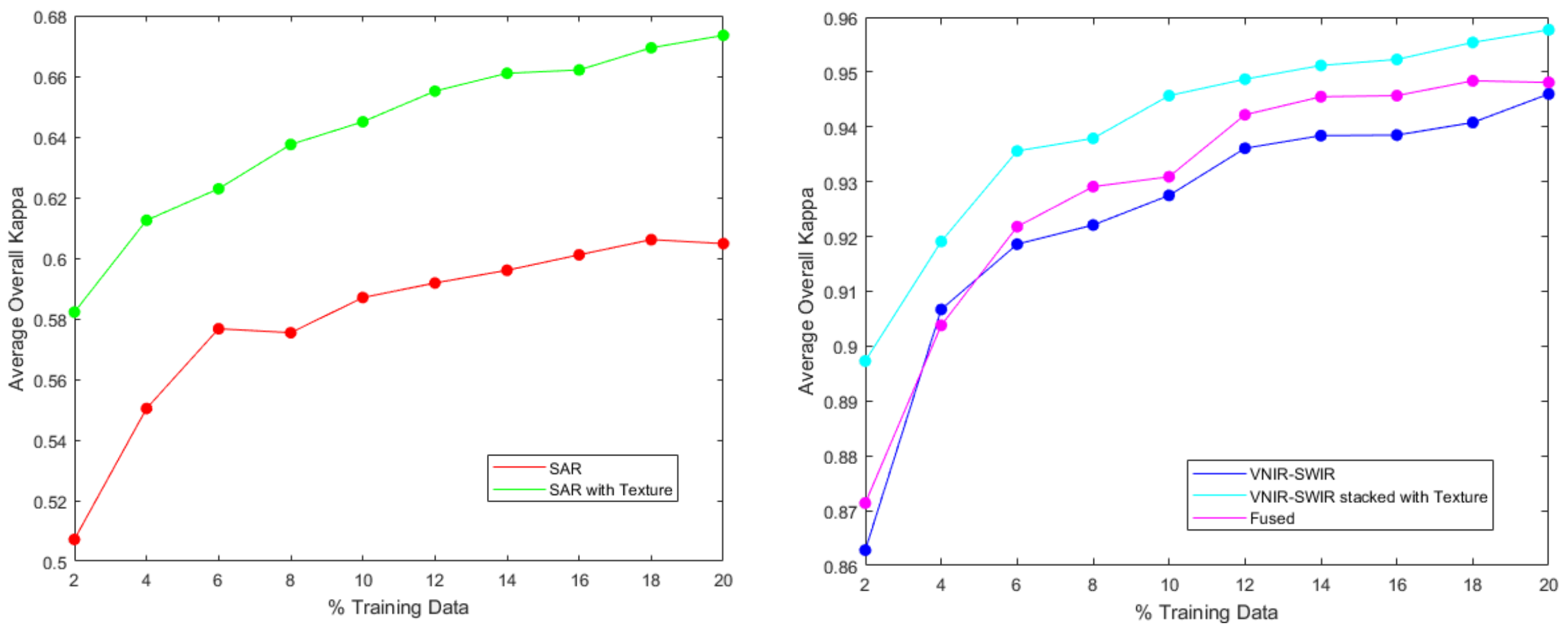

*(a)* *(b)*

*Figure 14 (a): Variation in average overall kappa for SAR bands and SAR bands stacked with texture band for random forest*
*Figure 14 (b): Variation in average overall kappa for VNIR-SWIR bands, VNIR-SWIR bands stacked with texture and VNIR-SWIR bands fused with texture band for random forest*

*Table 14: Overall kappa for different datasets for PCA-RFE*

| | SAR | SAR stacked with texture band | VNIR-SWIR | VNIR-SWIR stacked with texture band | VNIR-SWIR fused with texture band |
|---|---|---|---|---|---|
| **Percentage training data** | *Overall kappa* | | | | |
| **2** | 0.4857 | 0.5857 | 0.8929 | 0.9335 | 0.9009 |
| **4** | 0.5360 | 0.6112 | 0.9158 | 0.9408 | 0.9284 |
| **6** | 0.5624 | 0.6282 | 0.9280 | 0.9519 | 0.9433 |
| **8** | 0.5742 | 0.6480 | 0.9383 | 0.9582 | 0.9478 |
| **10** | 0.5807 | 0.6525 | 0.9436 | 0.9590 | 0.9487 |
| **12** | 0.5888 | 0.6644 | 0.9484 | 0.9637 | 0.9551 |
| **14** | 0.5972 | 0.6704 | 0.9497 | 0.9656 | 0.9555 |
| **16** | 0.5973 | 0.6752 | 0.9504 | 0.9667 | 0.9577 |
| **18** | 0.6032 | 0.6762 | 0.9511 | 0.9652 | 0.9584 |
| **20** | 0.6098 | 0.6823 | 0.9578 | 0.9683 | 0.9602 |

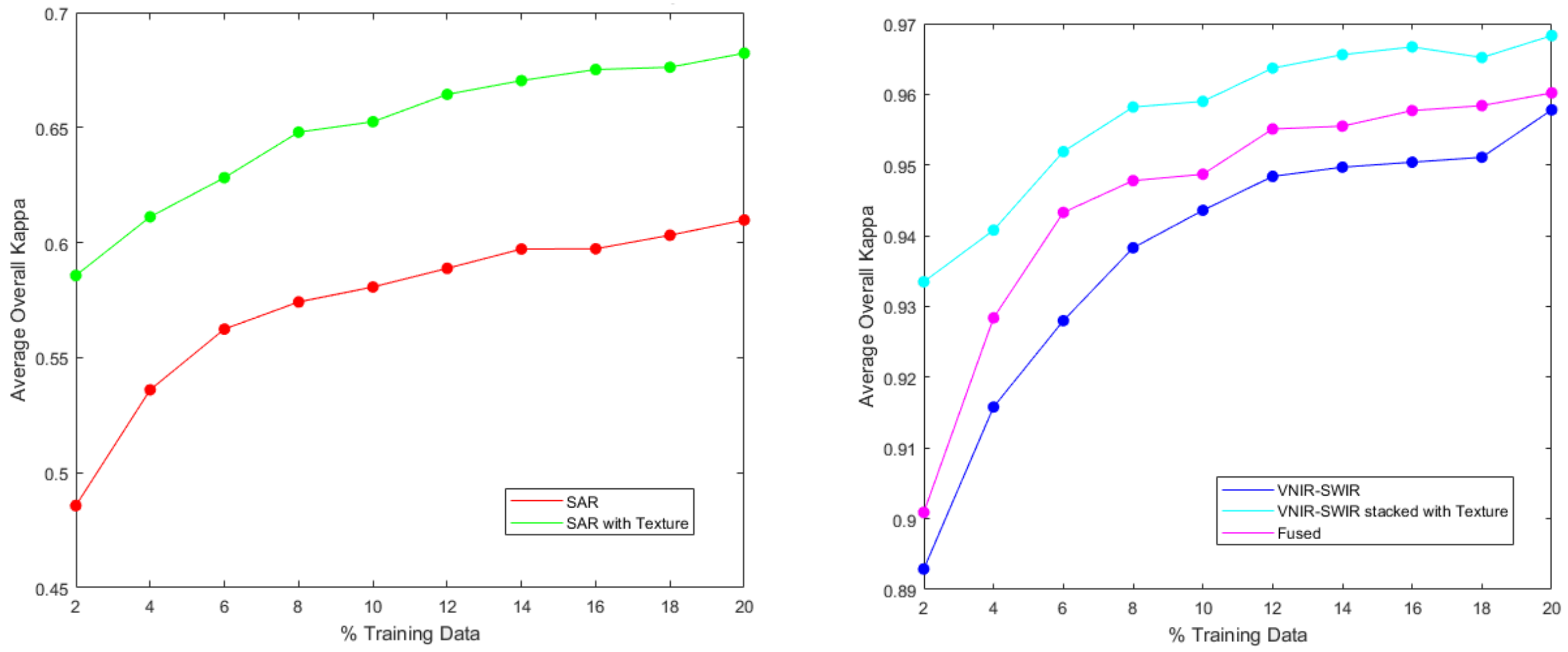

*(a)* *(b)*

*Figure 15 (a): Variation in average overall kappa for SAR bands and SAR bands stacked with texture band for PCA-RFE*
*Figure 15 (b): Variation in average overall kappa for VNIR-SWIR bands, VNIR-SWIR bands stacked with texture and VNIR-SWIR bands fused with texture band for PCA-RFE*

*Table 15: Overall kappa for different datasets for SRP-RFE*

| | **SAR** | **SAR stacked with texture band** | **VNIR-SWIR** | **VNIR-SWIR stacked with texture band** | **VNIR-SWIR fused with texture band** |
|---|---|---|---|---|---|
| **Percentage training data** | *Overall kappa* | | | | |
| **2** | 0.5218 | 0.5890 | 0.8983 | 0.9271 | 0.9135 |
| **4** | 0.5470 | 0.6209 | 0.9227 | 0.9448 | 0.9301 |
| **6** | 0.5763 | 0.6423 | 0.9339 | 0.9546 | 0.9432 |
| **8** | 0.5864 | 0.6493 | 0.9410 | 0.9556 | 0.9480 |
| **10** | 0.5874 | 0.6565 | 0.9435 | 0.9595 | 0.9509 |
| **12** | 0.5993 | 0.6680 | 0.9479 | 0.9626 | 0.9543 |
| **14** | 0.6076 | 0.6651 | 0.9510 | 0.9666 | 0.9566 |
| **16** | 0.6143 | 0.6763 | 0.9540 | 0.9660 | 0.9604 |
| **18** | 0.6109 | 0.6790 | 0.9544 | 0.9671 | 0.9575 |
| **20** | 0.6182 | 0.6825 | 0.9549 | 0.9694 | 0.9605 |

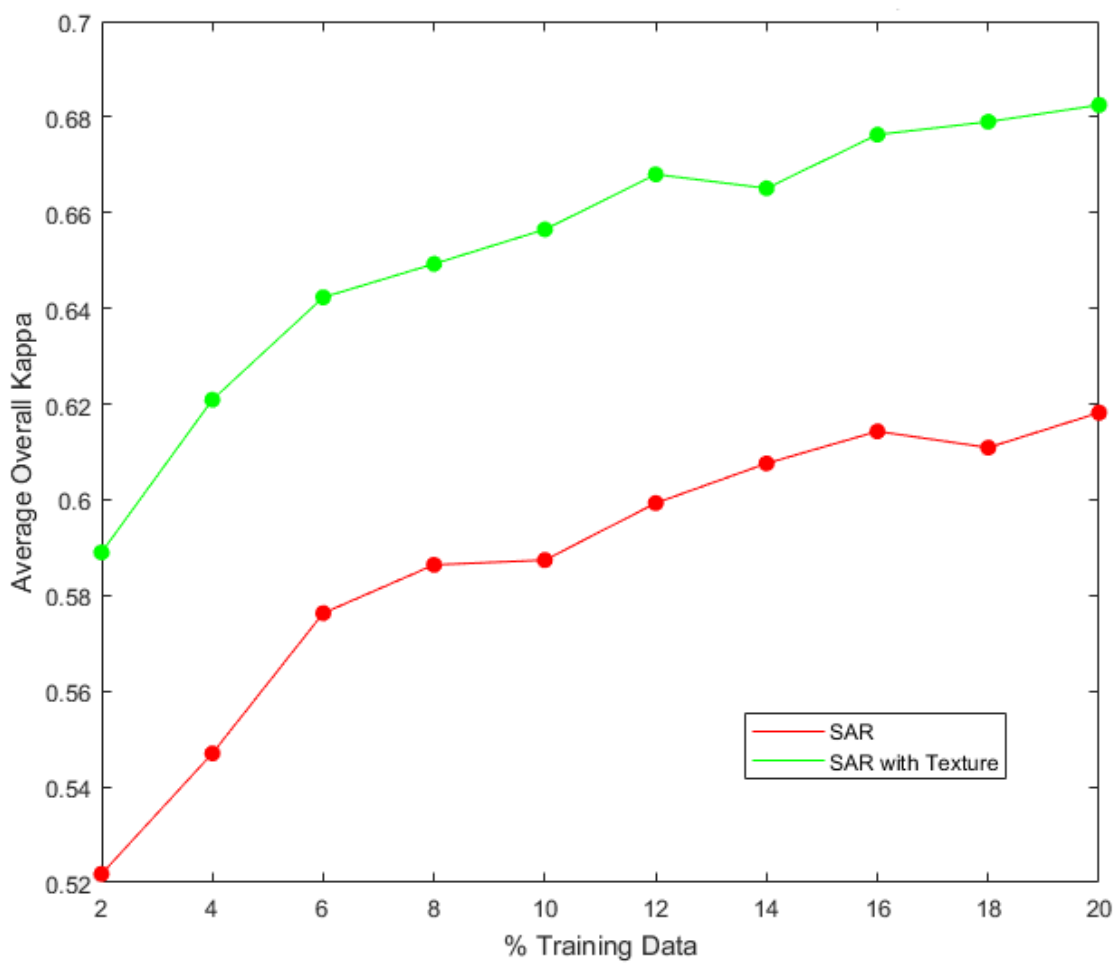

*(a)*

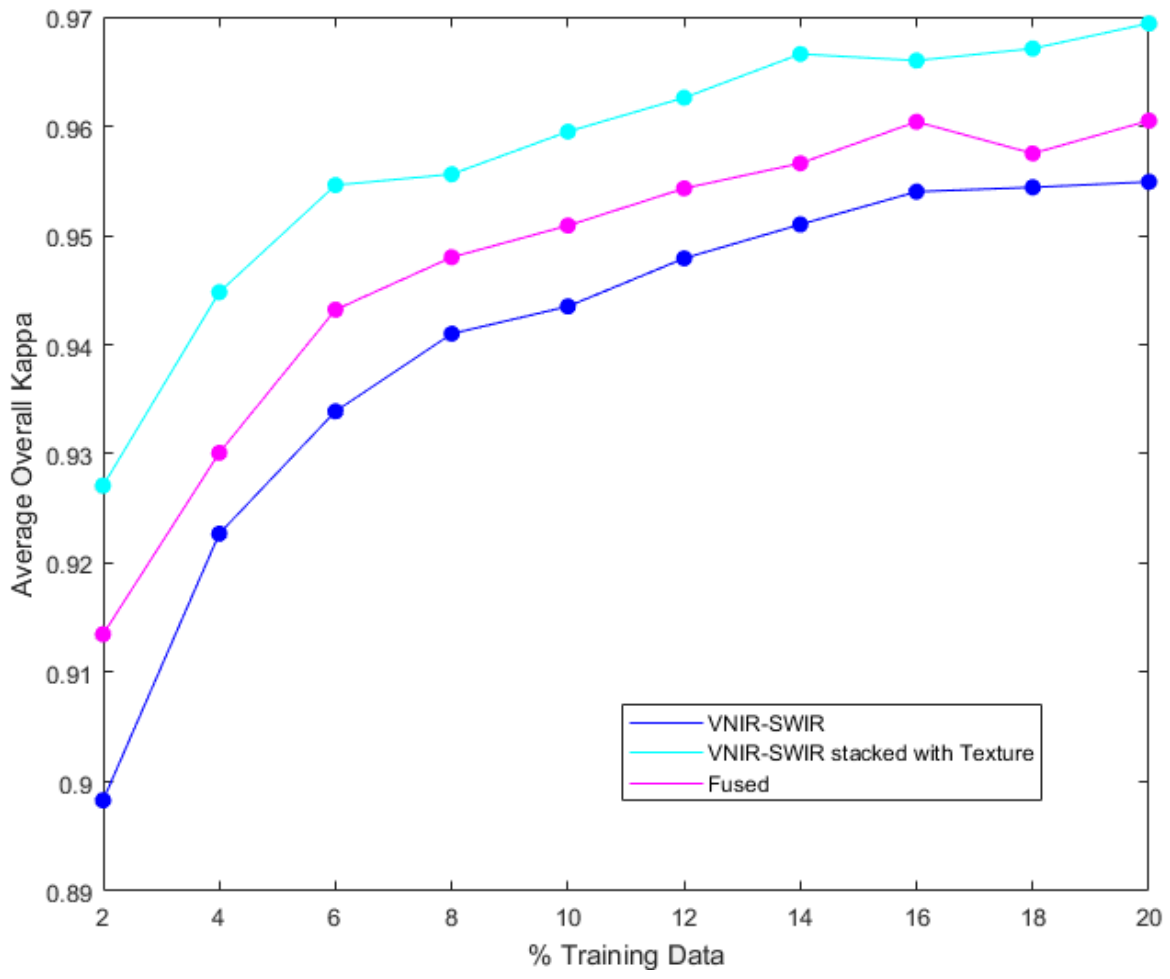

*(b)*

*Figure 16 (a): Variation in average overall kappa for SAR bands and SAR bands stacked with texture band for SRP-RFE*

*Figure 16 (b): Variation in average overall kappa for VNIR-SWIR bands, VNIR-SWIR bands stacked with texture and VNIR-SWIR bands fused with texture band for SRP-RFE*

*Table 16: Overall kappa for different datasets for CRP-RFE*

| | **SAR** | **SAR stacked with texture band** | **VNIR-SWIR** | **VNIR-SWIR stacked with texture band** | **VNIR-SWIR fused with texture band** |
|---|---|---|---|---|---|
| **Percentage training data** | *Overall kappa* | | | | |
| **2** | 0.4888 | 0.5758 | 0.9058 | 0.9280 | 0.9156 |
| **4** | 0.5219 | 0.6145 | 0.9199 | 0.9474 | 0.9350 |
| **6** | 0.5640 | 0.6301 | 0.9343 | 0.9541 | 0.9429 |
| **8** | 0.5694 | 0.6438 | 0.9419 | 0.9578 | 0.9507 |
| **10** | 0.5739 | 0.6446 | 0.9467 | 0.9601 | 0.9510 |
| **12** | 0.5883 | 0.6578 | 0.9523 | 0.9649 | 0.9572 |
| **14** | 0.5944 | 0.6582 | 0.9535 | 0.9649 | 0.9595 |
| **16** | 0.6004 | 0.6648 | 0.9547 | 0.9660 | 0.9587 |
| **18** | 0.6001 | 0.6666 | 0.9544 | 0.9668 | 0.9604 |
| **20** | 0.6061 | 0.6697 | 0.9590 | 0.9710 | 0.9633 |

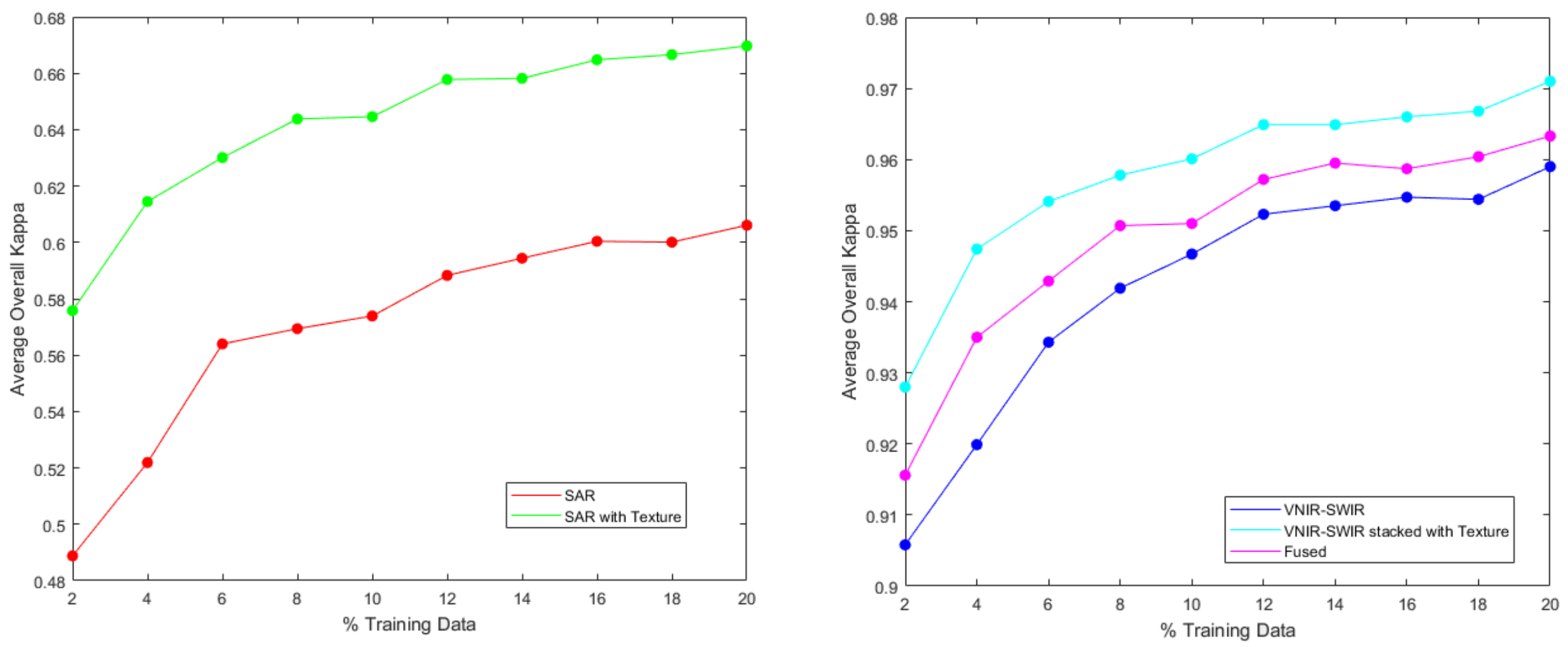

*(a)* *(b)*

*Figure 17 (a): Variation in average overall kappa for SAR bands and SAR bands stacked with texture band for CRP-RFE*

*Figure 17 (b): Variation in average overall kappa for VNIR-SWIR bands, VNIR-SWIR bands stacked with texture and VNIR-SWIR bands fused with texture band for CRP-RFE*

From the results, it is observed that the least classification accuracies are obtained when only SAR dataset is used for classification. This is because the number of features are quite less in SAR bands and also they lack spectral information about the area. When the texture band generated from SAR imagery is stacked with SAR bands, about 4-5% increase in the classification accuracy is observed. This proves that incorporating textural information in classification can increase classification accuracy significantly.

When only VNIR-SWIR bands are used for classification, a significant increase in accuracy is observed in comparison to only SAR bands and even SAR bands incorporated with texture band. This is because more number of features are available for classification as well as spectral information is being used. After texture information (obtained from SAR imagery) is incorporated with VNIR-SWIR bands, it is observed that better classification accuracies are obtained in comparison to using only VNIR-SWIR bands. However, the classification accuracies obtained for VNIR-SWIR stacked with texture band is still more than that obtained from VNIR-SWIR bands fused with texture band. This could be the case because whereas, in case of fusion, a fraction of information from texture band and VNIR-SWIR bands is used, in case of stacking the texture band, complete information from VNIR-SWIR bands as well texture bands are harnessed.

### 5.2.3 Comparison of accuracies for different classifiers on same dataset

In this section, the accuracies for different kinds of classifiers are compared for the same dataset. The value of trees in the decision trees for random forests and ensembles is kept to be 30. The value of *number of forests* in ensembles is fixed at 25. Tables 17- 21 display the overall kappa values for different classifiers on same dataset. In the tables, the highest average overall kappa values are emboldened with black colour. In addition, the red colour is used to highlight the Z-values where the ensembles underperform than the single random forest (i.e. $Z < 0$) whereas, the green colour is used to indicate the values where the ensembles perform significantly better than the single random forest (i.e. $Z > 1.645$).

Figures 18-22 are used to display the trend average overall kappa follows for the classifiers for each of the dataset.

*Table 17: Comparison of overall kappa for different classifiers on SAR bands*

| | **RF** | **PCA-RFE** | | **SRP-RFE** | | **CRP-RFE** | |
|---|---|---|---|---|---|---|---|
| **Percentage training data** | *Overall kappa* | *Overall kappa* | *Z statistic* | *Overall kappa* | *Z statistic* | *Overall kappa* | *Z statistic* |
| **2** | 0.5072 | 0.4942 | -0.8063 | **0.5146** | 0.4610 | 0.4811 | -1.6332 |
| **4** | 0.5504 | 0.5485 | -0.1173 | **0.5570** | 0.4094 | 0.5425 | -0.4900 |
| **6** | 0.5767 | 0.5603 | -1.0084 | **0.5805** | 0.2347 | 0.5592 | -1.0760 |
| **8** | 0.5754 | 0.5697 | -0.3475 | **0.5893** | 0.8510 | 0.5691 | -0.3840 |
| **10** | 0.5870 | 0.5778 | -0.5560 | **0.5970** | 0.6070 | 0.5802 | -0.4110 |
| **12** | 0.5918 | 0.5923 | 0.0300 | **0.5983** | 0.3912 | 0.5872 | -0.2757 |
| **14** | 0.5960 | 0.5961 | 0.0059 | **0.6072** | 0.6627 | 0.5963 | 0.0178 |
| **16** | 0.6011 | 0.5963 | -0.2817 | **0.6092** | 0.4753 | 0.5951 | -0.3521 |
| **18** | 0.6061 | 0.6030 | -0.1804 | **0.6154** | 0.5412 | 0.5977 | -0.4869 |
| **20** | 0.6048 | 0.6079 | 0.1775 | **0.6180** | 0.7588 | 0.6028 | -0.1145 |

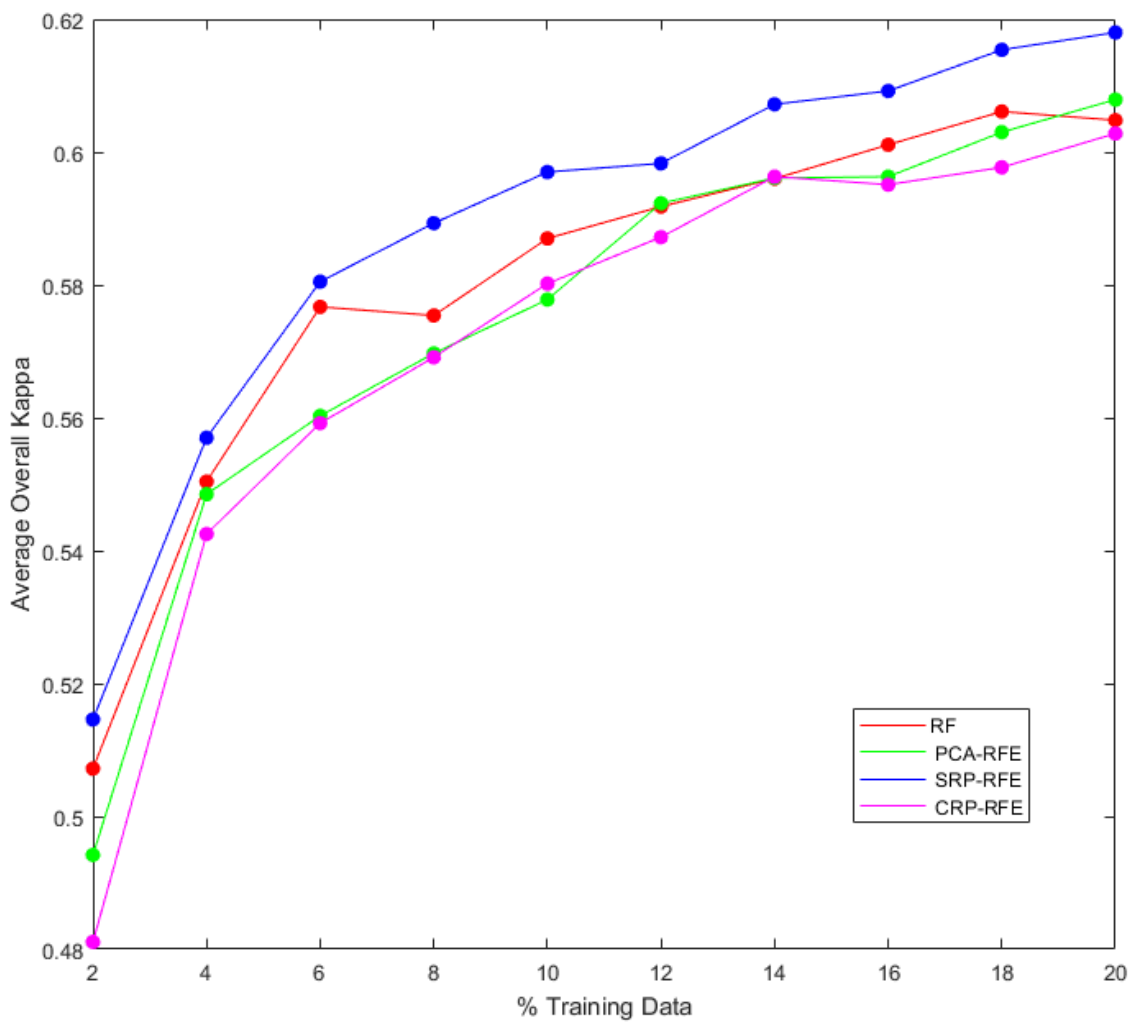

*Figure 18: Comparing average overall kappa for different classifiers on SAR dataset*

*Table 18: Comparison of overall kappa for different classifiers on SAR bands*

| | **RF** | **PCA-RFE** | | **SRP-RFE** | | **CRP-RFE** | |
|---|---|---|---|---|---|---|---|
| **Percentage training data** | *Overall kappa* | *Overall kappa* | *Z statistic* | *Overall kappa* | *Z statistic* | *Overall kappa* | *Z statistic* |
| **2** | 0.5822 | **0.5950** | 0.8154 | 0.5905 | 0.5287 | 0.5842 | 0.1268 |
| **4** | 0.6125 | 0.6101 | -0.1529 | **0.6193** | 0.4351 | 0.6139 | 0.0896 |
| **6** | 0.6229 | 0.6289 | 0.3805 | **0.6399** | 1.0829 | 0.6325 | 0.6115 |
| **8** | 0.6375 | **0.6523** | 0.9386 | 0.6491 | 0.7356 | 0.6423 | 0.3044 |
| **10** | 0.6450 | 0.6521 | 0.4463 | **0.6564** | 0.7165 | 0.6450 | 0.0000 |
| **12** | 0.6551 | **0.6696** | 0.9114 | 0.6660 | 0.6851 | 0.6548 | -0.0188 |
| **14** | 0.6610 | 0.6678 | 0.4236 | **0.6686** | 0.4735 | 0.6610 | 0.0000 |
| **16** | 0.6621 | 0.6734 | 0.6918 | **0.6781** | 0.9838 | 0.6672 | 0.3136 |
| **18** | 0.6694 | **0.6775** | 0.4959 | 0.6761 | 0.4102 | 0.6670 | -0.1463 |
| **20** | 0.6735 | 0.6812 | 0.4634 | **0.6818** | 0.5016 | 0.6696 | -0.2347 |

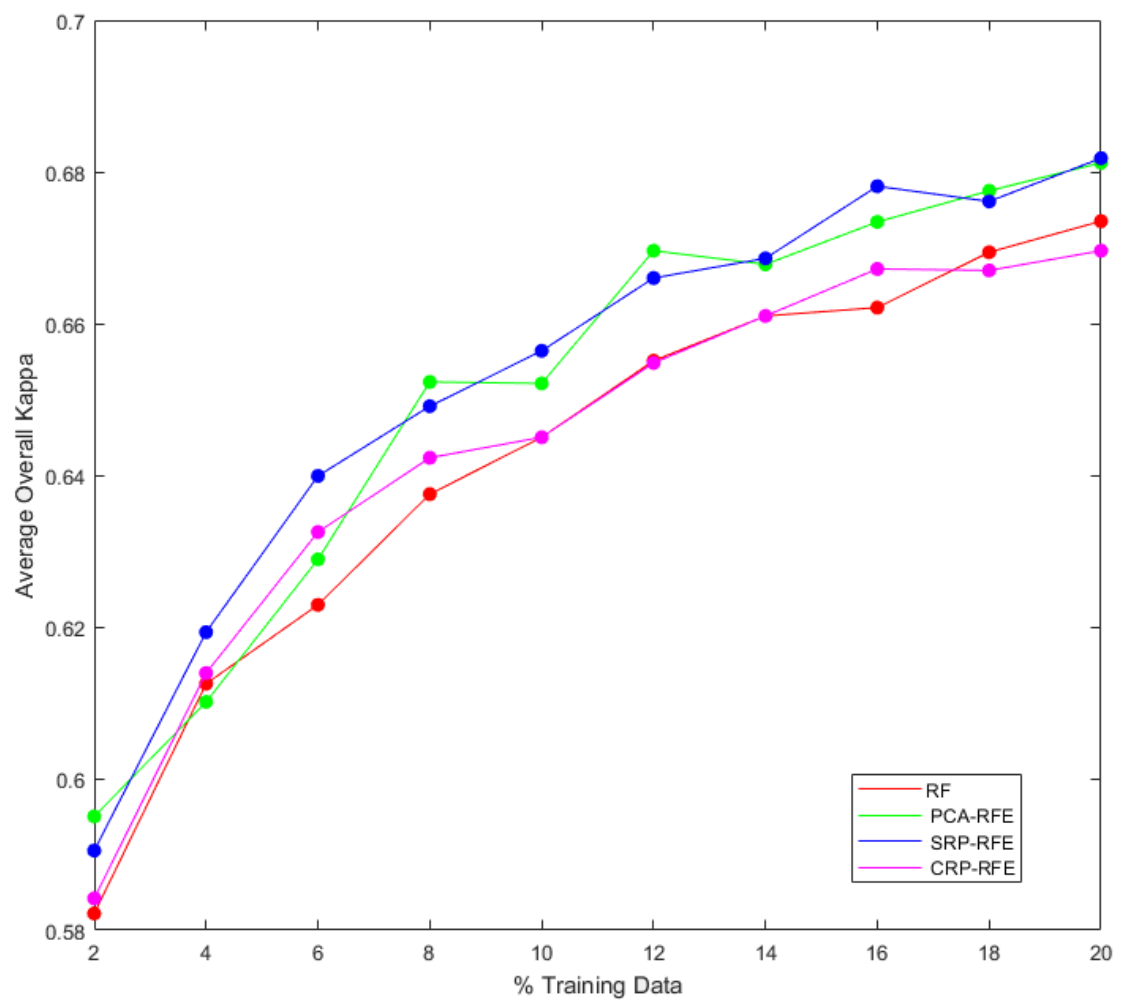

*Figure 19: Comparing average overall kappa for different classifiers on SAR dataset stacked with texture band*

*Table 19: Comparison of overall kappa for different classifiers on VNIR-SWIR bands*

| | RF | PCA-RFE | | SRP-RFE | | CRP-RFE | |
|---|---|---|---|---|---|---|---|
| **Percentage training data** | *Overall kappa* | *Overall kappa* | *Z statistic* | *Overall kappa* | *Z statistic* | *Overall kappa* | *Z statistic* |
| **2** | 0.8628 | 0.8969 | 3.3195 | 0.8962 | 3.2513 | **0.9040** | 4.0625 |
| **4** | 0.9067 | 0.9203 | 1.5133 | **0.9256** | 2.1189 | 0.9217 | 1.6690 |
| **6** | 0.9186 | 0.9299 | 1.3092 | 0.9349 | 1.9186 | **0.9389** | 2.4274 |
| **8** | 0.9221 | **0.9403** | 2.1960 | 0.9399 | 2.1477 | 0.9391 | 2.0512 |
| **10** | 0.9275 | 0.9442 | 2.0499 | 0.9422 | 1.7898 | **0.9469** | 2.4007 |
| **12** | 0.9361 | **0.9500** | 1.7844 | 0.9499 | 1.7716 | 0.9498 | 1.7587 |
| **14** | 0.9384 | 0.9490 | 1.3491 | 0.9501 | 1.5020 | **0.9530** | 1.8904 |
| **16** | 0.9385 | 0.9504 | 1.5146 | 0.9513 | 1.6432 | **0.9550** | 2.1547 |
| **18** | 0.9408 | 0.9530 | 1.5662 | 0.9540 | 1.6945 | **0.9562** | 2.0111 |
| **20** | 0.9460 | 0.9552 | 1.2039 | 0.9570 | 1.4523 | **0.9599** | 1.8680 |

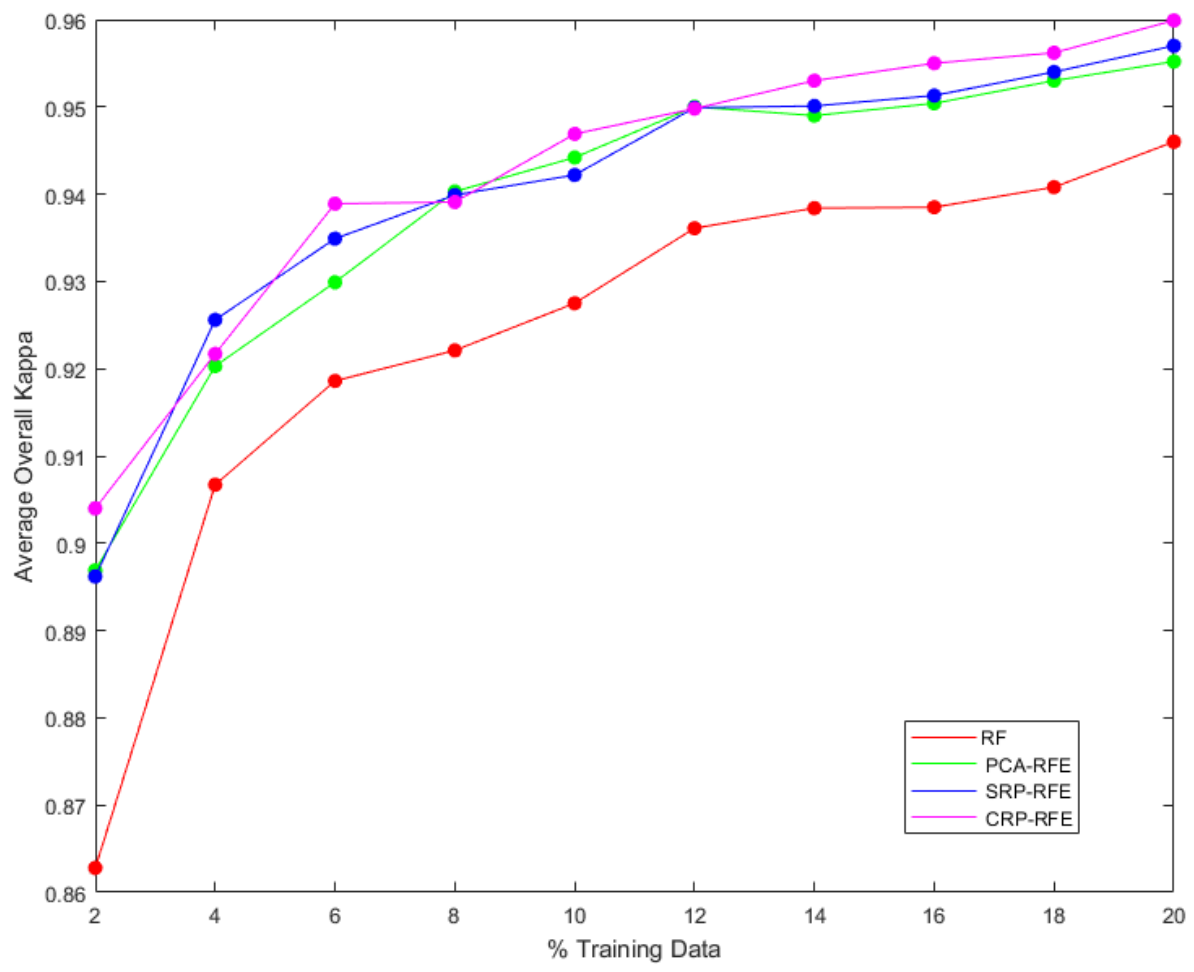

*Figure 20: Comparing average overall kappa for different classifiers on VNIR-SWIR dataset*

*Table 20: Comparison of overall kappa for different classifiers on VNIR-SWIR bands stacked with texture band*

| | RF | PCA-RFE | | SRP-RFE | | CRP-RFE | |
|---|---|---|---|---|---|---|---|
| **Percentage training data** | *Overall kappa* | *Overall kappa* | *Z statistic* | *Overall kappa* | *Z statistic* | *Overall kappa* | *Z statistic* |
| **2** | 0.8973 | 0.9232 | 2.8560 | **0.9280** | 3.4349 | 0.9256 | 3.1435 |
| **4** | 0.9191 | 0.9463 | 3.3881 | 0.9416 | 2.7585 | **0.9482** | 3.6535 |
| **6** | 0.9356 | **0.9556** | 2.7356 | 0.9543 | 2.5354 | 0.9542 | 2.5218 |
| **8** | 0.9379 | 0.9561 | 2.4894 | **0.9579** | 2.7356 | **0.9579** | 2.7356 |
| **10** | 0.9457 | 0.9594 | 1.9522 | 0.9594 | 1.9522 | **0.9611** | 2.2150 |
| **12** | 0.9487 | 0.9636 | 2.1667 | 0.9635 | 2.1727 | **0.9644** | 2.3048 |
| **14** | 0.9512 | **0.9665** | 2.2675 | 0.9634 | 1.7741 | 0.9642 | 1.9085 |
| **16** | 0.9523 | 0.9650 | 1.8468 | 0.9634 | 1.5988 | **0.9664** | 2.0700 |
| **18** | 0.9554 | 0.9659 | 1.5438 | 0.9649 | 1.3968 | **0.9670** | 1.7222 |
| **20** | 0.9577 | 0.9678 | 1.5164 | 0.9664 | 1.2933 | **0.9693** | 1.7590 |

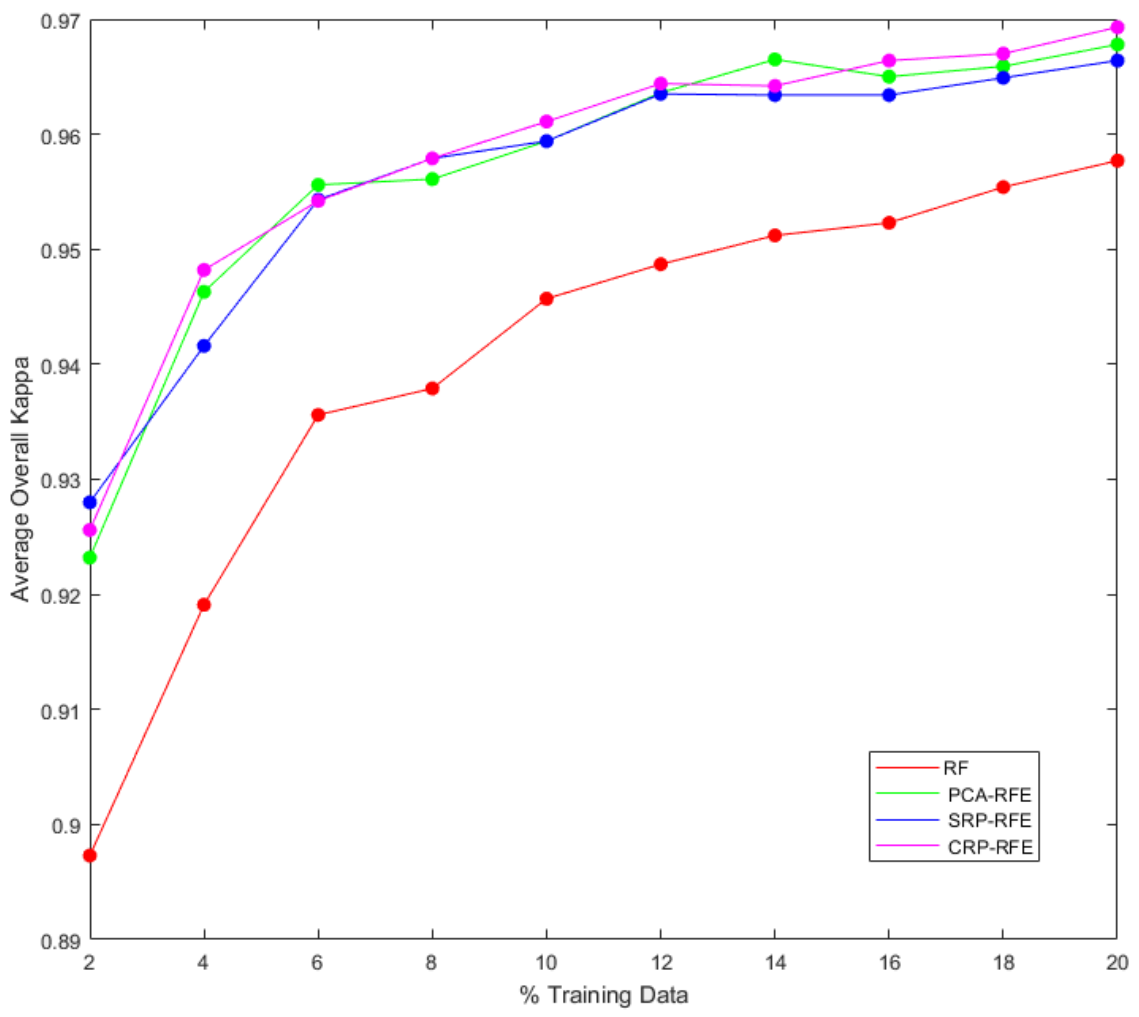

*Figure 21: Comparing average overall kappa for different classifiers on VNIR-SWIR dataset stacked with texture band*

*Table 21: Comparison of overall kappa for different classifiers on VNIR-SWIR bands fused with texture band*

| | **RF** | **PCA-RFE** | | **SRP-RFE** | | **CRP-RFE** | |
|---|---|---|---|---|---|---|---|
| **Percentage training data** | *Overall kappa* | *Overall kappa* | *Z statistic* | *Overall kappa* | *Z statistic* | *Overall kappa* | *Z statistic* |
| **2** | 0.8714 | 0.9055 | 3.4132 | 0.9056 | 3.4459 | **0.9146** | 4.4105 |
| **4** | 0.9038 | 0.9272 | 2.6211 | **0.9324** | 3.2513 | 0.9318 | 3.1831 |
| **6** | 0.9218 | **0.9433** | 2.6359 | 0.9384 | 2.0029 | 0.9421 | 2.4690 |
| **8** | 0.9291 | 0.9437 | 1.8409 | 0.9445 | 1.9418 | **0.9494** | 2.6030 |
| **10** | 0.9309 | 0.9508 | 2.5517 | 0.9499 | 2.4363 | **0.9517** | 2.6895 |
| **12** | 0.9422 | 0.9540 | 1.5875 | **0.9578** | 2.1370 | 0.9561 | 1.8870 |
| **14** | 0.9455 | 0.9559 | 1.4119 | 0.9573 | 1.6164 | **0.9574** | 1.6301 |
| **16** | 0.9457 | 0.9569 | 1.5205 | 0.9583 | 1.7105 | **0.9621** | 2.2873 |
| **18** | 0.9484 | 0.9570 | 1.1686 | 0.9601 | 1.6194 | **0.9619** | 1.8858 |
| **20** | 0.9481 | 0.9599 | 1.6019 | 0.9609 | 1.7534 | **0.9630** | 2.0595 |

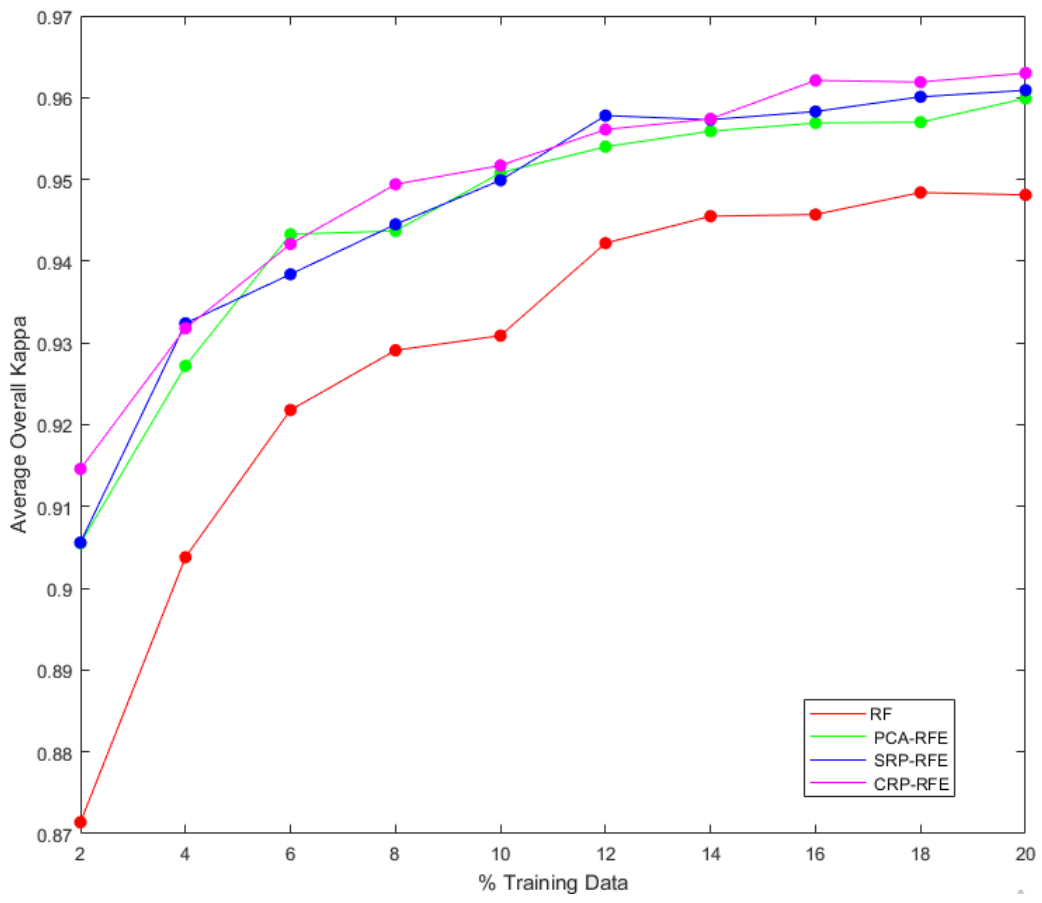

*Figure 22: Comparing average overall kappa for different classifiers on VNIR-SWIR dataset fused with texture band*

Here, it is observed that all the red coloured Z-values occur in the case of SAR bands and even SAR bands stacked with texture band i.e. the performance of random forest ensembles is worse than that of a single random forest classifier in those cases. But, in none of the cases, the random forest ensembles perform significantly worse than the single random forest since Z-value is never less than -1.645.

All the green coloured Z-values occur in case of VNIR-SWIR bands (both with and without texture information). This means that the random forest ensembles are performing significantly better than a single random forest in most of the cases. We also observe that the CRP-RFE perform the best among all the three ensembles for VNIR-SWIR (with and without texture) as the maximum number of green coloured Z-values are obtained for them along with maximum number of emboldened average overall kappa. The SRP-RFE stand on the second position while the PCA-RFE stand on the third.

However, in case of SAR and SAR with texture, CRP-RFE perform worst among all the three ensembles as maximum red values occur in it. The PCA-RFE perform the second worst while SRP-RFE performs better than the other two as there is no red Z-values in it and has maximum number of emboldened average overall kappa.

The reason that the ensembles do not perform better than the single random forest in first two cases could be that both datasets contain derived features that are already combinations of the original VV and VH bands (section 3.2).

It is also seen that the Z-values are higher (specifically in last 3 cases) for less number of training pixels. This means that the ensembles are performing much better than the single random forest for less number of training pixels and as the number of training pixels increase, the performance of both of them tends to become similar.

### 5.2.4 Comparison of average producer kappa

Tables 22-26 show the values of producer kappa ($\hat{\kappa}_{+j}$) and their standard deviation ($\sigma_{\hat{\kappa}_{+j}}$) for each class for all classifiers and datasets and figures 23-27 show the bar graphs for the producer kappa for each class for all classifiers and datasets. 10% training data has been used for this analysis because ensembles are performing much better than single random forest at less training data.

#### *1. SAR bands*

*Table 22: Producer kappa and its standard deviation for SAR bands for all classifiers*

| Classes | **RF** | | **PCA-RFE** | | **SRP-RFE** | | **CRP-RFE** | |
|---|---|---|---|---|---|---|---|---|
| | $\hat{\kappa}_{+j}$ | $\sigma_{\hat{\kappa}_{+j}}$ | $\hat{\kappa}_{+j}$ | $\sigma_{\hat{\kappa}_{+j}}$ | $\hat{\kappa}_{+j}$ | $\sigma_{\hat{\kappa}_{+j}}$ | $\hat{\kappa}_{+j}$ | $\sigma_{\hat{\kappa}_{+j}}$ |
| Water | 0.8337 | 0.0306 | 0.7697 | 0.0350 | 0.7734 | 0.0345 | 0.7191 | 0.0371 |
| Sand | 0.3935 | 0.0506 | 0.3865 | 0.0500 | 0.4003 | 0.0506 | 0.3615 | 0.0498 |
| Built-up area | 0.6219 | 0.0208 | 0.6344 | 0.0207 | 0.6280 | 0.0204 | 0.6378 | 0.0205 |
| Trees | 0.4076 | 0.0352 | 0.4210 | 0.0358 | 0.4759 | 0.0379 | 0.4766 | 0.0383 |
| Grass | 0.5504 | 0.0233 | 0.5352 | 0.0232 | 0.5653 | 0.0232 | 0.5536 | 0.0235 |
| Algae | 0.4310 | 0.0343 | 0.3992 | 0.0971 | 0.5097 | 0.0447 | 0.4342 | 0.0427 |
| Unused land | 0.6567 | 0.0242 | 0.6314 | 0.0243 | 0.6448 | 0.0240 | 0.5980 | 0.0240 |

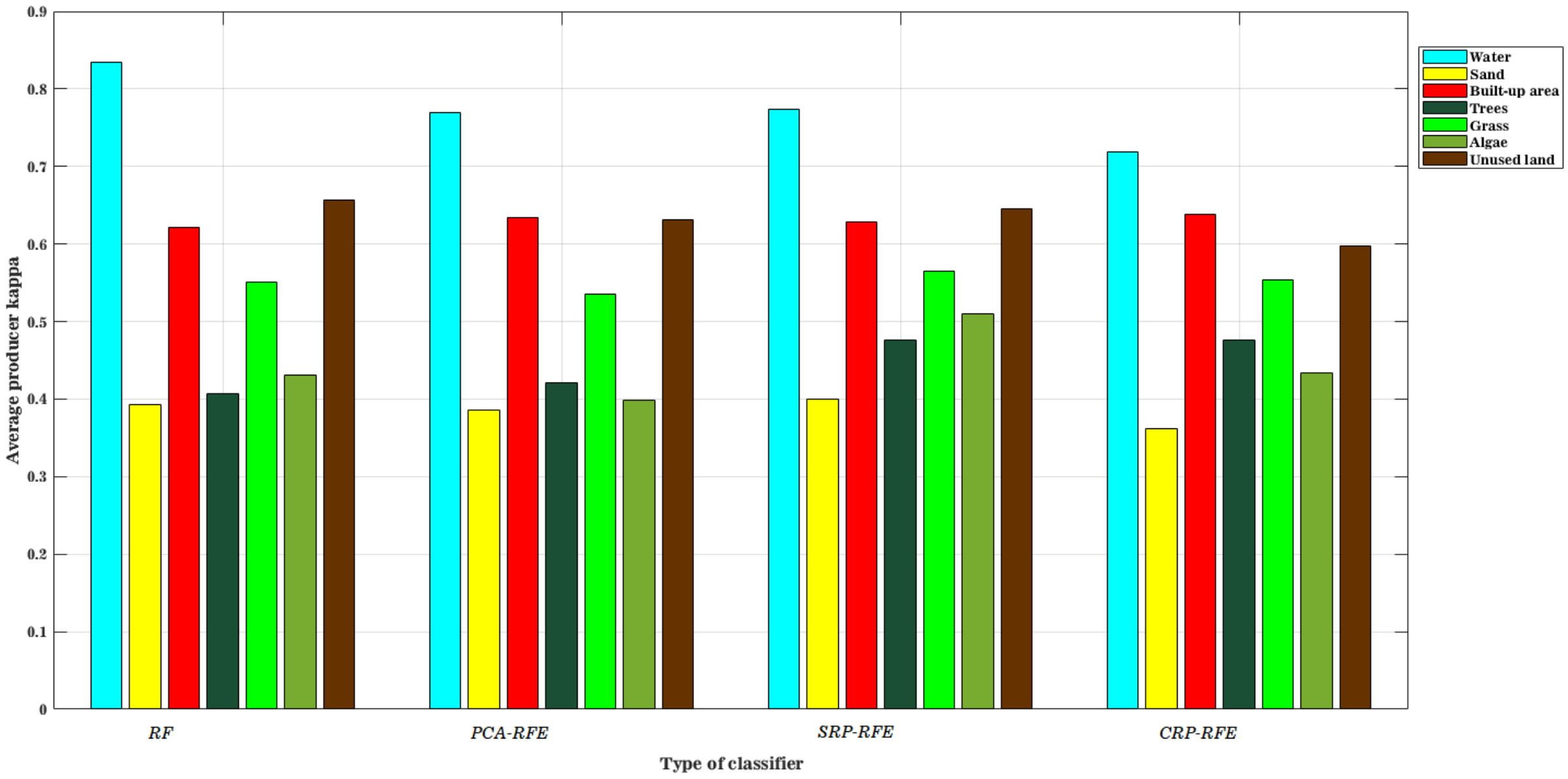

*Figure 23: Producer kappa for SAR bands for all classifiers*

## 2. *SAR bands stacked with texture band*

*Table 23: Producer kappa and its standard deviation for SAR bands stacked with texture band for all classifiers*

| Classes | RF | | PCA-RFE | | SRP-RFE | | CRP-RFE | |
|---|---|---|---|---|---|---|---|---|
| | $\hat{\kappa}_{+j}$ | $\sigma_{\hat{\kappa}_{+j}}$ | $\hat{\kappa}_{+j}$ | $\sigma_{\hat{\kappa}_{+j}}$ | $\hat{\kappa}_{+j}$ | $\sigma_{\hat{\kappa}_{+j}}$ | $\hat{\kappa}_{+j}$ | $\sigma_{\hat{\kappa}_{+j}}$ |
| Water | 0.8403 | 0.0292 | 0.7892 | 0.0329 | 0.8074 | 0.0332 | 0.7451 | 0.0367 |
| Sand | 0.4794 | 0.0538 | 0.4710 | 0.0552 | 0.4667 | 0.0551 | 0.4414 | 0.0571 |
| Built-up area | 0.7076 | 0.0196 | 0.7087 | 0.0195 | 0.7082 | 0.0193 | 0.7085 | 0.0192 |
| Trees | 0.4310 | 0.0344 | 0.4418 | 0.0345 | 0.4882 | 0.0364 | 0.5000 | 0.0371 |
| Grass | 0.6165 | 0.0223 | 0.6333 | 0.0219 | 0.6265 | 0.0216 | 0.6186 | 0.0217 |
| Algae | 0.1153 | 0.0295 | 0.1389 | 0.0145 | 0.4051 | 0.0362 | 0.4779 | 0.0004 |
| Unused land | 0.7041 | 0.0235 | 0.7225 | 0.0231 | 0.7162 | 0.0233 | 0.6842 | 0.0239 |

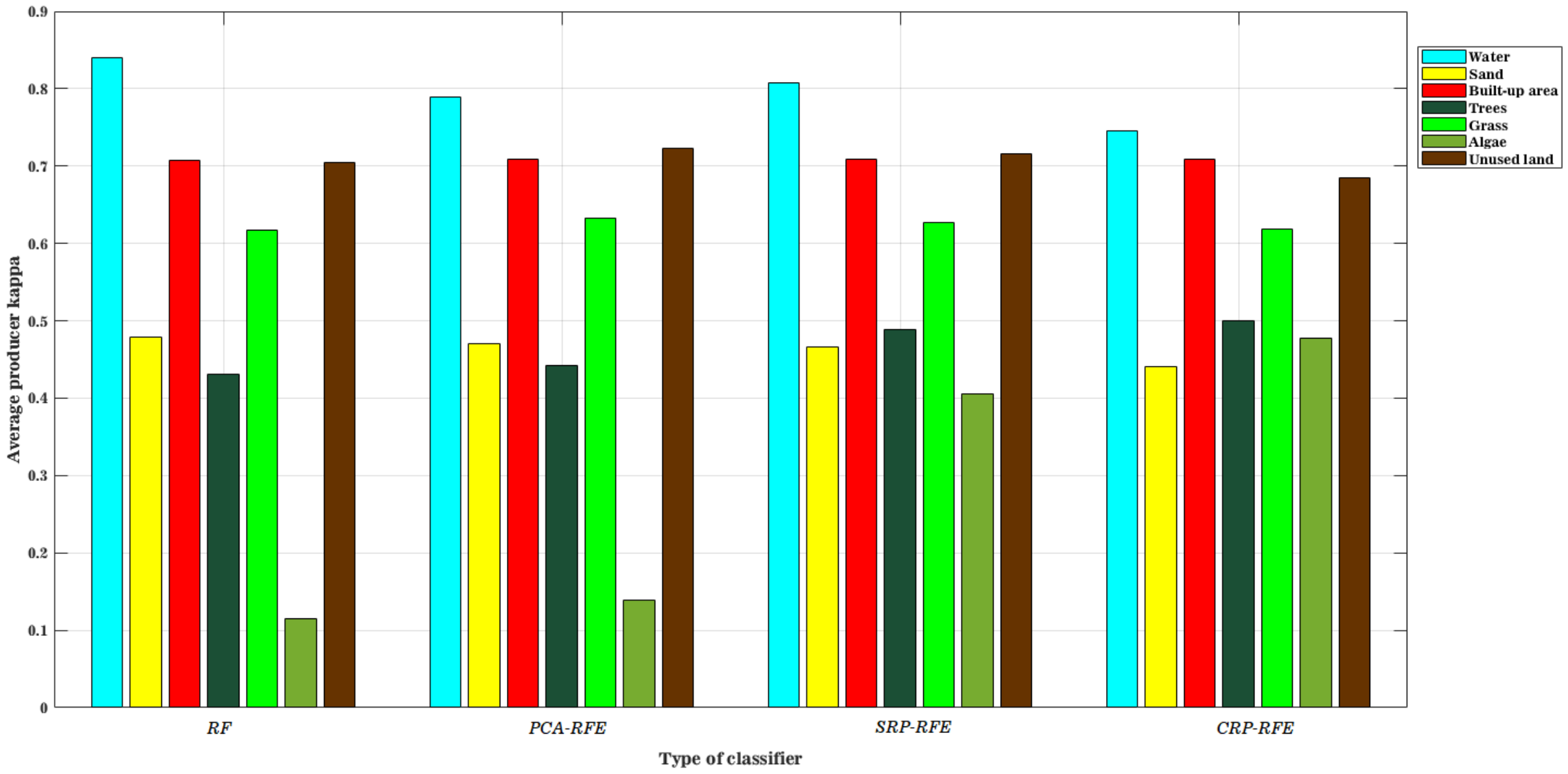

*Figure 24: Producer kappa for SAR bands stacked with texture band for all classifiers*

## 3. *VNIR-SWIR bands*

*Table 24: Producer kappa and its standard deviation for VNIR-SWIR bands for all classifiers*

| Classes | RF | | PCA-RFE | | SRP-RFE | | CRP-RFE | |
|---|---|---|---|---|---|---|---|---|
| | $\hat{\kappa}_{+j}$ | $\sigma_{\hat{\kappa}_{+j}}$ | $\hat{\kappa}_{+j}$ | $\sigma_{\hat{\kappa}_{+j}}$ | $\hat{\kappa}_{+j}$ | $\sigma_{\hat{\kappa}_{+j}}$ | $\hat{\kappa}_{+j}$ | $\sigma_{\hat{\kappa}_{+j}}$ |
| Water | 0.9875 | 0.0081 | 0.9731 | 0.0126 | 0.9765 | 0.0116 | 0.9760 | 0.0119 |
| Sand | 0.8206 | 0.0337 | 0.8834 | 0.0287 | 0.8834 | 0.0294 | 0.8825 | 0.0288 |
| Built-up area | 0.9661 | 0.0080 | 0.9668 | 0.0080 | 0.9637 | 0.0083 | 0.9687 | 0.0080 |
| Trees | 0.8784 | 0.0188 | 0.8953 | 0.0175 | 0.8935 | 0.0177 | 0.9065 | 0.0169 |
| Grass | 0.9420 | 0.0113 | 0.9600 | 0.0096 | 0.9554 | 0.0100 | 0.9550 | 0.0100 |
| Algae | 0.6966 | 0.1311 | 0.8054 | 0.0720 | 0.8471 | 0.0769 | 0.7698 | 0.1178 |
| Unused land | 0.9278 | 0.0131 | 0.9515 | 0.0109 | 0.9509 | 0.0109 | 0.9579 | 0.0102 |

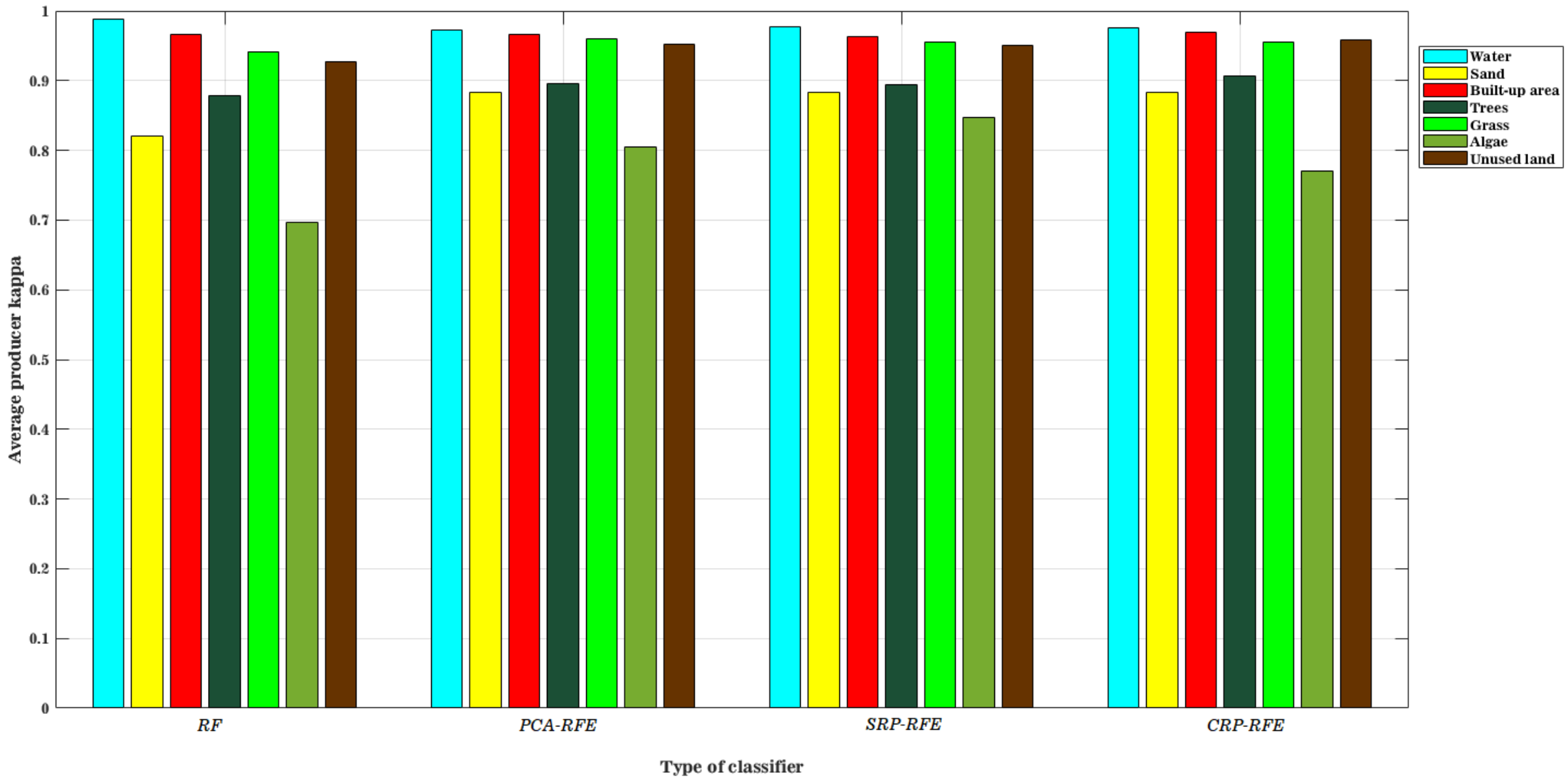

*Figure 25: Producer kappa for VNIR-SWIR bands for all classifiers*

## *4. VNIR-SWIR bands stacked with texture band*

*Table 25: Producer kappa and its standard deviation for VNIR-SWIR bands stacked with texture band for all classifiers*

| Classes | **RF** | | **PCA-RFE** | | **SRP-RFE** | | **CRP-RFE** | |
|---|---|---|---|---|---|---|---|---|
| | $\hat{\kappa}_{+j}$ | $\sigma_{\hat{\kappa}_{+j}}$ | $\hat{\kappa}_{+j}$ | $\sigma_{\hat{\kappa}_{+j}}$ | $\hat{\kappa}_{+j}$ | $\sigma_{\hat{\kappa}_{+j}}$ | $\hat{\kappa}_{+j}$ | $\sigma_{\hat{\kappa}_{+j}}$ |
| Water | 0.9808 | 0.0105 | 0.9773 | 0.0117 | 0.9813 | 0.0099 | 0.9760 | 0.0121 |
| Sand | 0.9075 | 0.0258 | 0.9505 | 0.0189 | 0.9663 | 0.0159 | 0.9623 | 0.0172 |
| Built-up area | 0.9816 | 0.0060 | 0.9840 | 0.0057 | 0.9864 | 0.0053 | 0.9851 | 0.0055 |
| Trees | 0.8918 | 0.0179 | 0.9109 | 0.0165 | 0.9082 | 0.0167 | 0.9149 | 0.0163 |
| Grass | 0.9508 | 0.0104 | 0.9552 | 0.0100 | 0.9554 | 0.0099 | 0.9544 | 0.0101 |
| Algae | 0.7230 | 0.1006 | 0.7789 | 0.1313 | 0.8424 | 0.0894 | 0.8098 | 0.0741 |
| Unused land | 0.9450 | 0.0116 | 0.9720 | 0.0083 | 0.9623 | 0.0097 | 0.9709 | 0.0085 |

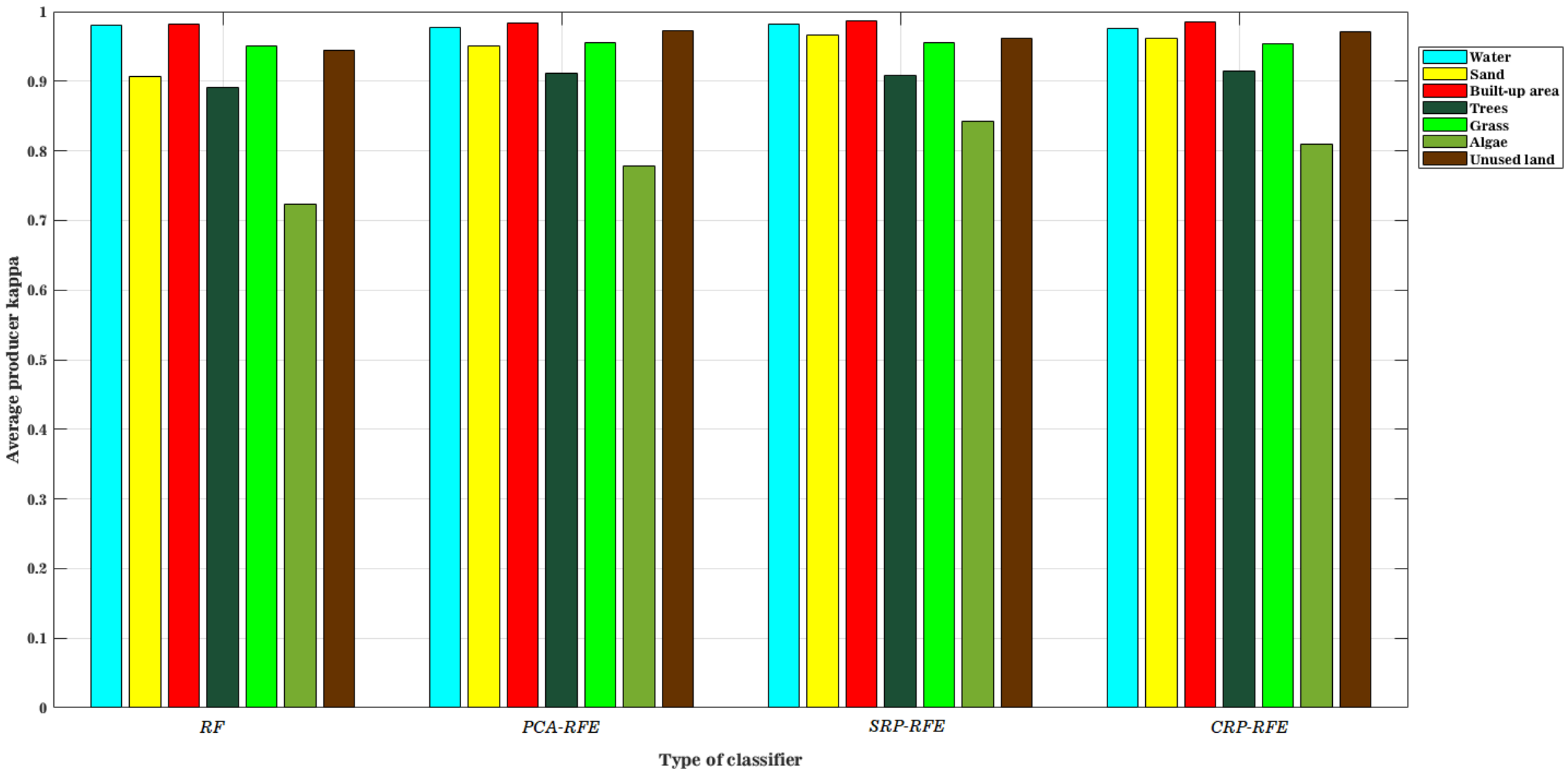

*Figure 26: Producer kappa for VNIR-SWIR bands stacked with texture band for all classifiers*

## 5. VNIR-SWIR bands fused with texture band

*Table 26: Producer kappa and its standard deviation for VNIR-SWIR bands fused with texture band for all classifiers*

| Classes | RF | | PCA-RFE | | SRP-RFE | | CRP-RFE | |
|---|---|---|---|---|---|---|---|---|
| | $\hat{\kappa}_{+j}$ | $\sigma_{\hat{\kappa}_{+j}}$ | $\hat{\kappa}_{+j}$ | $\sigma_{\hat{\kappa}_{+j}}$ | $\hat{\kappa}_{+j}$ | $\sigma_{\hat{\kappa}_{+j}}$ | $\hat{\kappa}_{+j}$ | $\sigma_{\hat{\kappa}_{+j}}$ |
| Water | 0.9838 | 0.0090 | 0.9766 | 0.0116 | 0.9808 | 0.0109 | 0.9808 | 0.0101 |
| Sand | 0.8531 | 0.0317 | 0.9048 | 0.0263 | 0.8988 | 0.0272 | 0.9201 | 0.0251 |
| Built-up area | 0.9593 | 0.0086 | 0.9713 | 0.0072 | 0.9762 | 0.0068 | 0.9731 | 0.0072 |
| Trees | 0.8834 | 0.0185 | 0.9034 | 0.0171 | 0.9024 | 0.0173 | 0.9007 | 0.0172 |
| Grass | 0.9382 | 0.0117 | 0.9572 | 0.0098 | 0.9528 | 0.0102 | 0.9589 | 0.0095 |
| Algae | 0.7045 | 0.1252 | 0.7159 | 0.1380 | 0.8399 | 0.092 | 0.8128 | 0.0785 |
| Unused land | 0.9393 | 0.0122 | 0.9677 | 0.0090 | 0.9601 | 0.0099 | 0.9611 | 0.0098 |

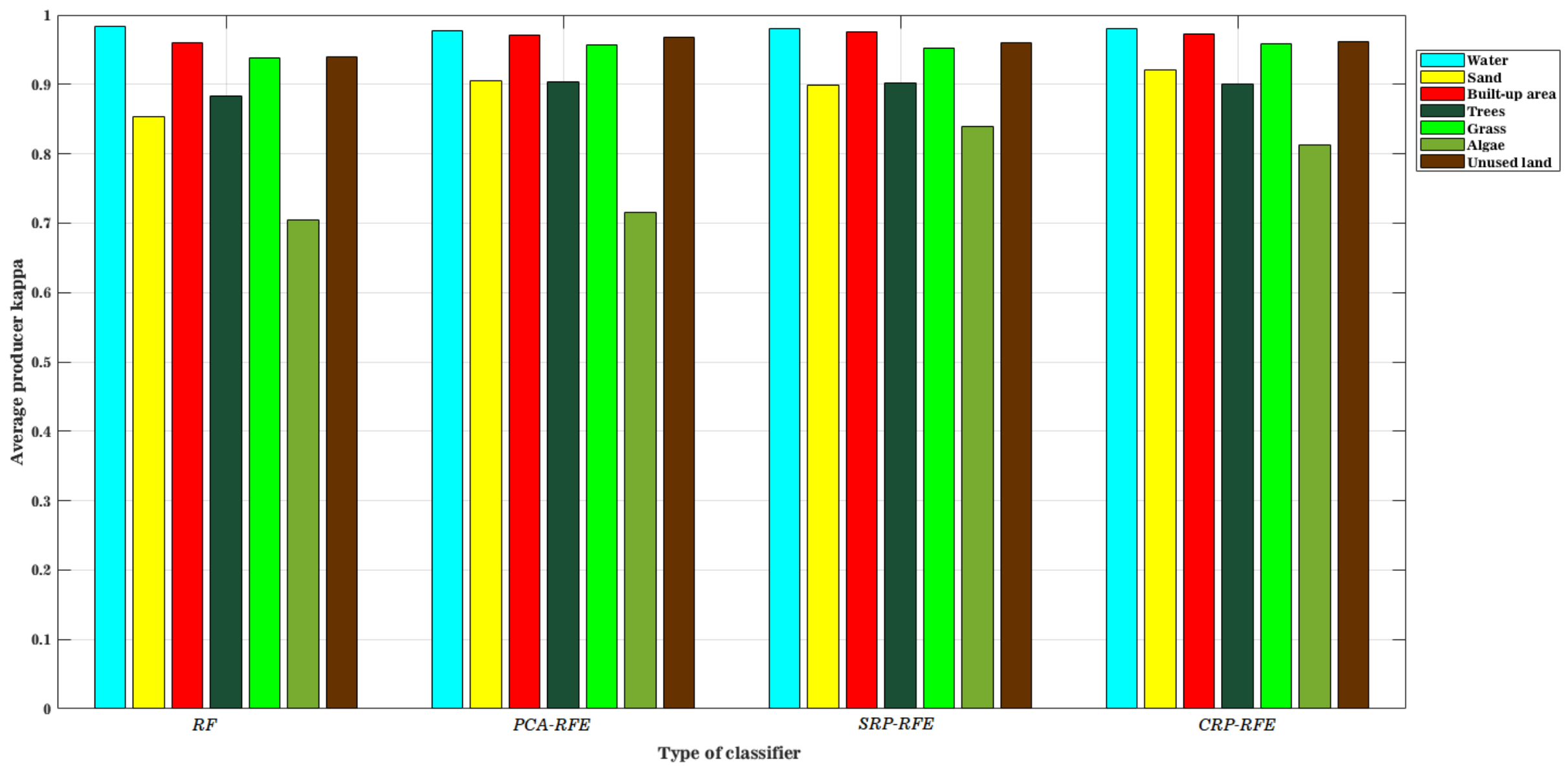

*Figure 27: Producer kappa for VNIR-SWIR bands fused with texture band for all classifiers*

Following inferences can be drawn after observing the average producer kappa for all the aforementioned cases:

a. In case of SAR imagery, for the *water* class, $\hat{\kappa}_{+j}$ decreases for the ensembles as compared to a single random forest. For the *sand*, *built-up area* and *grass* classes, the performance seems similar for all the classifiers. For the *trees* significant improvement is seen for SRP-RFE and CRP-RFE over single RF. For class *algae*, improvement is observed for SRP-RFE while for the class *unused land*, the performance decreases in case of CRP-RFE.

b. For SAR imagery stacked with texture band, there is no improvement in accuracy of ensembles for *water* class but the accuracy decreases for CRP-RFE. Similar performances of the classifiers are observed for classes *sand*, *built-up area*, *grass* and *unused land*. The RF and PCA-RFE classifiers have underperformed for class *algae* however, improvements are observed in SRP-RFE and CRP-RFE.

c. In case of VNIR-SWIR imagery, classes *water*, *built-up area*, *grass* and *trees*, perform similar for all the classifiers. However, ensembles have shown to perform better for classes *sand*, *algae* and *unused land*.

d. For VNIR-SWIR imagery stacked with texture band, classes *water, built-up area*, *grass* and *trees*, perform similar for all the classifiers. However, ensembles have shown to perform better for classes *sand, algae* and *unused land*.

e. For VNIR-SWIR bands fused with texture band, accuracies for classes *sand, algae* and *unused land* have increased greatly for the ensembles however only minor improvements are observed for the remaining classes.

The classes where major improvements are recorded had much less training pixels that the other classes. This shows that the ensembles work well with limited number of training samples for same number of features.

### 5.2.5 Comparison of execution time for different classifiers

The average execution time calculated for classifying VNIR-SWIR stacked with texture band at 20% training data using all four classifiers has been tabulated in table 27.

*Table 27: Average execution time for all the classifiers*

| Classifier | Mean execution time (seconds) | Standard deviation for execution time (seconds) |
|---|---|---|
| RF | 0.1894 | 0.0142 |
| PCA-RFE | 2.0935 | 0.0373 |
| SRP-RFE | 2.0683 | 0.0330 |
| CRP-RFE | 2.0913 | 0.0360 |

Since, the ensembles are made of random forests themselves, it is expected that the execution time taken by the ensembles would be:

$$ET_C = C \text{ x } t_1 + t_2 \qquad (5.1)$$

Here,

$ET_C$ : Execution time taken by an ensemble of *C* base classifiers (RF in this case)

*C* : Numbers of random forests in ensemble

$t_1$ : Execution time of single random forest

$t_2$ : Other programming overheads

Using this, one would calculate the execution time of an ensemble of 30 base classifiers using $t_1 = 0.1894$ from table 27 and $t_2 = 0$ (an assumption since this is not practically viable) as:

$$ET_{30} = 30 \text{ x } 0.1894 + 0$$

$$ET_{30} = 5.6820 \text{ s}$$

However, since parallel processing has been used in the random forest ensembles, therefore execution time is much less than the calculated value.

### 5.2.6 Comparison of ensembles for different parameters

In this section, different parameters of the same classification ensemble are varied and then the corresponding accuracies are compared. The value of *number of forests* is varied progressively in the difference of one classifier from 1-10 (since accuracy was changing rapidly in this case) then in the difference of 2 and 5 upto 30 (since the stagnation in accuracy was starting to creep in) and finally 40 and 50 to ensure that stagnation in accuracy has been achieved. The overall kappa is calculated for all the five datasets for each ensemble in tables 28-30 while the graphical representation is shown in figures 28-30.

### 5.2.6.1 *Ensemble using PCA rotation*

*Table 28: Average overall kappa for PCA-RFE for all datasets*

| Number of trees | SAR | SAR+Texture | VNIR-SWIR | VNIR-SWIR+Texture | Fused |
|---|---|---|---|---|---|
| **1** | 0.5910 | 0.6624 | 0.9496 | 0.9644 | 0.9538 |
| **2** | 0.5933 | 0.6686 | 0.9532 | 0.9654 | 0.9576 |
| **3** | 0.5974 | 0.6766 | 0.9537 | 0.9649 | 0.9591 |
| **4** | 0.6015 | 0.6737 | 0.9547 | 0.9672 | 0.9603 |
| **5** | 0.6028 | 0.6763 | 0.9543 | 0.9668 | 0.9581 |
| **6** | 0.6064 | 0.6796 | 0.9551 | 0.9682 | 0.9599 |
| **7** | 0.6056 | 0.6817 | 0.9542 | 0.9681 | 0.9590 |
| **8** | 0.6037 | 0.6807 | 0.9560 | 0.9687 | 0.9595 |
| **9** | 0.6064 | 0.6822 | 0.9559 | 0.9674 | 0.9600 |
| **10** | 0.6053 | 0.6854 | 0.9554 | 0.9683 | 0.9602 |
| **12** | 0.6033 | 0.6762 | 0.9559 | 0.9698 | 0.9613 |
| **14** | 0.6057 | 0.6814 | 0.9545 | 0.9684 | 0.9612 |
| **15** | 0.6045 | 0.6772 | 0.9541 | 0.9673 | 0.9605 |
| **16** | 0.6120 | 0.6815 | 0.9558 | 0.9698 | 0.9602 |
| **18** | 0.6075 | 0.6802 | 0.9548 | 0.9688 | 0.9598 |
| **20** | 0.6098 | 0.6823 | 0.9578 | 0.9683 | 0.9602 |
| **22** | 0.6080 | 0.6820 | 0.9566 | 0.9673 | 0.9604 |
| **24** | 0.6042 | 0.6836 | 0.9558 | 0.9669 | 0.9597 |
| **25** | 0.6079 | 0.6812 | 0.9552 | 0.9678 | 0.9599 |
| **26** | 0.6067 | 0.6821 | 0.9556 | 0.9676 | 0.9597 |
| **28** | 0.6031 | 0.6841 | 0.9548 | 0.9670 | 0.9607 |
| **30** | 0.6088 | 0.6874 | 0.9553 | 0.9685 | 0.9595 |
| **40** | 0.6086 | 0.6796 | 0.9540 | 0.9683 | 0.9592 |
| **50** | 0.6064 | 0.6816 | 0.9568 | 0.9684 | 0.9606 |

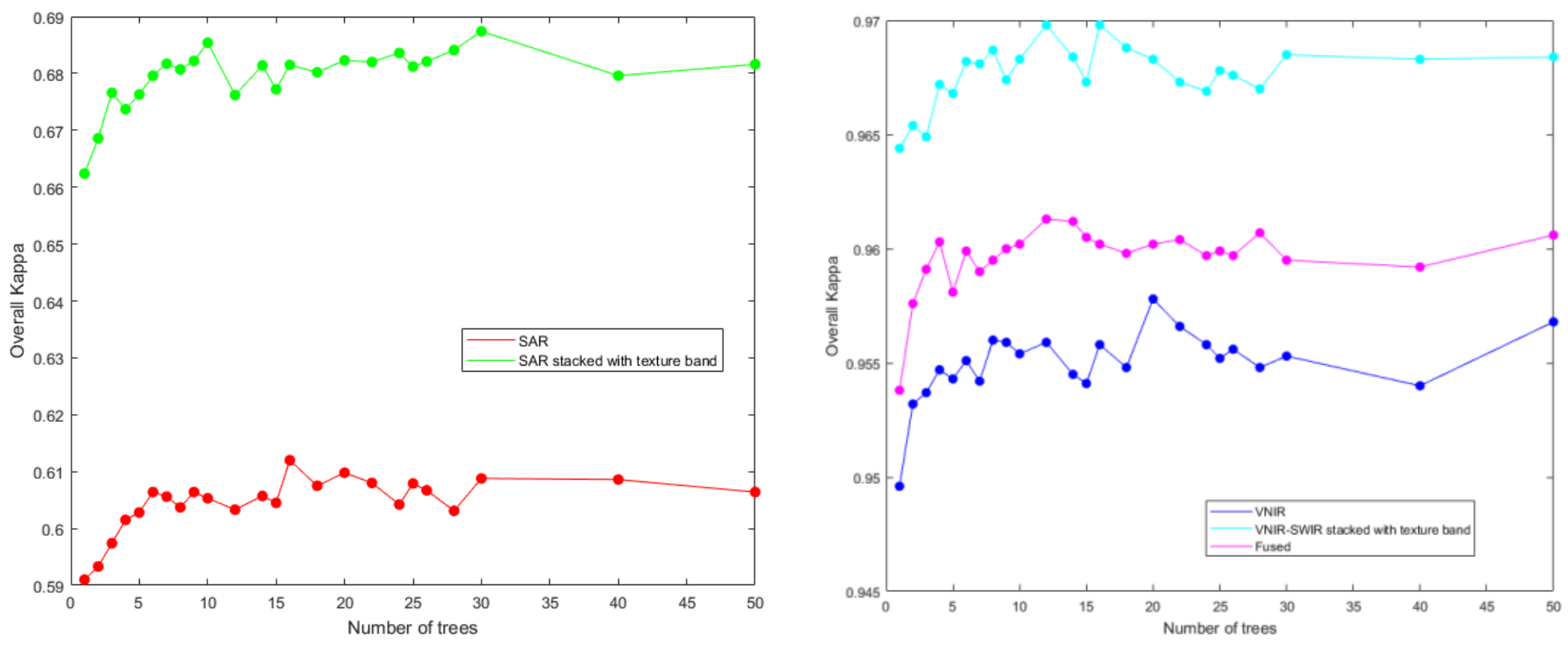

*(a)* *(b)*

*Figure 28(a): Average overall kappa for PCA-RFE for SAR and SAR stacked with texture*
*Figure 28 (b): Average overall kappa for PCA-RFE for VNIR-SWIR, VNIR-SWIR stacked with texture and VNIR-SWIR fused with texture*

### *5.2.6.2 Ensemble using SRP rotation*

*Table 29: Average overall kappa for SRP-RFE for all datasets*

| | SAR | SAR+Texture | VNIR-SWIR | VNIR-SWIR+Texture | Fused |
|---|---|---|---|---|---|
| **Number of trees** | | | | | |
| **1** | 0.5670 | 0.6473 | 0.9432 | 0.9557 | 0.9490 |
| **2** | 0.5943 | 0.6584 | 0.9503 | 0.9603 | 0.9546 |
| **3** | 0.6022 | 0.6673 | 0.9533 | 0.9643 | 0.9552 |
| **4** | 0.6056 | 0.6743 | 0.9536 | 0.9660 | 0.9599 |
| **5** | 0.6087 | 0.6740 | 0.9541 | 0.9653 | 0.9566 |
| **6** | 0.6112 | 0.6797 | 0.9539 | 0.9667 | 0.9607 |
| **7** | 0.6124 | 0.6768 | 0.9556 | 0.9671 | 0.959 |
| **8** | 0.6095 | 0.6746 | 0.9563 | 0.9677 | 0.9608 |
| **9** | 0.6137 | 0.6790 | 0.9569 | 0.9671 | 0.9611 |
| **10** | 0.6129 | 0.6812 | 0.9571 | 0.9669 | 0.9609 |
| **12** | 0.6175 | 0.6791 | 0.9549 | 0.9672 | 0.9597 |
| **14** | 0.6121 | 0.6849 | 0.9570 | 0.9681 | 0.9620 |
| **15** | 0.6163 | 0.6824 | 0.9571 | 0.9677 | 0.9587 |
| **16** | 0.6160 | 0.6818 | 0.9566 | 0.9683 | 0.9616 |
| **18** | 0.6181 | 0.6795 | 0.9571 | 0.9671 | 0.9612 |
| **20** | 0.6182 | 0.6825 | 0.9549 | 0.9694 | 0.9605 |
| **22** | 0.6126 | 0.6838 | 0.9591 | 0.9659 | 0.9612 |
| **24** | 0.6179 | 0.6837 | 0.9580 | 0.9680 | 0.9612 |
| **25** | 0.6180 | 0.6818 | 0.9570 | 0.9664 | 0.9609 |
| **26** | 0.6162 | 0.6807 | 0.9577 | 0.9686 | 0.9605 |
| **28** | 0.6176 | 0.6841 | 0.9549 | 0.9675 | 0.9612 |
| **30** | 0.6180 | 0.6805 | 0.9566 | 0.9684 | 0.9617 |
| **40** | 0.6182 | 0.6840 | 0.9585 | 0.9666 | 0.9619 |
| **50** | 0.6172 | 0.6856 | 0.9578 | 0.9682 | 0.9617 |

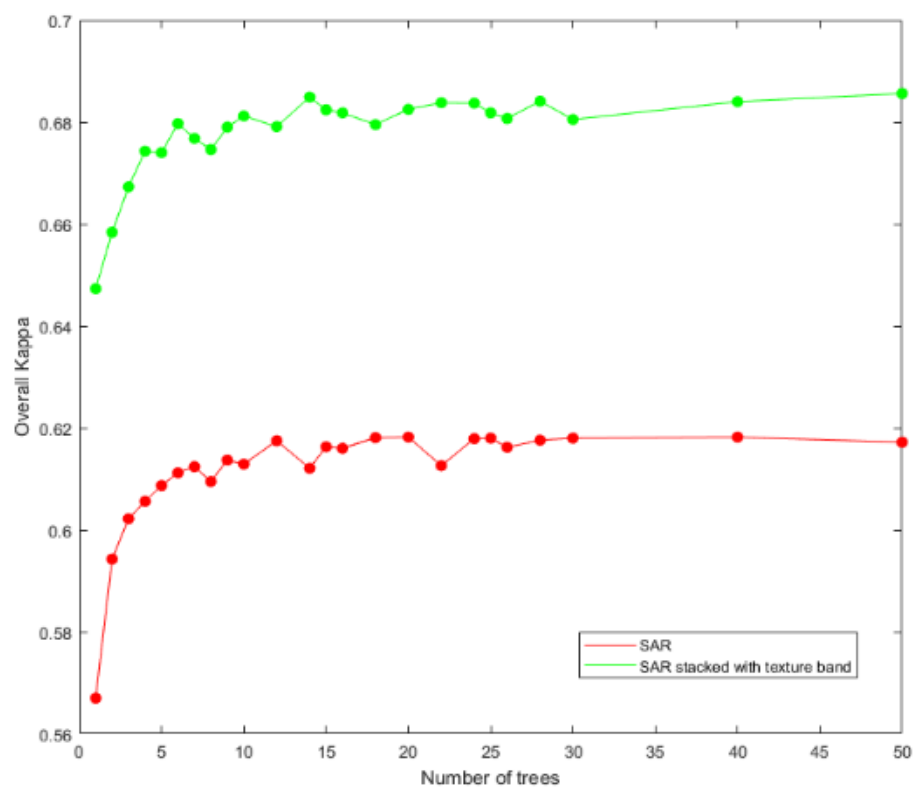

*(a)*

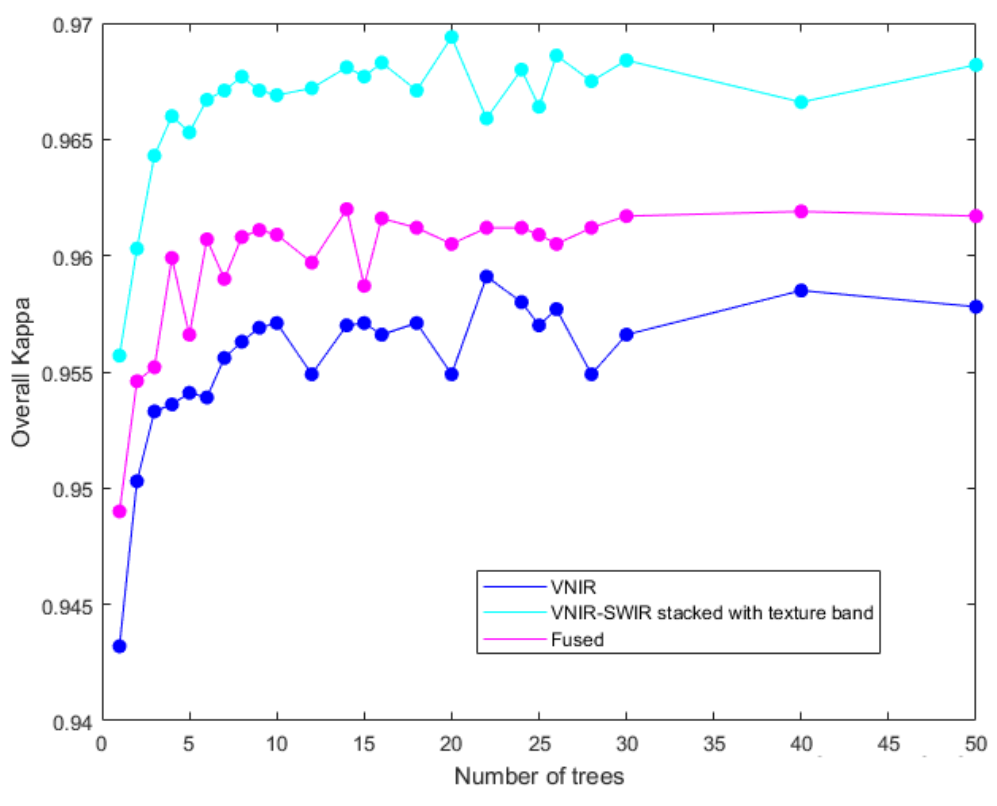

*(b)*

*Figure 29 (a): Average overall kappa for SRP-RFE for SAR and SAR stacked with texture*
*Figure 29 (b): Average overall kappa for SRP-RFE for VNIR-SWIR, VNIR-SWIR stacked with texture and VNIR-SWIR fused with texture*

### *5.2.6.3 Ensemble using CRP rotation*

*Table 30: Average overall kappa for CRP-RFE for all datasets*

| | **SAR** | **SAR+Texture** | **VNIR-SWIR** | **VNIR-SWIR+Texture** | **Fused** |
|---|---|---|---|---|---|
| **Number of trees** | | | | | |
| **1** | 0.5489 | 0.6332 | 0.9466 | 0.9572 | 0.9512 |
| **2** | 0.5737 | 0.6382 | 0.9496 | 0.9636 | 0.9562 |
| **3** | 0.5875 | 0.6510 | 0.9542 | 0.9657 | 0.9568 |
| **4** | 0.5866 | 0.6595 | 0.9557 | 0.9672 | 0.9596 |
| **5** | 0.5955 | 0.6628 | 0.9566 | 0.9662 | 0.9606 |
| **6** | 0.5956 | 0.6593 | 0.9560 | 0.9677 | 0.9625 |
| **7** | 0.5930 | 0.6613 | 0.9538 | 0.9685 | 0.9619 |
| **8** | 0.6013 | 0.6661 | 0.9571 | 0.9692 | 0.9618 |
| **9** | 0.5990 | 0.6684 | 0.9559 | 0.9665 | 0.9608 |
| **10** | 0.6022 | 0.6710 | 0.9591 | 0.9693 | 0.9622 |
| **12** | 0.5999 | 0.6699 | 0.9571 | 0.9685 | 0.9641 |
| **14** | 0.6005 | 0.6698 | 0.9589 | 0.9695 | 0.9615 |
| **15** | 0.6043 | 0.6712 | 0.9575 | 0.9688 | 0.9633 |
| **16** | 0.5977 | 0.6708 | 0.9576 | 0.9685 | 0.9640 |
| **18** | 0.6026 | 0.6678 | 0.9574 | 0.9691 | 0.9618 |
| **20** | 0.6061 | 0.6697 | 0.9590 | 0.9710 | 0.9633 |
| **22** | 0.6050 | 0.6755 | 0.9585 | 0.9679 | 0.9649 |
| **24** | 0.6065 | 0.6716 | 0.9591 | 0.9690 | 0.9642 |
| **25** | 0.6028 | 0.6696 | 0.9599 | 0.9693 | 0.9630 |
| **26** | 0.6035 | 0.6698 | 0.9597 | 0.9695 | 0.9645 |
| **28** | 0.6024 | 0.6725 | 0.9584 | 0.9694 | 0.9634 |
| **30** | 0.6029 | 0.6752 | 0.9570 | 0.9703 | 0.9625 |
| **40** | 0.6048 | 0.6748 | 0.9576 | 0.9688 | 0.9624 |
| **50** | 0.6063 | 0.6733 | 0.9592 | 0.9694 | 0.9624 |

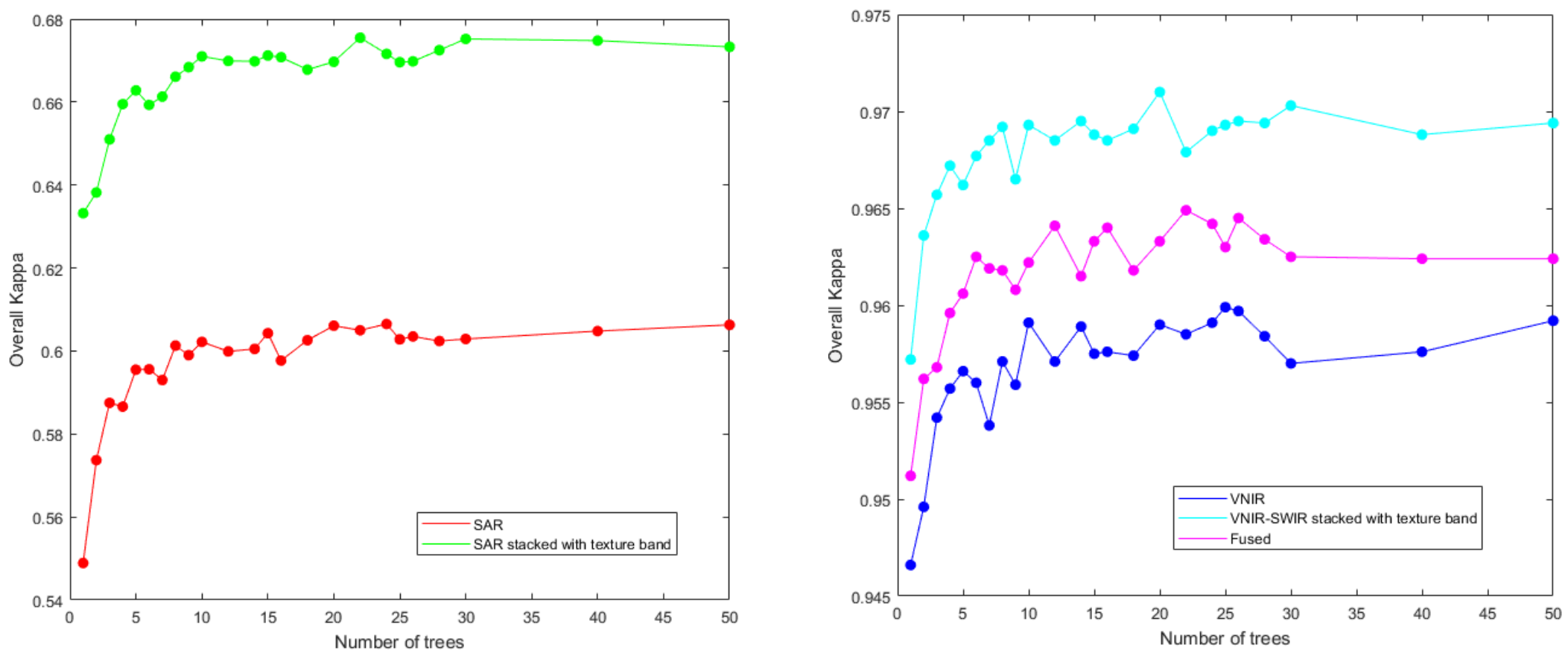

(a) (b)

*Figure 30 (a): Average overall kappa for CRP-RFE for SAR and SAR stacked with texture*
*Figure 30 (b): Average overall kappa for CRP-RFE for VNIR-SWIR, VNIR-SWIR stacked with texture and VNIR-SWIR fused with texture*

The optimum values for the number of RFs in the ensembles (table 31) are calculated by performing hypothesis testing between the first classifier and the $i^{th}$ classifier. The Z-values obtained for all the classifiers are compared and the number of RFs corresponding to the highest Z-value is considered as the most optimum value that can be used for training the ensembles.

*Table 31: Optimum number of random forests in each ensemble*

| Ensembles | SAR | SAR + texture | VNIR-SWIR | VNIR-SWIR+ texture | Fused |
|---|---|---|---|---|---|
| **PCA-RFE** | 16 | 30 | 22 | 16 | 12 |
| **SRP-RFE** | 20 | 14 | 22 | 20 | 15 |
| **CRP-RFE** | 24 | 22 | 25 | 20 | 22 |

### 5.2.7 Classified imageries

For each classifier, 5 classified images are generated (i.e. one for each dataset). They are represented in figures 31-50.

#### *5.2.7.1 Random forest*

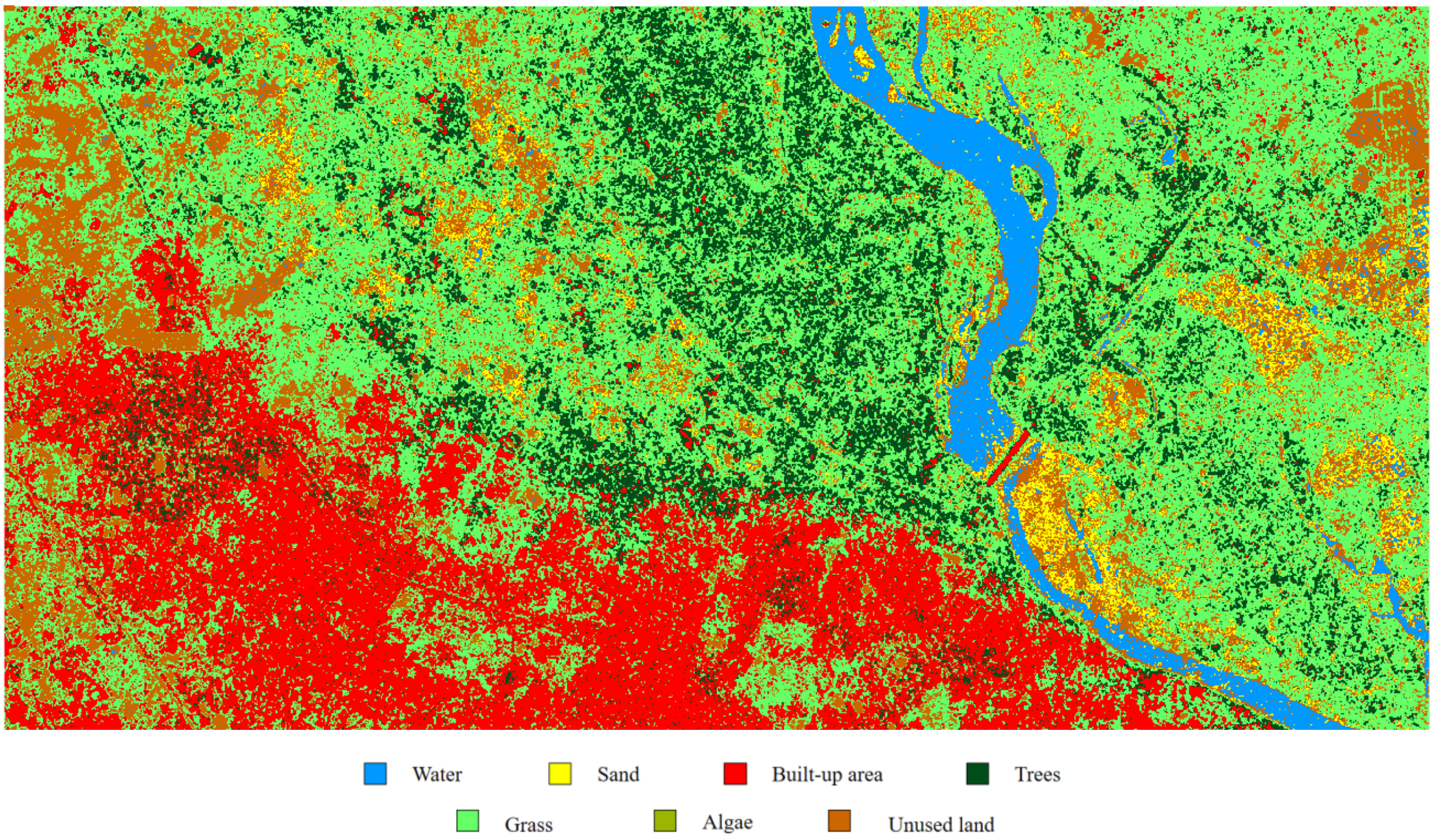

*Figure 31: Classified imagery for SAR bands using RF*

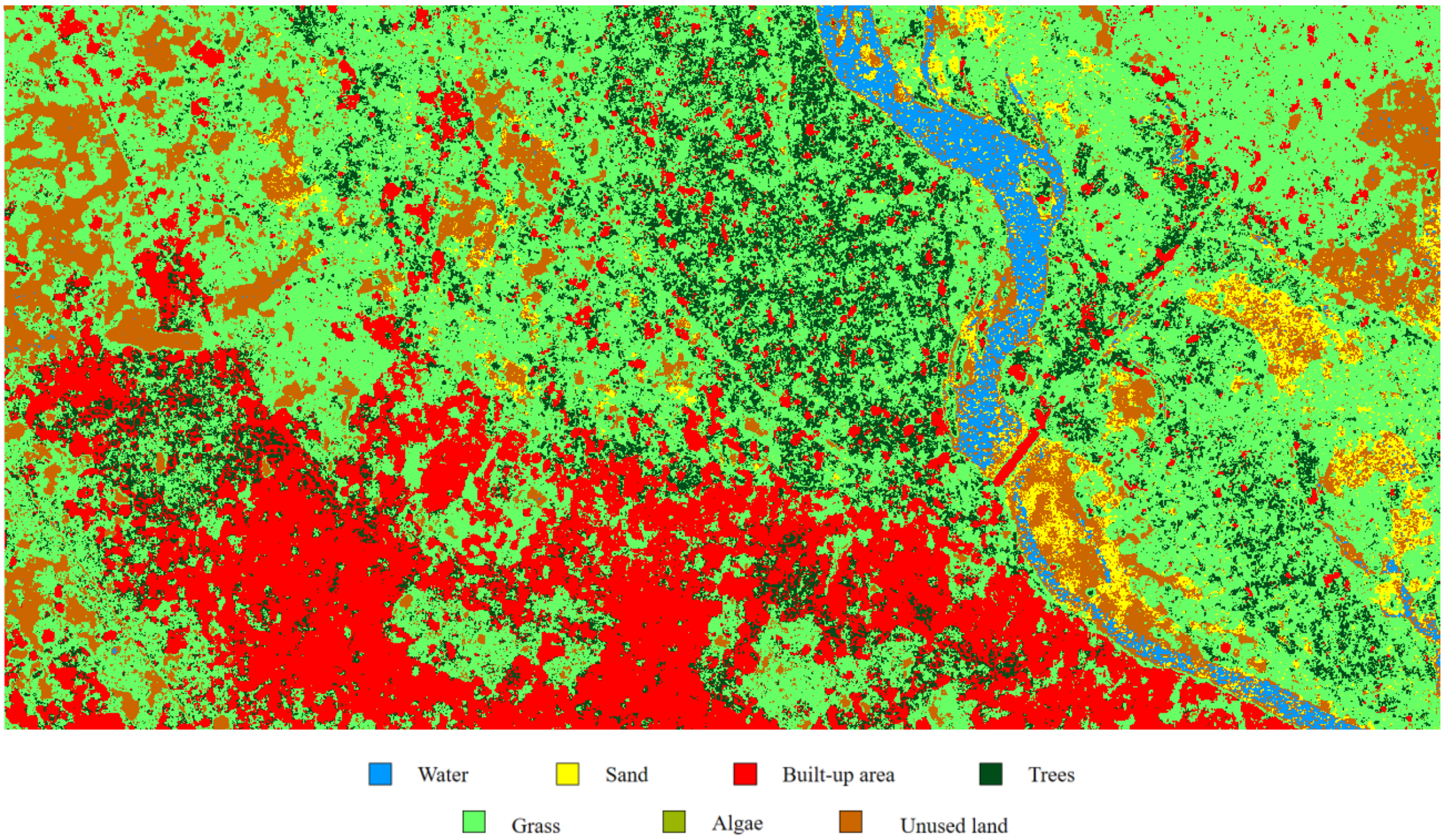

*Figure 32: Classified imagery for SAR bands stacked with texture band using RF*

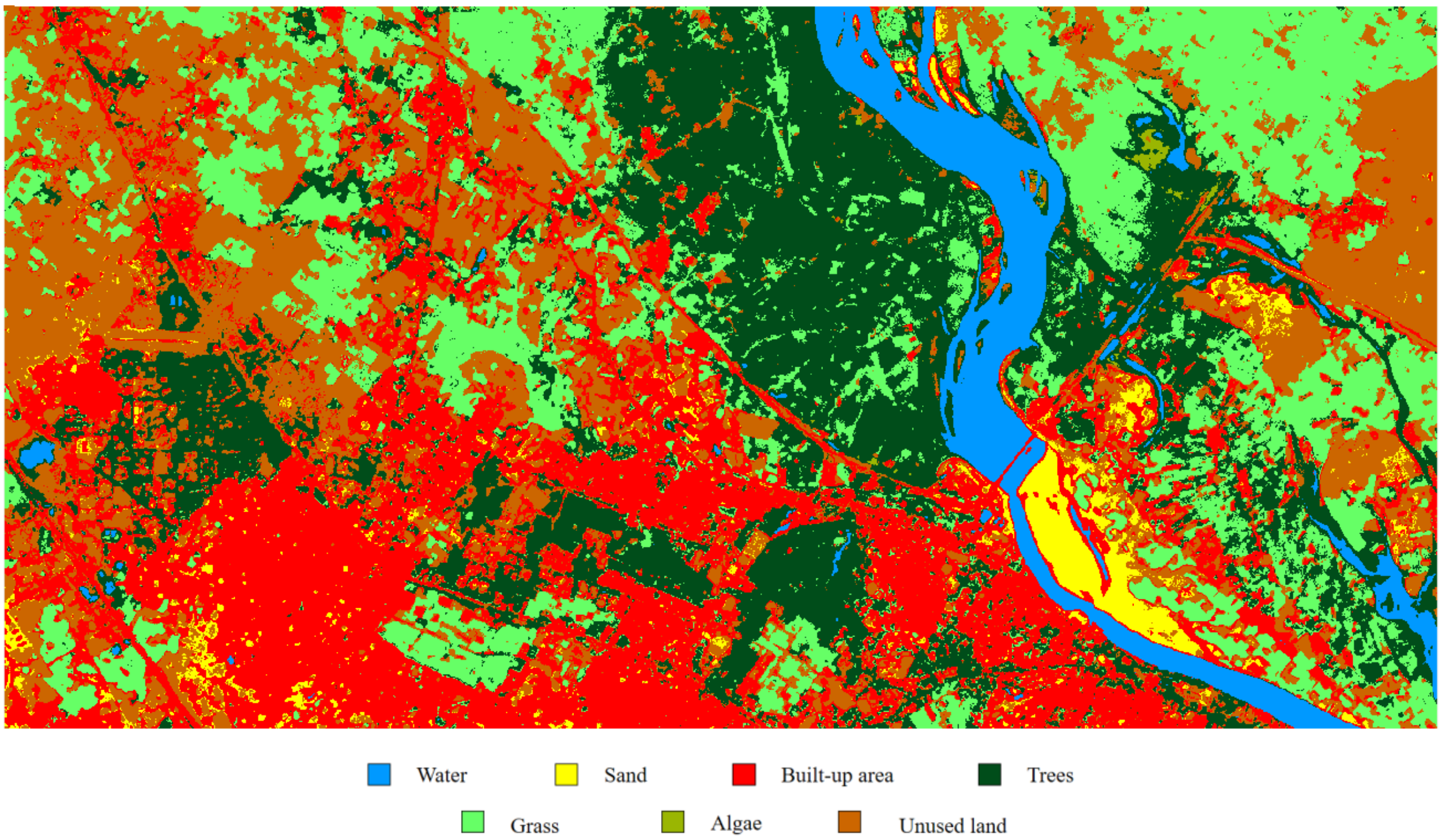

*Figure 33: Classified imagery for VNIR-SWIR bands using RF*

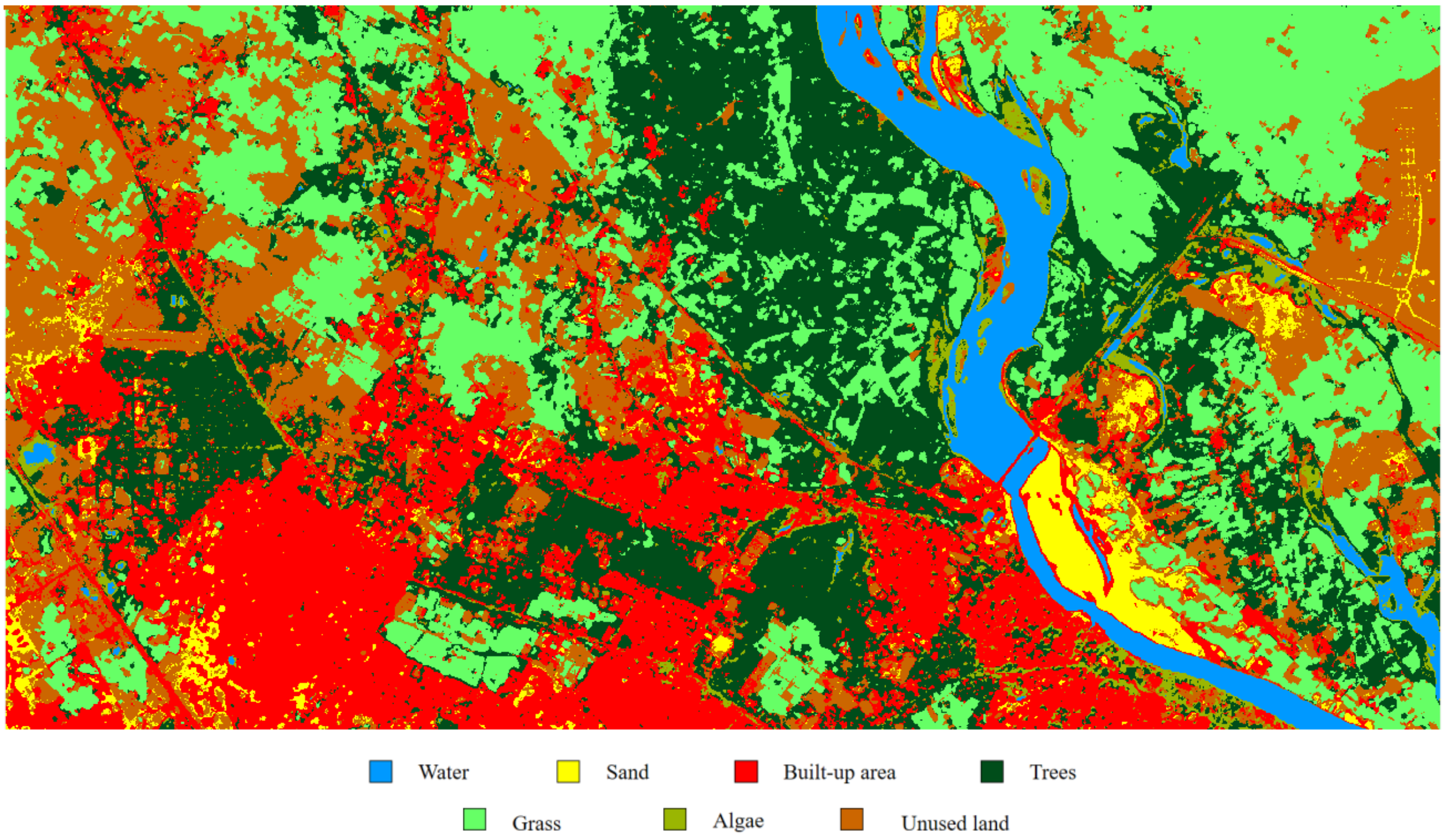

*Figure 34: Classified imagery for VNIR-SWIR bands stacked with texture band using RF*

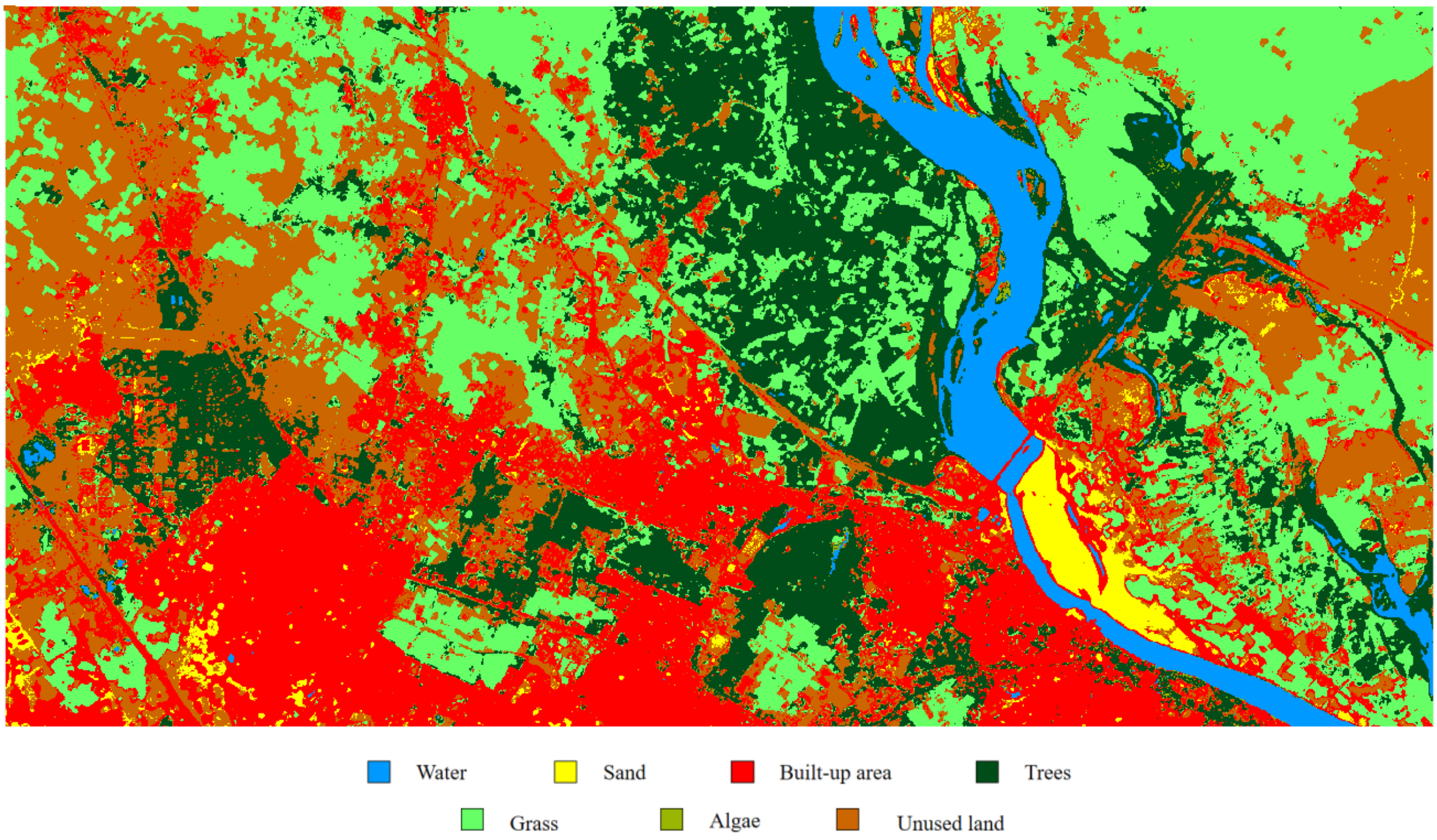

*Figure 35: Classified imagery for VNIR-SWIR bands fused with texture band using RF*

### 5.2.7.2 PCA-RFE

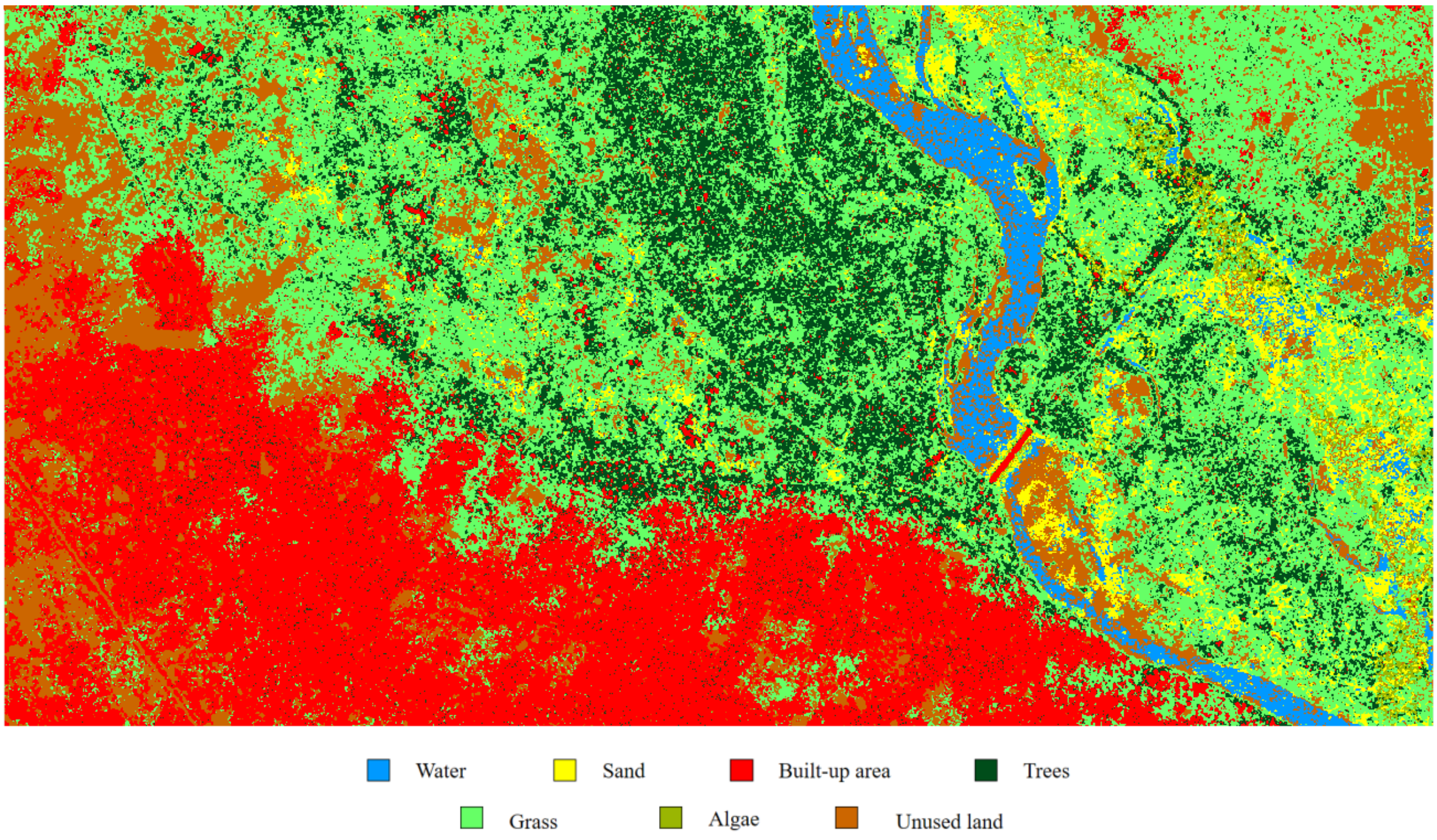

*Figure 36: Classified imagery for SAR bands using PCA-RFE*

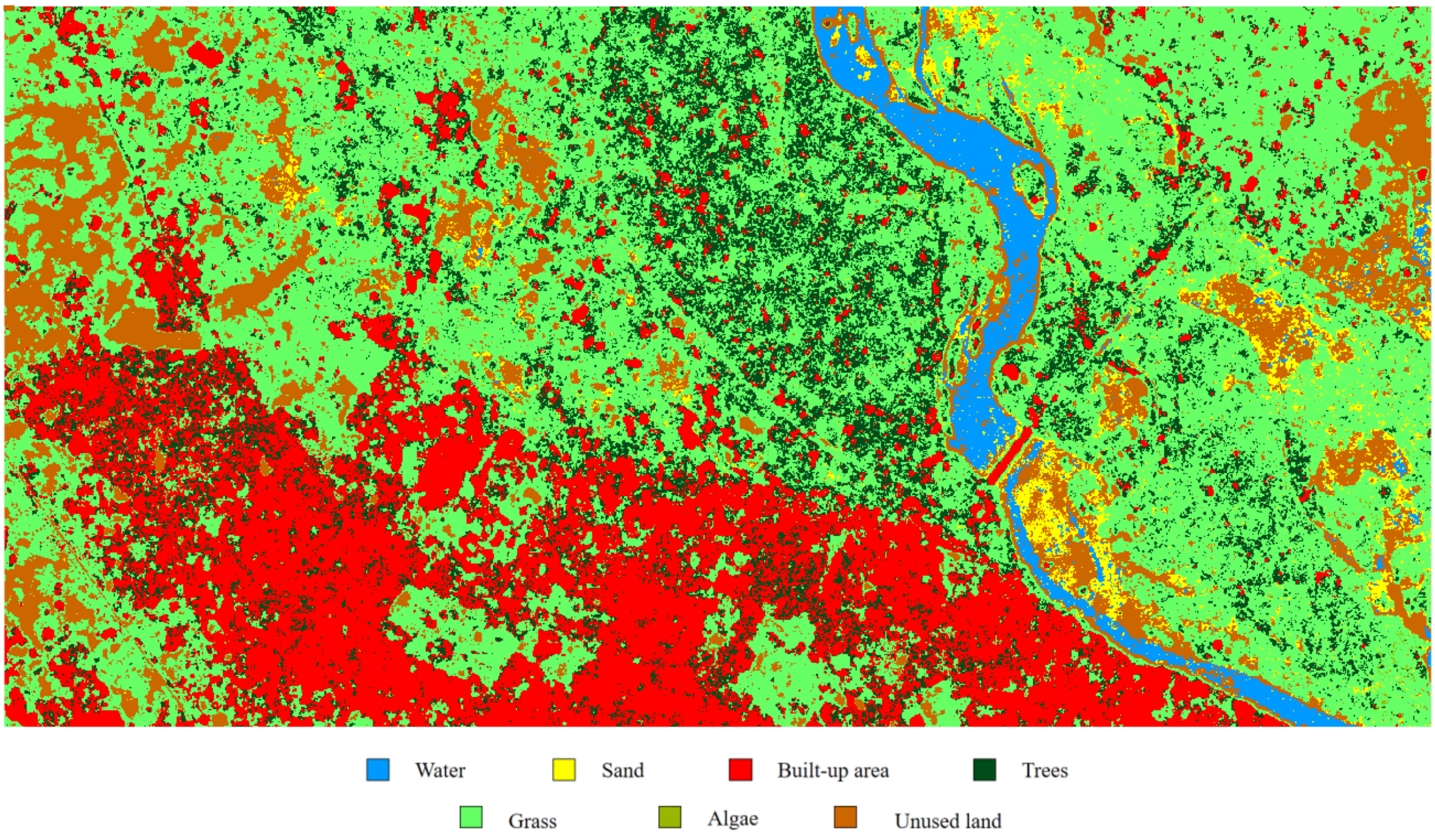

*Figure 37: Classified imagery for SAR bands stacked with texture band using PCA-RFE*

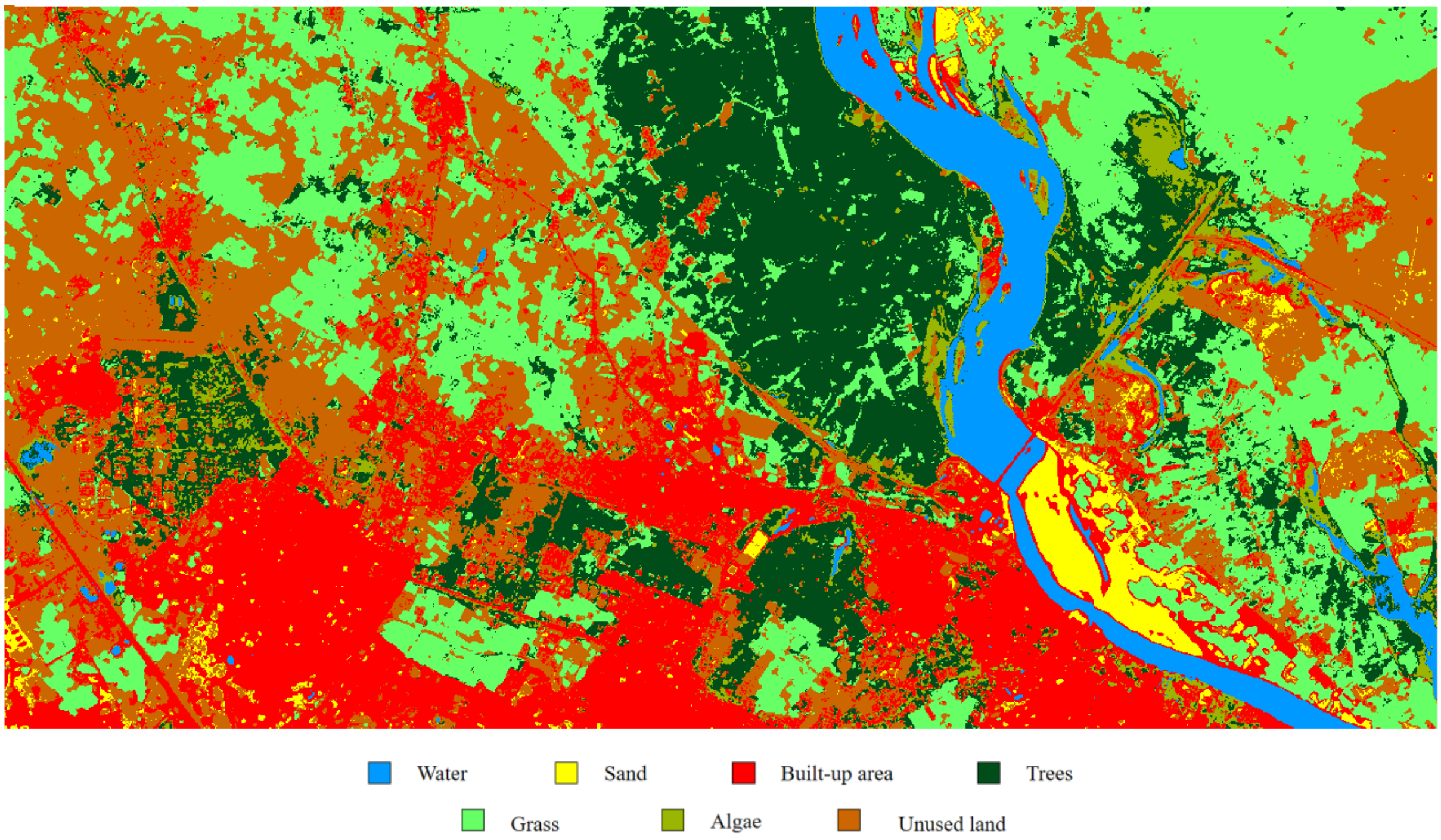

*Figure 38: Classified imagery for VNIR-SWIR bands using PCA-RFE*

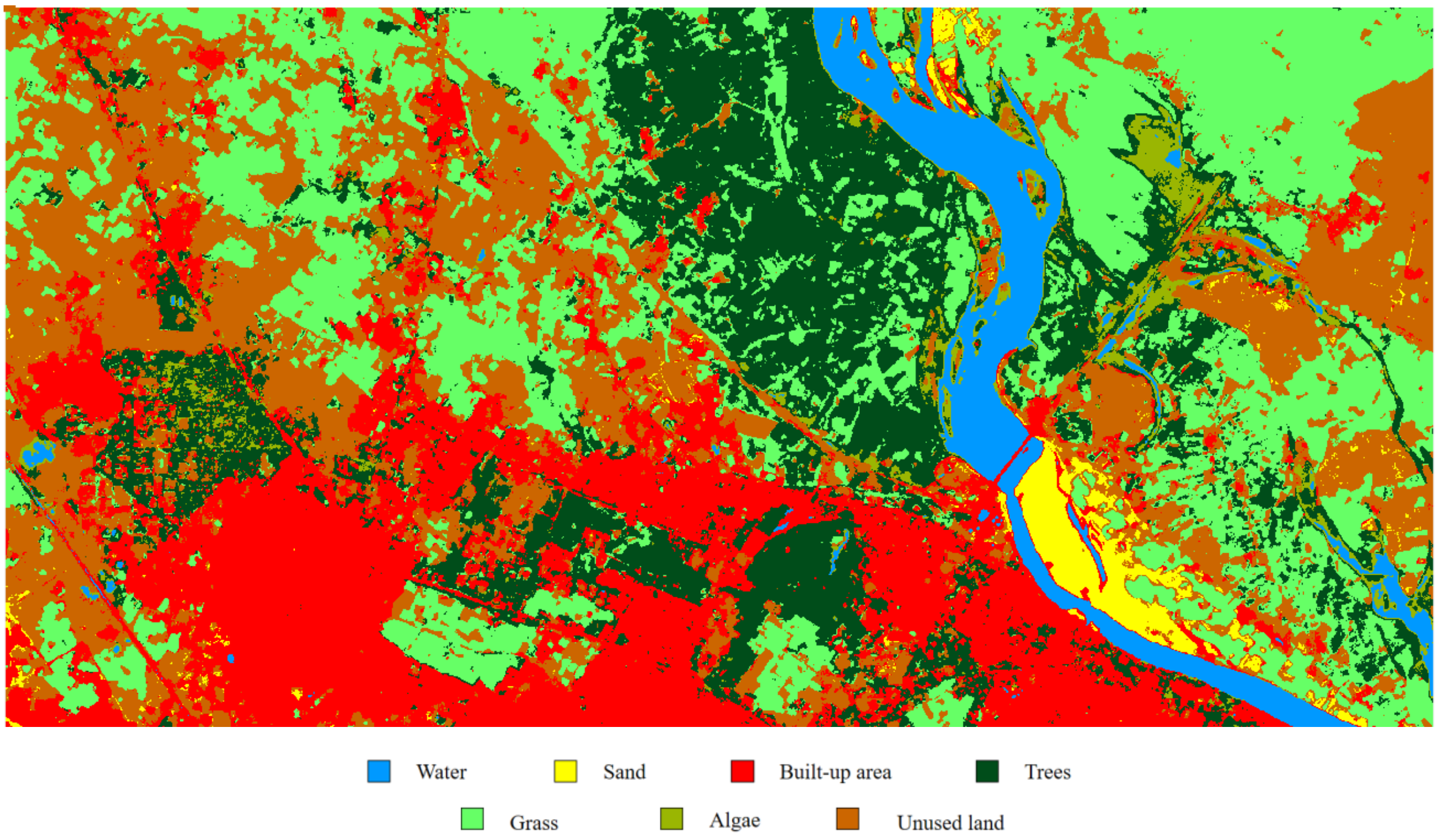

*Figure 39: Classified imagery for VNIR-SWIR bands stacked with texture band using PCA-RFE*

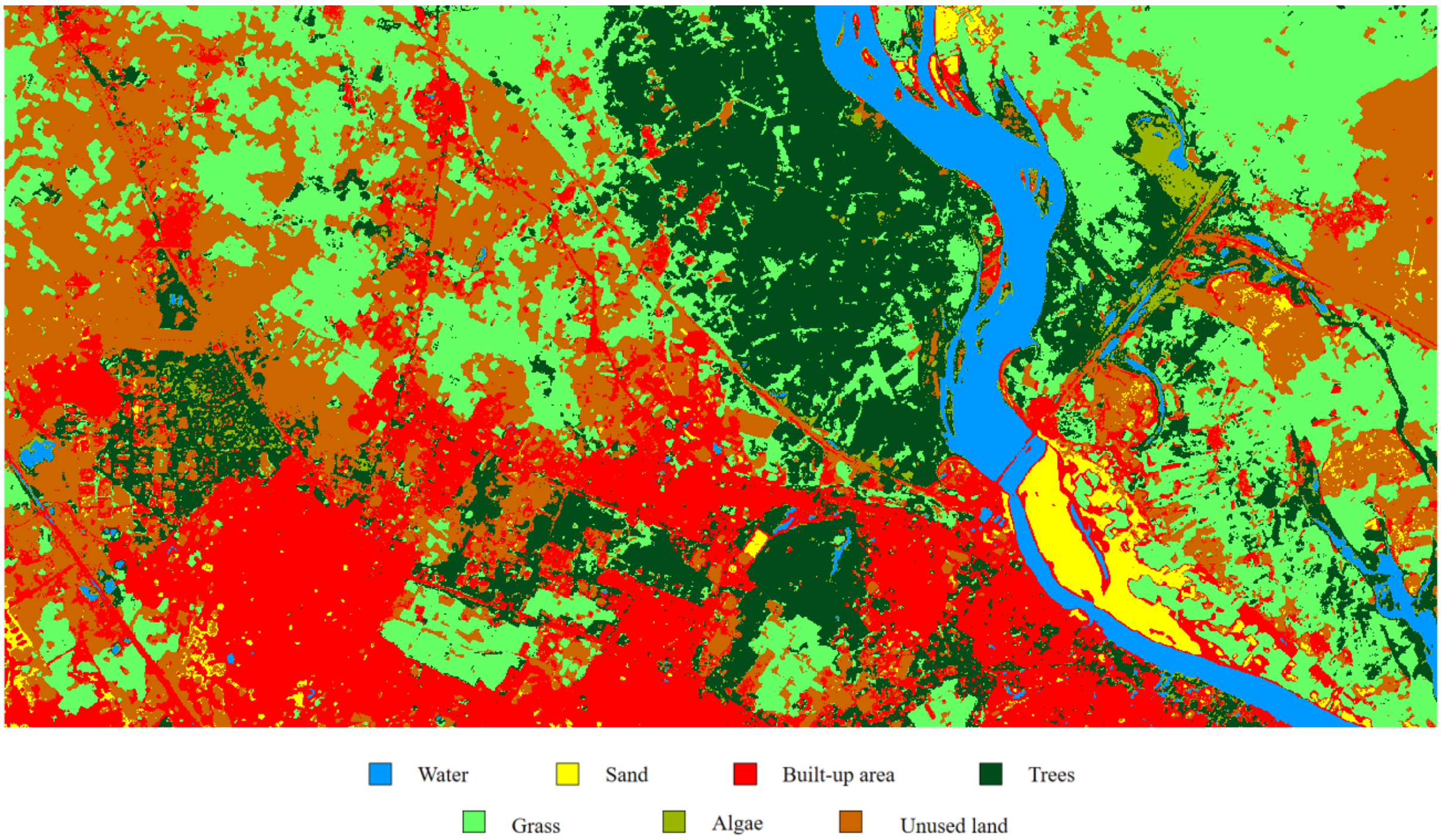

*Figure 40: Classified imagery for VNIR-SWIR bands fused with texture band using PCA-RFE*

#### 5.2.7.3 SRP-RFE

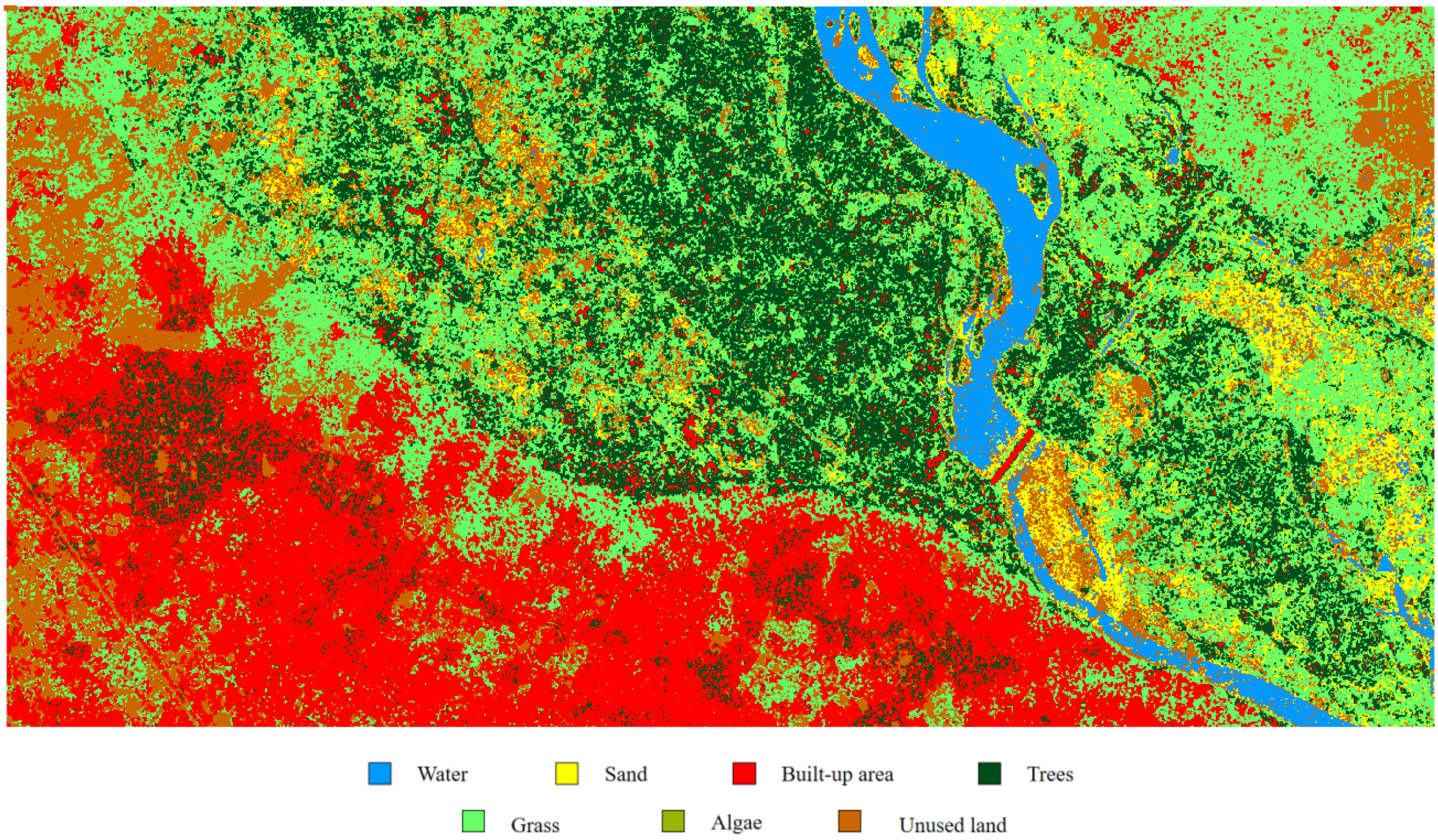

*Figure 41: Classified imagery for SAR bands using SRP-RFE*

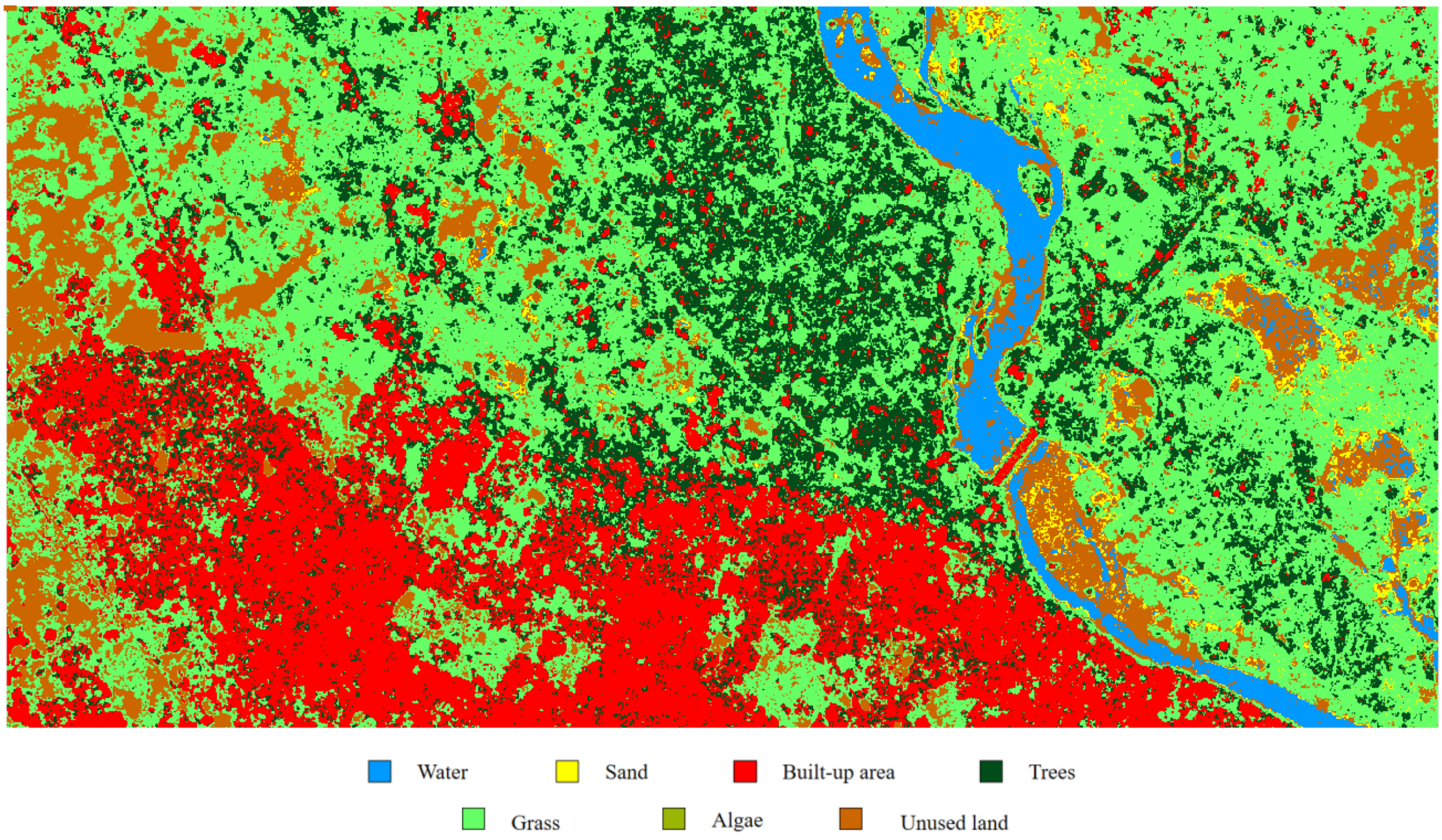

*Figure 42: Classified imagery for SAR bands stacked with texture band using SRP-RFE*

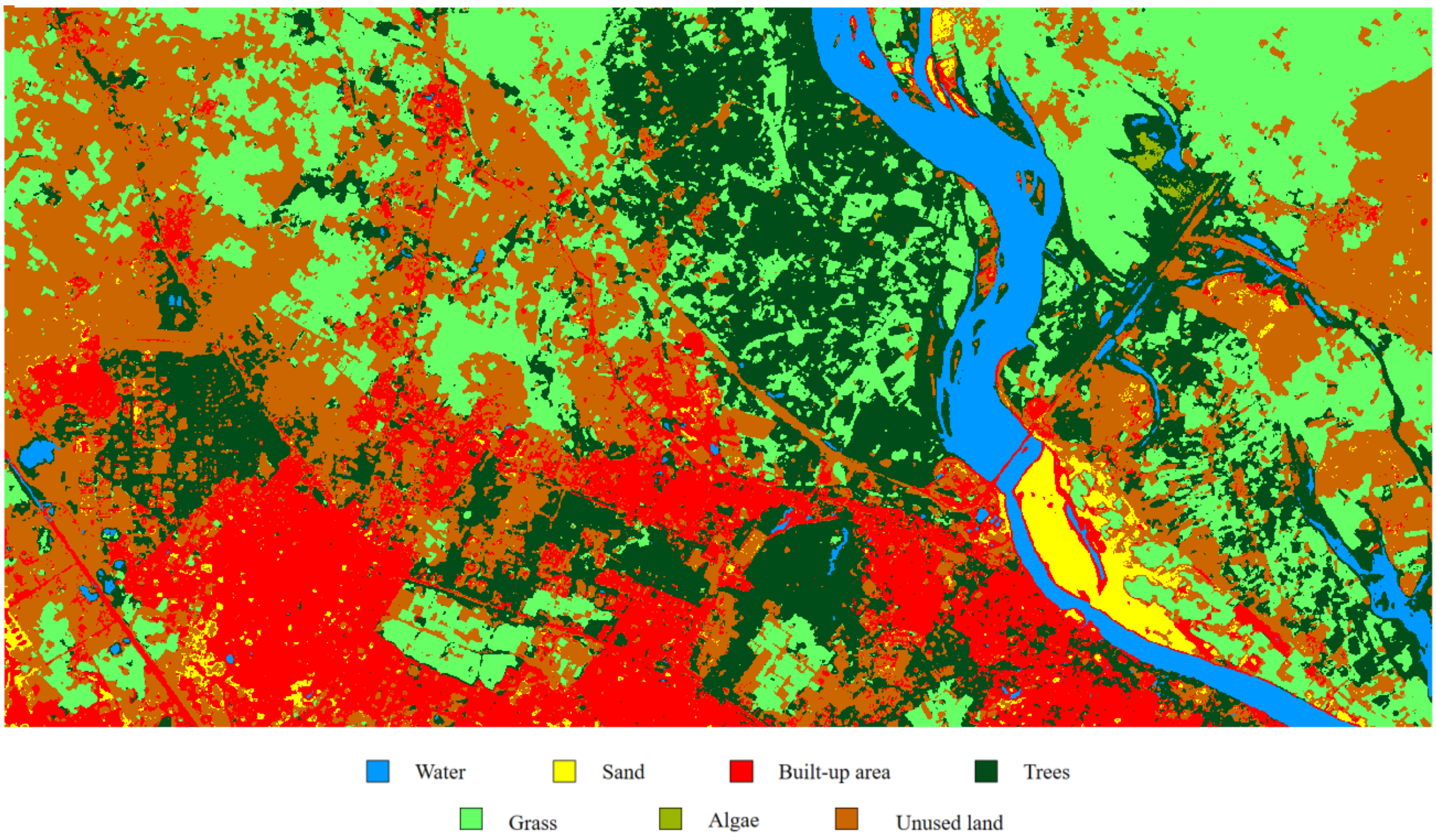

*Figure 43: Classified imagery for VNIR-SWIR bands using SRP-RFE*

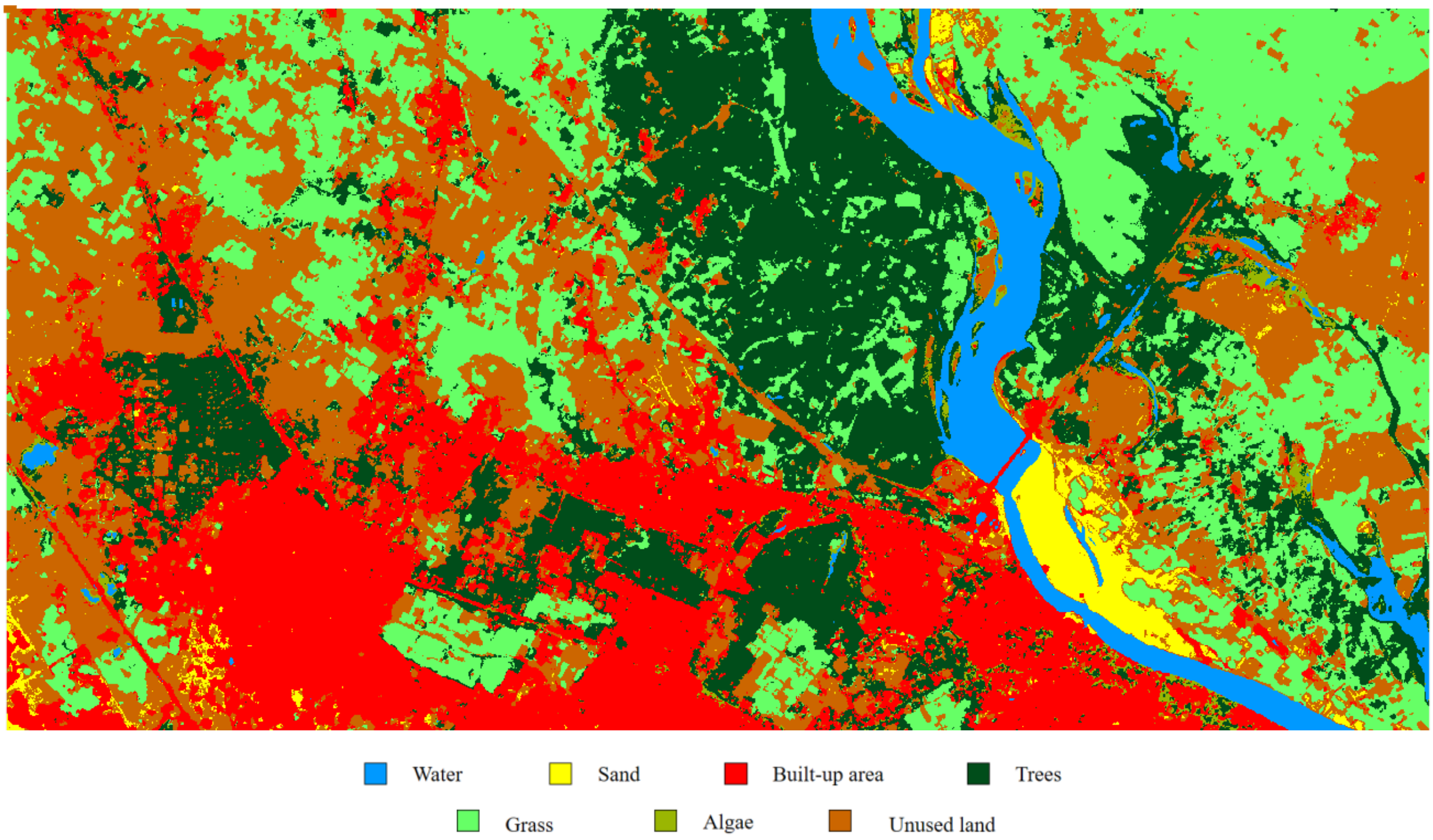

*Figure 44: Classified imagery for VNIR-SWIR bands stacked with texture band using SRP-RFE*

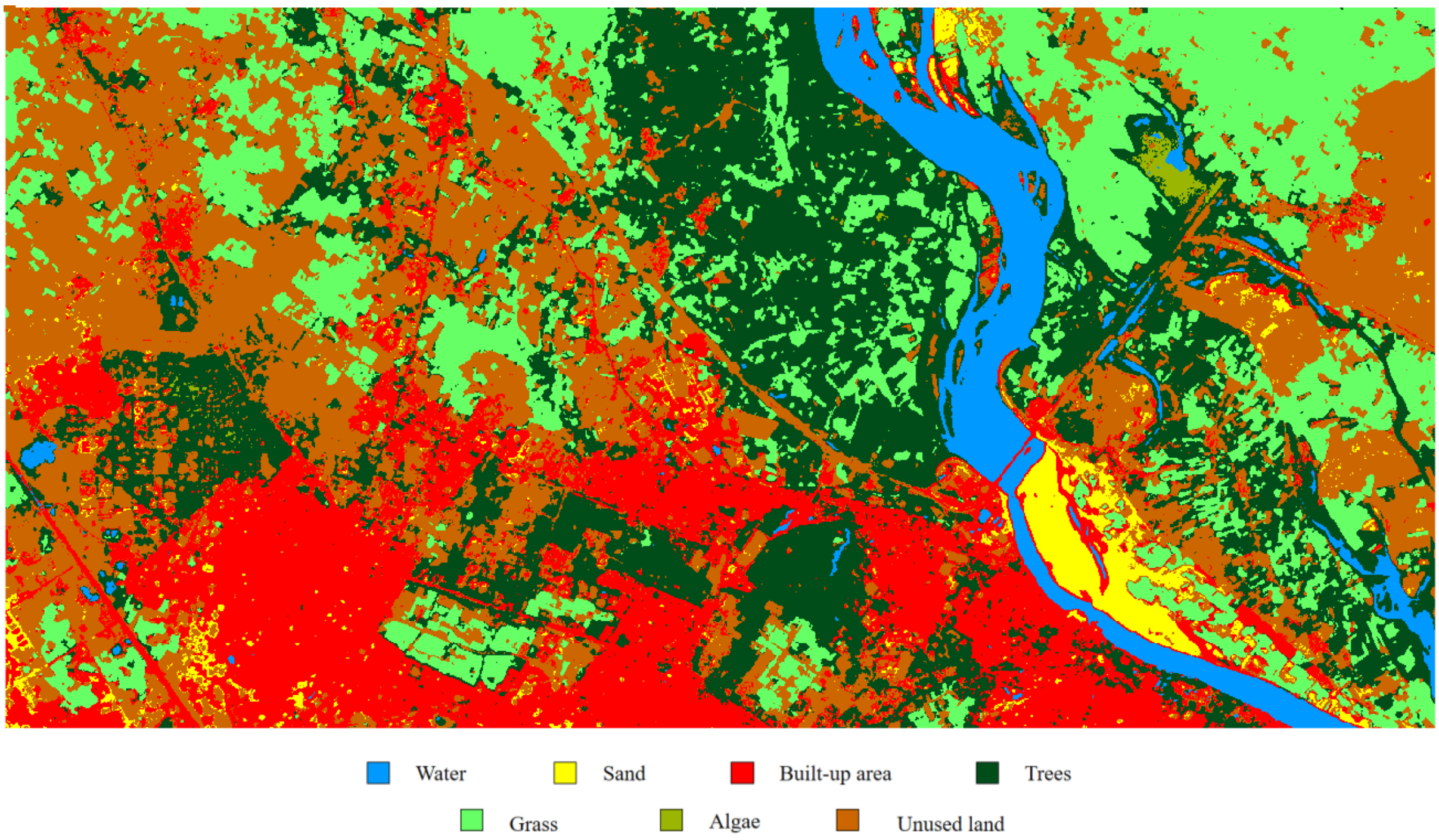

*Figure 45: Classified imagery for VNIR-SWIR bands fused with texture band using SRP-RFE*

### 5.2.7.4 CRP-RFE

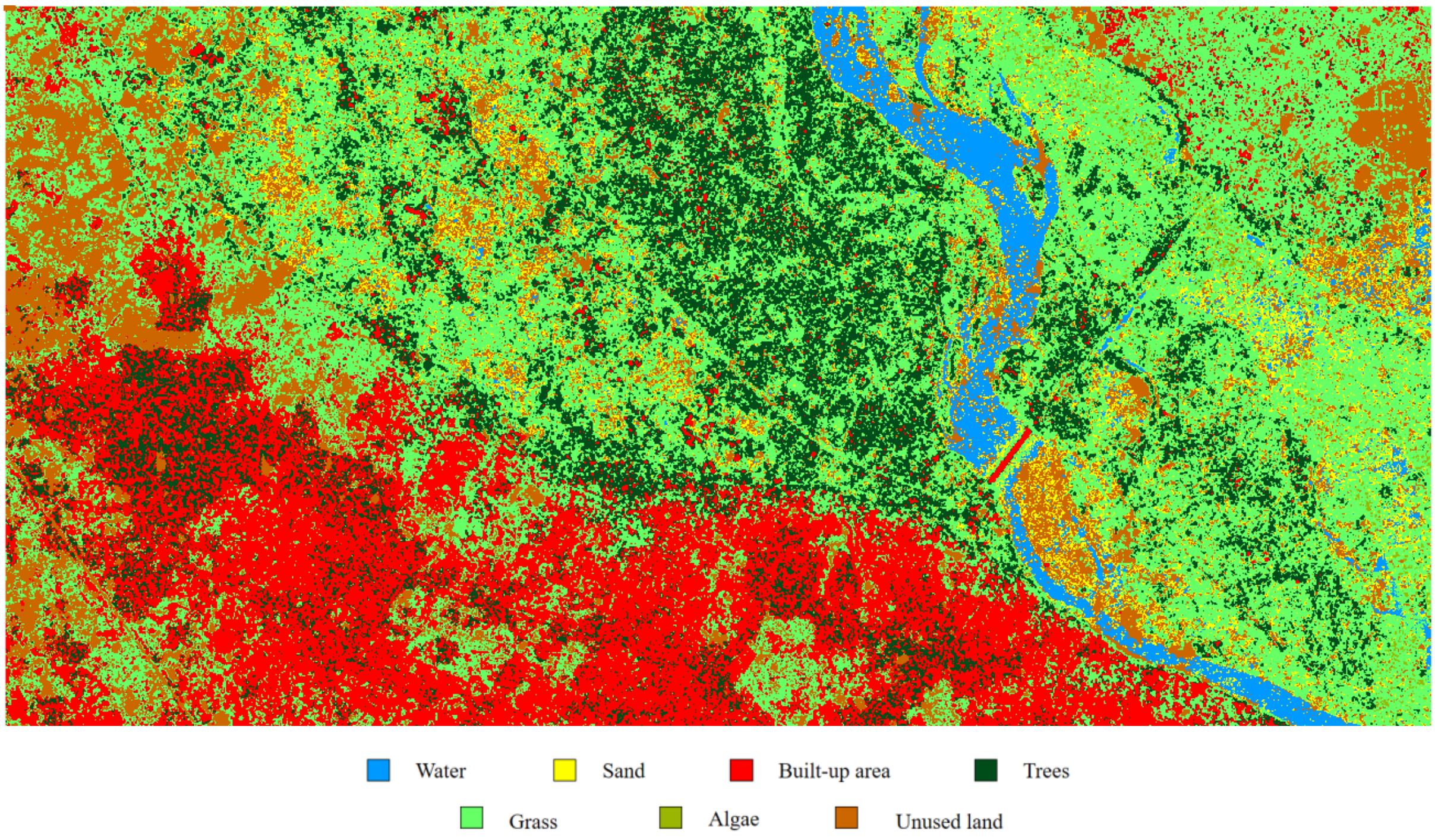

*Figure 46: Classified imagery for SAR bands using CRP-RFE*

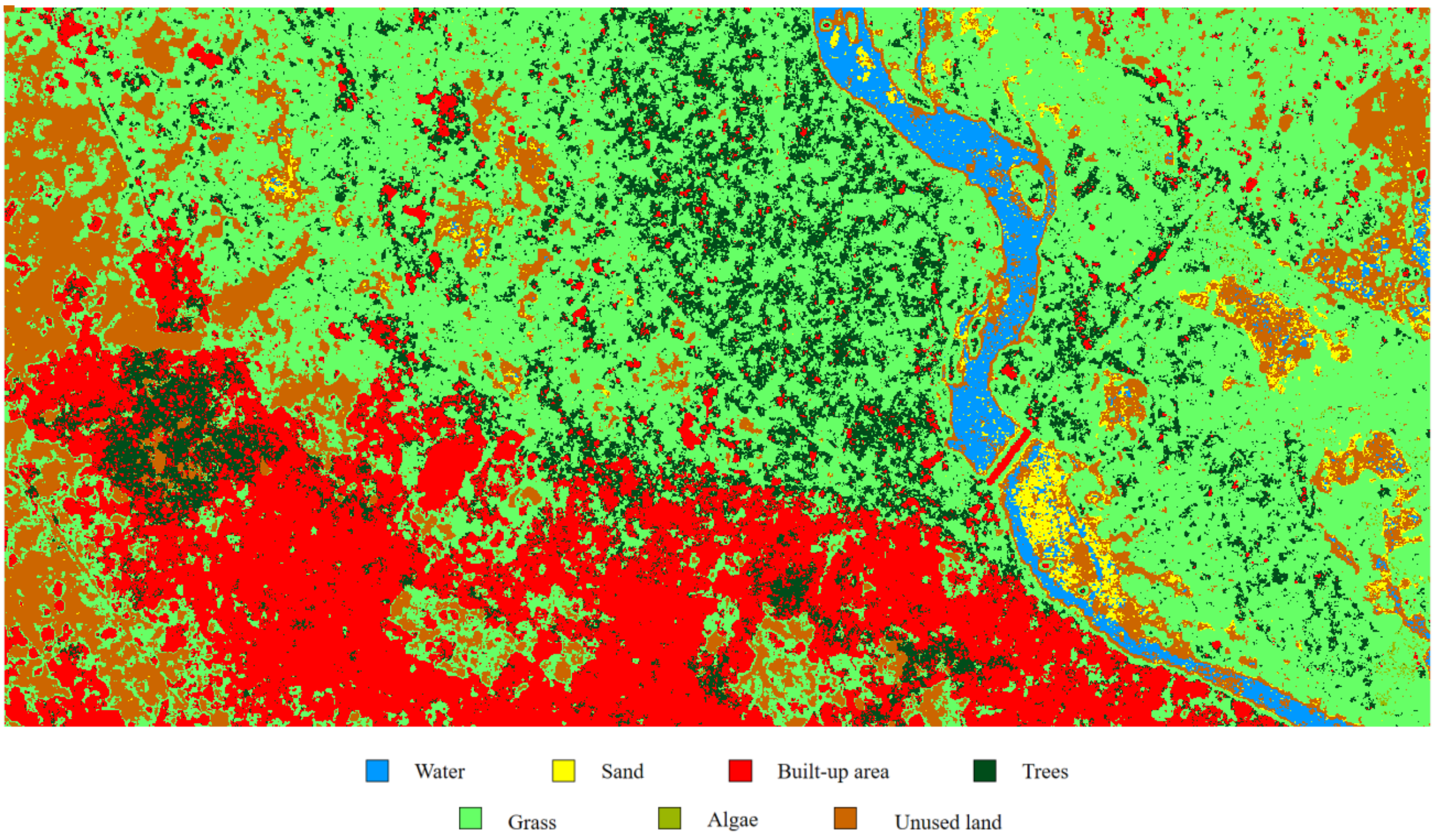

*Figure 47: Classified imagery for SAR bands stacked with texture band using CRP-RFE*

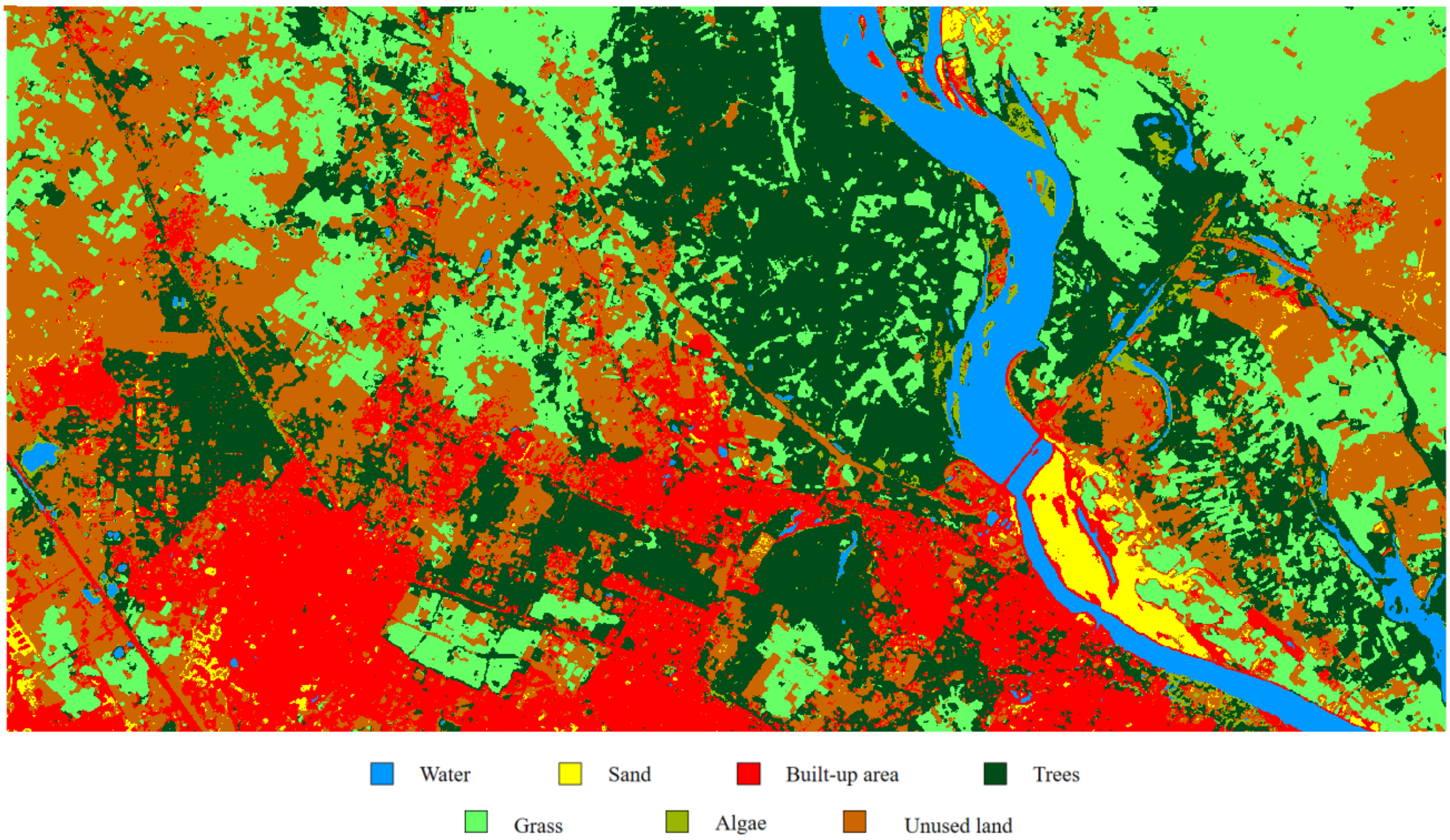

*Figure 48: Classified imagery for VNIR-SWIR bands using CRP-RFE*

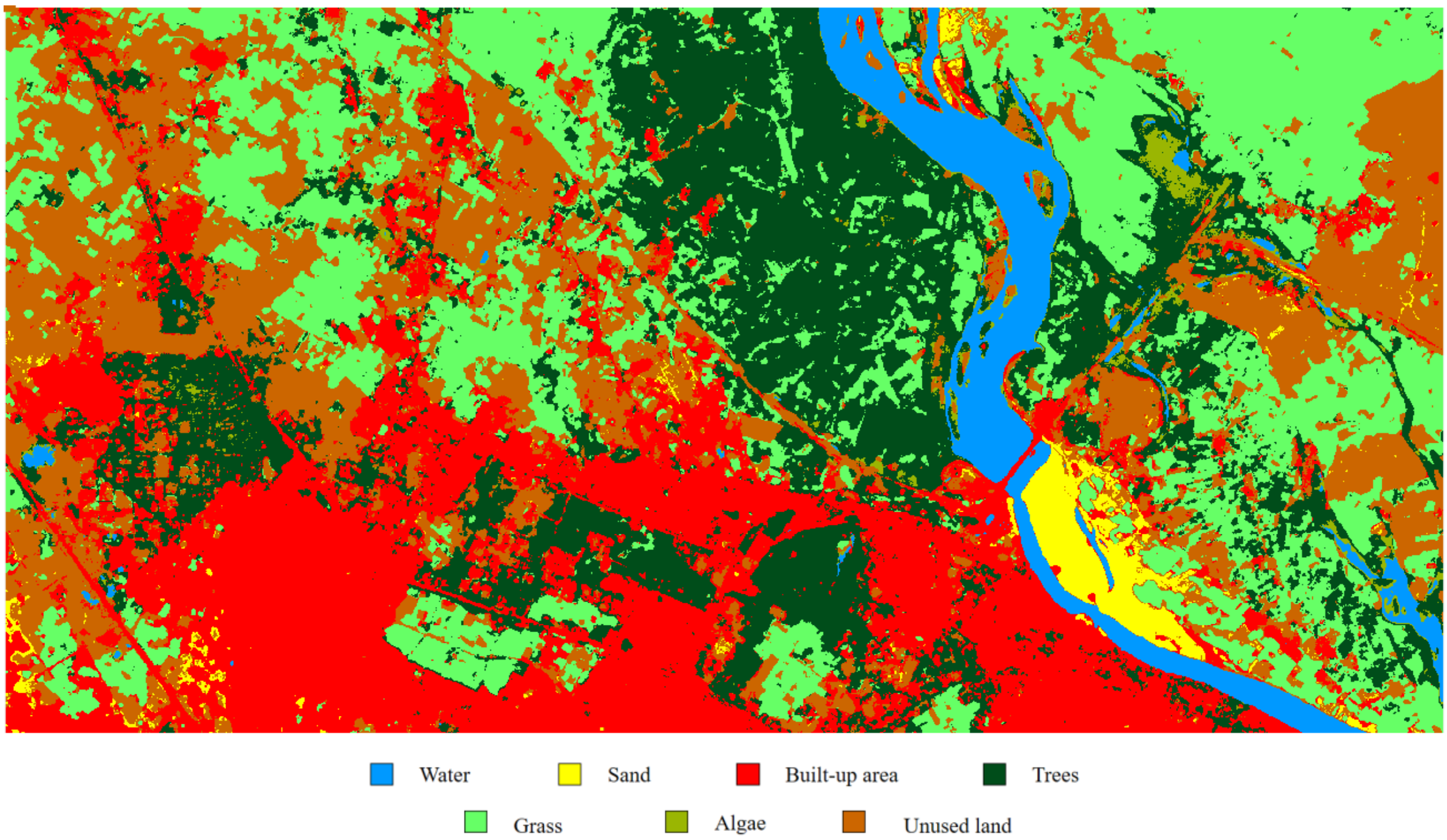

*Figure 49: Classified imagery for VNIR-SWIR bands stacked with texture band using CRP-RFE*

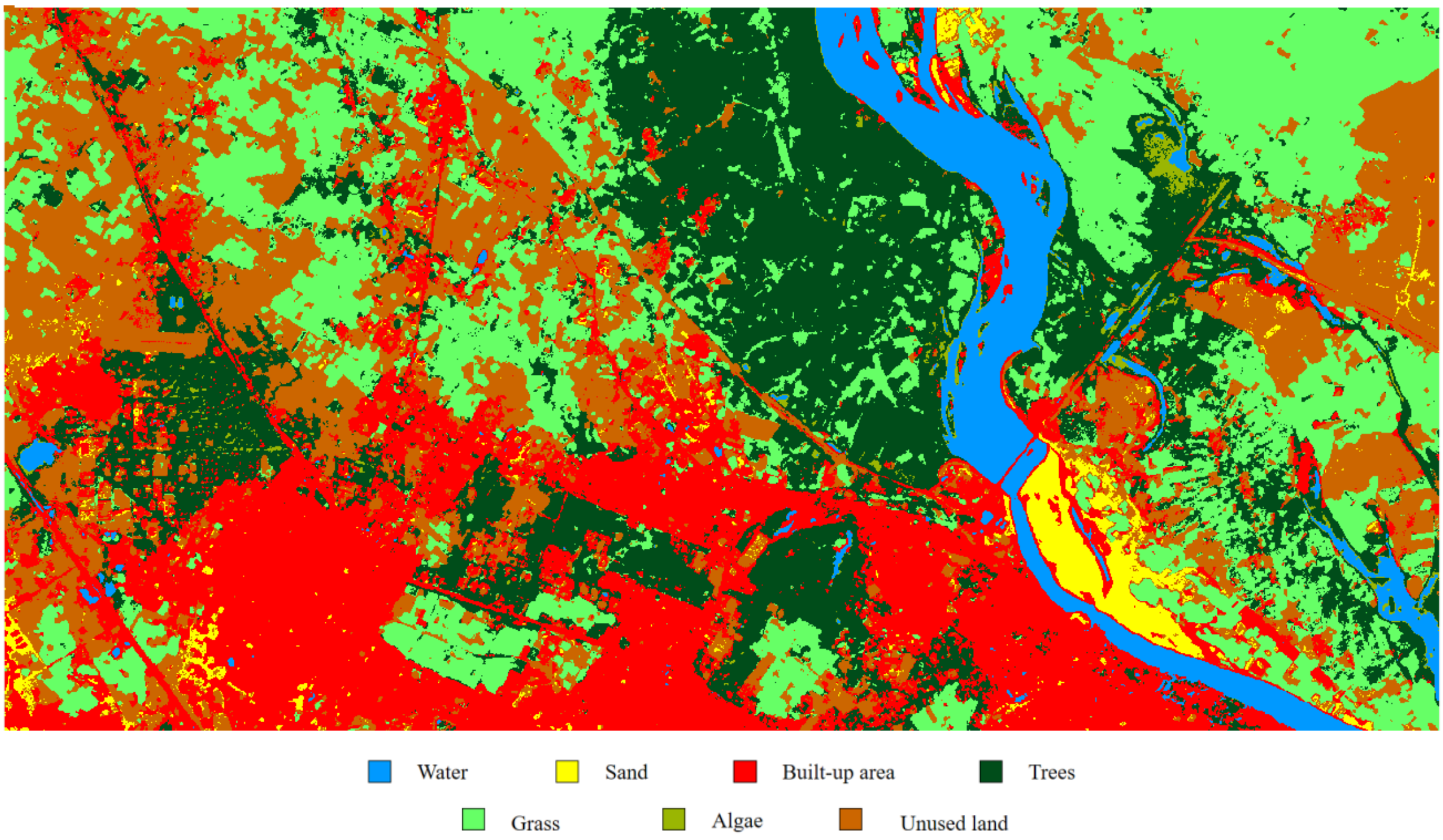

*Figure 50: Classified imagery for VNIR-SWIR bands fused with texture band using CRP-RFE*

Following inferences can be drawn from the classified imageries:

a. The classified imageries obtained from SAR bands for each of the classifier look least similar to the ground truth image generated while the ones obtained from VNIR-SWIR bands staked with texture band bear the most resemblance to the ground truth. The classified imageries obtained from VNIR-SWIR fused with texture band comes second in resembling the ground truth. The ones obtained from only VNIR-SWIR stand third while the ones from SAR dataset stacked with texture stand fourth.

b. The edges are most distinctively visible in the case of imageries obtained from VNIR-SWIR bands stacked with texture band. This could be observed clearly at the demarcation of *sand* class and *water* class.

c. In almost all the cases the commission error is highest among the algae class that is though it correctly identified in the regions it is actually present in, it is also included in the regions where it is not expected to grow.

# 6 Conclusion and future scope

This study attempts to generate ensembles of random forest classifier using the concept of Forest-RC algorithm by rotation of the training dataset in order to overcome the issue of stagnation of accuracy in random forests and harness the advantage of the linear combination of features in classifying the satellite imagery. In addition, to study the effect of image fusion and textural features on image classification, texture bands generated from Sentinel-1 (SAR) bands have been incorporated with Sentinel-2 (VNIR-SWIR) bands using stacking as well as image fusion. The image classification results for SAR and VNIR-SWIR bands (both with and without texture information) are compared with each other for all the classifiers. Also, as an extra effort, the code for the Bayesian fusion algorithm is speeded up (about **3000** times for 700 x700 image with 11 bands).

## 6.1 Conclusions

Following conclusions can be drawn from the results obtained:

a. The lowest classification accuracies are obtained for SAR bands because the number of features are quite less (originally only two, VH and VV while rest four are derived from them) and there is neither any spectral nor any textural information involved with it. When texture information is added to them, classification accuracies show an improvement. Similarly, for VNIR-SWIR bands, though the classification accuracies are much higher than the previous two cases because of incorporation of spectral features, yet they are still less when texture information is added to them. Also, the accuracies obtained for the product created by stacking texture band to VNIR-SWIR bands is more than their fused product because in the former case, 100% of information from both the kinds of bands is being used whereas in the latter case, only a fraction of information is being harnessed from both kinds of bands (i.e. 60% from texture band whereas 40% form VNIR-SWIR bands).

b. While comparing the random forest ensembles with the single random forest classifier, it is observed that the ensembles do not perform significantly better in case of only SAR bands and SAR bands stacked with texture. This could be due to the reason the additional bands in SAR are created by the combinations of existing bands so the effect of random rotations is suppressed. The classifiers have shown to perform quite well in case of VNIR-SWIR, VNIR-SWIR stacked with texture and VNIR-SWIR fused with texture. Among the three classifiers, CRP-RFE has performed better than PCA-RFE and SRP-RFE with highest average overall kappa value of **96.93%** for 20% training data, whereas SRP-RFE outperforms only PCA-RFE. The reason could be that the degree of randomization is much more in case of CRP-RFE, whereas SRP-RFE stand second in that respect as well.

c. The performance of the ensembles is observed to be better for the classes with less training samples such as *sand* and *algae*. This shows that incorporation of random combination in the training dataset tends to generate the additional information that helps in better classification. Also, the ensemble classifiers, especially in case of VNIR-SWIR stacked with texture band, seem to perform better in demarcation the boundaries between the classes.

d. The accuracies of random forest ensembles also tend to stagnate after a certain value. These values are shown in table 31.

e. The execution time for all the three ensemble classifiers is observed to be almost same. Hence, they seem to be equally computationally expensive.

## 6.2 Future scope of work

Following points can be taken up as the future scope of the present work:

a. The classes in this study lie in spectrally wide region. The classifiers used in this study can be used with classes having narrow spectral separation such as those in farmlands with different crops.
b. While working with Sentinel-2 imagery, vegetation indices can be incorporated as well to see how they would affect the classification accuracy.
c. This thesis does not discuss any other fusion technique besides Bayesian fusion. One can make use of other fusion techniques along with classification ensembles and observe their effects on the accuracy of classification.
d. This thesis does not focus on object based image analysis. Hence, segmentation based approaches can also be incorporated with these classifiers along with spectral bands.
e. From the perspective of ensemble classifiers, other multivariate data analysis techniques such as canonical correlation analysis or feature extraction techniques such as independent component analysis can also be used to create classification ensembles.